\documentclass[runningheads]{llncs}

\usepackage{eccv}

\usepackage{eccvabbrv}

\usepackage{graphicx}
\usepackage{booktabs}
\usepackage{multirow}
\usepackage{multicol}
\usepackage{wrapfig}
\usepackage{algorithm}
\usepackage{algpseudocode}

\usepackage[accsupp]{axessibility}  

\usepackage{hyperref}

\usepackage{orcidlink}

\DeclareMathOperator{\softmax}{softmax}
\DeclareMathOperator{\interpolate}{interpolate}
\DeclareMathOperator{\mlp}{MLP}
\DeclareMathOperator{\sampling}{sampling}
\DeclareMathOperator{\act}{act}
\DeclareMathOperator*{\argmax}{argmax} 

\newcommand{\fra}[1]{\textcolor{black}{#1}}

\begin{document}

\title{Boost Your NeRF: A Model-Agnostic Mixture of Experts Framework for High Quality and Efficient Rendering} 

\titlerunning{Boost Your NeRF}

\author{Francesco Di Sario\inst{1}\orcidID{0009-0005-6969-1246} \and\\
Riccardo Renzulli\inst{1}\orcidID{0000-0003-0532-5966} \and\\  
Enzo Tartaglione\inst{2}\orcidID{0000-0003-4274-8298} \and\\
 Marco Grangetto\inst{1}\orcidID{0000-0002-2709-7864}}

\authorrunning{F. Di Sario et al.}

\institute{University of Turin, Italy \and
LTCI, T\'el\'ecom Paris, Institut Polytechnique de Paris, France
\email{francesco.disario@unito.it}}
\maketitle

\begin{abstract}
Since the introduction of NeRFs, considerable attention has been focused on improving their training and inference times, leading to the development of Fast-NeRFs models. Despite demonstrating impressive rendering speed and quality, the rapid convergence of such models poses challenges for further improving reconstruction quality. Common strategies to improve rendering quality involves augmenting model parameters or increasing the number of sampled points. However, these computationally intensive approaches encounter limitations in achieving significant quality enhancements. This study introduces a model-agnostic framework inspired by Sparsely-Gated Mixture of Experts to enhance rendering quality without escalating computational complexity. Our approach enables specialization in rendering different scene components by employing a mixture of experts with varying resolutions. We present a novel gate formulation designed to maximize expert capabilities and propose a resolution-based routing technique to effectively induce sparsity and decompose scenes. Our work significantly improves reconstruction quality while maintaining competitive performance. Project page: \url{https://eidoslab.github.io/boost-your-nerf/}.
\end{abstract}

\section{Introduction}
Neural Radiance Fields (NeRFs)~\cite{mildenhall2020nerf} have recently shown impressive results in synthesizing photo-realistic 3D scenes from a set of 2D images. However, NeRFs suffer from limited scene diversity, long training time, and sensitivity to training data~\cite{liu2021probabilistic}.
Since the introduction of NeRFs, significant attention has been directed toward improving their training and inference times, resulting in the development of Fast-NeRFs. \fra{
By using auxiliary data structures such as voxel grids to store scene geometry and skip empty spaces, the training and inference process can be accelerated by several orders of magnitude. As a result, the neural component of these models, which is usually used to transform learned features into view-dependent color representations, becomes much smaller and can sometimes be replaced entirely by spherical harmonics~\cite{fridovich2022plenoxels}.
}
Despite Fast NeRFs being an improvement from the original NeRF framework, achieving high-quality and efficient reconstructed geometry is still an open problem.

\begin{figure*}[t]
\centering
  \begin{minipage}[b]{0.32\textwidth}
    \includegraphics[width=\linewidth]{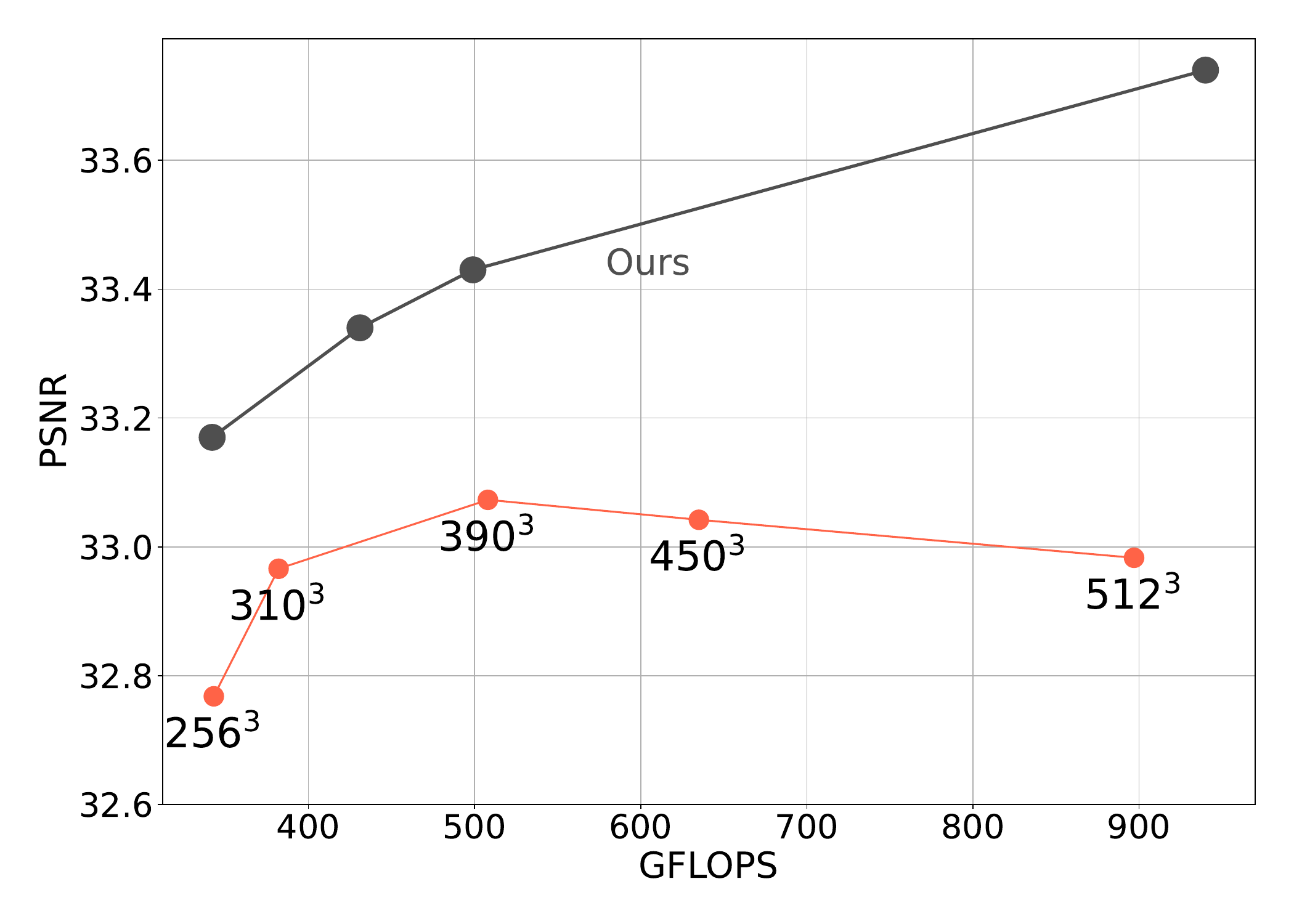}
 
    \label{fig:fig1.1}
  \end{minipage}
    \begin{minipage}[b]{0.32\textwidth}
    \includegraphics[width=\linewidth]{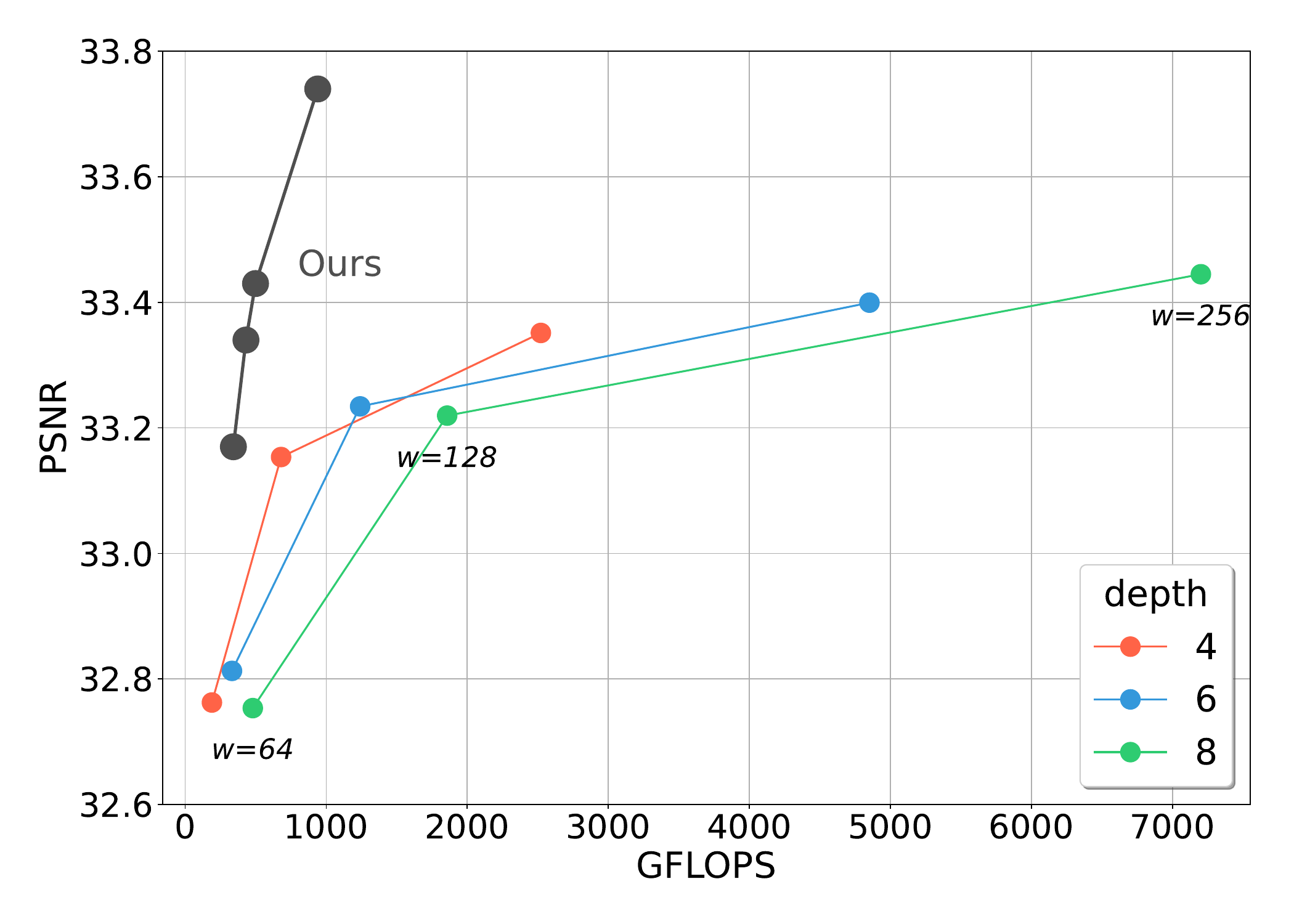}

    \label{fig:fig1.2}
  \end{minipage}
    \begin{minipage}[b]{0.32\textwidth}
    \includegraphics[width=\linewidth]{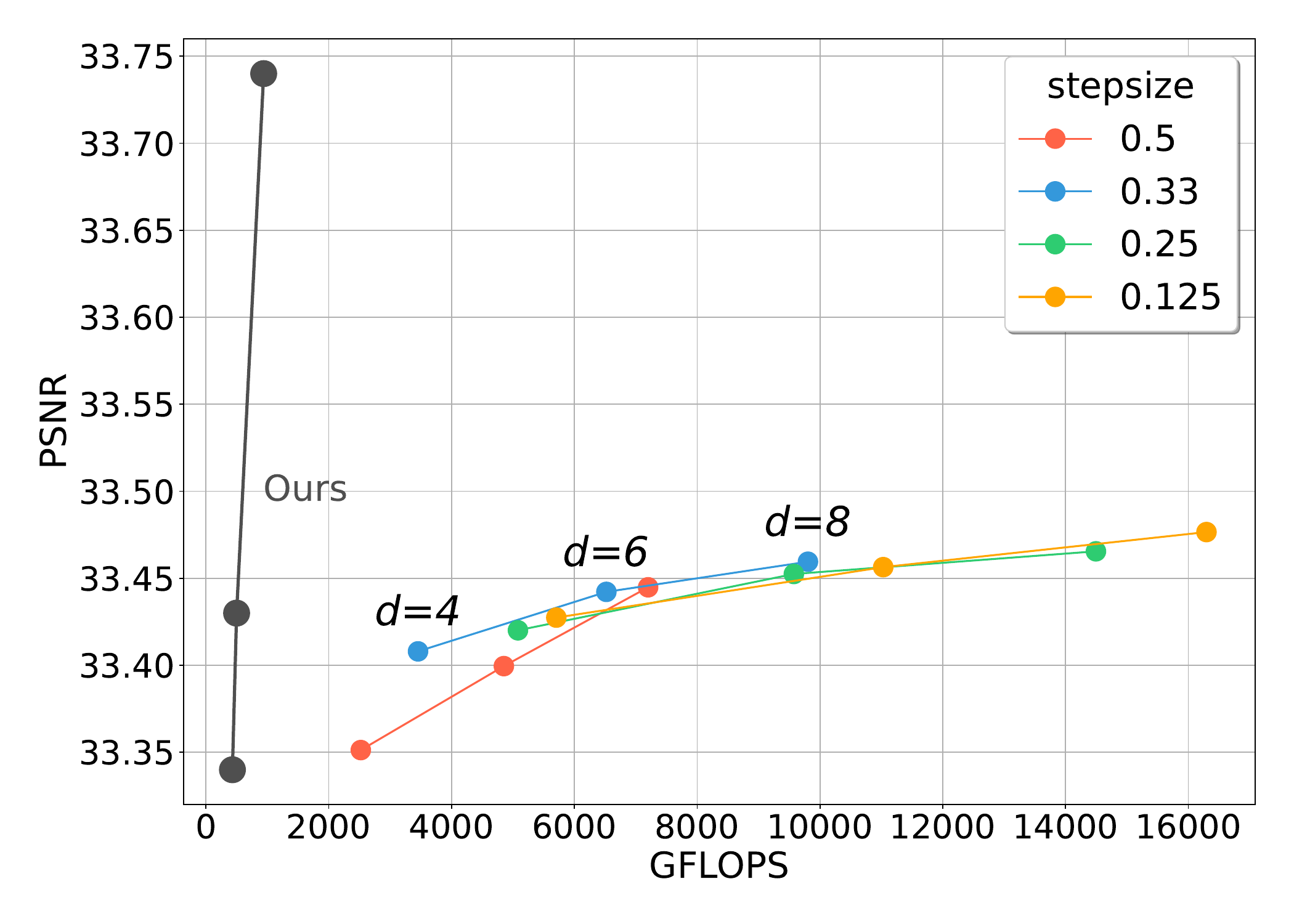}
  \end{minipage}

\caption{Different strategies for improving reconstruction quality with Fast-NeRFs (DVGO~\cite{sun2022direct}). 
Increasing the resolution of data structures like voxel grids can improve render quality, but only up to a certain point, after which quality declines (left). The MLP component's impact on rendering quality is analyzed by varying its depth and width while keeping other variables constant (center). The effect of the number of sampled points along each ray is examined by decreasing the step size (right). In gray, our method's performance, which significantly improves PSNR at a low computational cost.}
\label{fig:fig1}
\end{figure*}

Typical ``naive'' approaches to enhance the reconstruction quality of these models include:

\begin{enumerate}
    \item Increasing the parameters and resolution of the used data structures (e.g., voxel grid, hash grid, etc.).
    \item Increasing the number of sampled points per ray.
    \item Increasing the number of parameters in the neural network or the order of spherical harmonics.
\end{enumerate}
However, as evident from Figure~\ref{fig:fig1}, the increase in reconstruction quality results in a significant increase in computational costs.
 In the first approach, increasing the resolution can lead to a significant improvement in the reconstruction quality, but a plateau is reached beyond which overfitting occurs, and reconstruction quality degrades. This comes at the expense of a considerable increase in spatial complexity (especially with dense voxel grids) and training times. Increasing the number of sampled points per ray can marginally improve reconstruction quality further but at a significant increase in computational complexity. In fact, the higher the number of sampled rays, the higher forward passes through a neural network are needed. Augmenting the number of parameters in the neural network is another solution. However, as the neural network grows in depth and width, it tends towards a fully implicit model, deviating from the principles of fast models (limiting the neural part as much as possible - or even removing it altogether). 

We propose a technique capable of significantly enhancing the reconstruction quality of such models while maintaining competitive training and rendering times. Inspired by the Sparse Mixture of Experts (MoE) paradigm~\cite{shazeer2017outrageously}, we have developed a model-agnostic framework capable of improving the reconstruction quality of various state-of-the-art models. Intuitively, our mixture of experts consists of different-capacity (resolution) models. During training, each model specializes in rendering the most suitable parts of the scene, \ie low-resolution models render low-frequency parts of the scene, while high-resolution models render high-frequency parts. Our formulation of the gate allows our method to be model-agnostic, enabling, on the one hand, the insertion of the gate function at the early stages of the MoE mechanism and, on the other hand, the multiplication of the gate's output with the experts' output as late as possible. This maximizes the capabilities of the mixture of experts and allows for working directly with the output of each expert, which is considered a black box. This is why our method is inherently model-agnostic.
In summary, our contributions are as follows:
\begin{enumerate}
    \item We propose the first model-agnostic framework based on Sparse MoE of models at different resolutions, which significantly improves the rendering quality of such models while maintaining competitive training and inference times (Sec.~\ref{sec:method}).
    \item We provide a novel gate formulation inspired by Fast-NeRF models, which maximizes the capability of the mixture of experts (Sec.~\ref{subsec:gate}).
    \item We introduce a new routing technique based on resolution, favoring the assignment of tokens to low-resolution models and discouraging the assignment of tokens to high-resolution models, inducing increasing sparsity in high-resolution models and decomposing the scene based on frequency (Sec.~\ref{subsec:smoe}).
    \item We conduct extensive experiments to test our method, including different NeRFs architectures and datasets, showing higher rendering quality and efficiency (Sec.~\ref{sec:experiments}).
\end{enumerate}

\section{Related works}
\label{sec:related}

\subsection{Neural Radiance Field}\label{sec:nerf_basics}
NeRF~\cite{mildenhall2020nerf} (Neural Radiance Field) has emerged as a prominent method for synthesizing novel views, showing significant progress. This approach requires a moderate number of input images along with their known camera poses. Unlike traditional methods that rely on explicit and discretized volumetric representations such as voxel grids and multiplane images, NeRF employs a coordinate-based multilayer perceptron (MLP) to create an implicit and continuous volumetric representation.
NeRFs can represent a 3D scene as a MLP $F_{\theta}$, with $\theta$ being the set of its trainable parameters, such that $F_{\theta}: (\mathbf{x}, \mathbf{d}) \rightarrow (\mathbf{c}, \sigma)$ which maps $(\mathbf{x}, \mathbf{d})$, a 3D position $\mathbf{x}$ and a viewing direction $\mathbf{d}$, to a view-dependent color emission $\mathbf{c}$ and density value $\sigma$.
To render the color of a pixel $\hat{C}(\mathbf{r})$, a ray $\mathbf{r}$ traverses the center of the camera through the pixel of the image plane from an origin point $\mathbf{o}$ to a ray having position $\mathbf{r}(t) = \mathbf{o} + t\mathbf{d}$. Then, $N$ points are sampled on $\mathbf{r}$, and the MLP is queried for each point, obtaining a density and a color value. Finally, these results are accumulated into a single color with the volume rendering equation~\cite{max1995optical}:

\begin{equation}\label{eq:vreq}
            \hat{{C}}(\mathbf{r}) = \sum_{i=1}^{K}T_i\alpha_ic_i,\qquad
            T_i = \prod_{j=1}^{i-1}(1 - \alpha_j), \qquad
            \alpha_i = 1 - \exp(\sigma_i\delta_i),
\end{equation}

where $\alpha_i$ is the probability of termination at point $i$ and $T_i$ is the accumulated transmittance from the near plane to point $i$.
NeRFs are trained by minimizing a photometric loss, which is an L2 loss between the rendered and the ground truth pixels. In more detail, given a batch $B$ of randomly sampled rays, the loss is defined as

\begin{equation}
    L_{\text{nerf}} = \frac{1}{|B|}\sum_{r \in B}^{} \left\| \hat{C}(\mathbf{r}) - C(\mathbf{r}) \right\|^2_2,
\end{equation}

where $\hat{C}(\mathbf{r})$ is the predicted color and $C(\mathbf{r})$ the ground truth color for the ray $\mathbf{r}$.

\subsection{Fast Neural Radiance Fields}
Since the introduction of NeRF, significant effort has been directed towards the development of faster models. \fra{Several studies focus on speeding up rendering times, for example by working on ray sampling efficiency~\cite{chan2021pi, deng2022gram, hou2023infamous, zhuang2022mofanerf }, by integrating explicit volumetric representations~\cite{garbin2021fastnerf, jeong2021self, wizadwongsa2021nex, yu2021plenoctrees, xian2021space, garbin2021fastnerf, chen2023mobilenerf} or by utilizing thousands of tiny MLPs~\cite{reiser2021kilonerf}}. However, all these methods still require large training times.
One noteworthy development is represented by the introduction of explicit volumetric representations in the training pipeline, directly optimizing such representations~\cite{fridovich2022plenoxels, sun2022direct, chen2022tensorf, mueller2022instant, sfk_kplanes_2023, hu2023tri, kerbl3Dgaussians}. These models enable fast training and inference, with render quality only slightly inferior to full-implicit models~\cite{barron2021mip, barron2023zip}. We refer to these as \emph{Fast-NeRFs}. 
Plenoxels~\cite{fridovich2022plenoxels} is the first important work in this context, as it represents a scene as a sparse 3D grid where each voxel stores spherical harmonic coefficients and density. Spherical harmonics serve as an orthogonal basis for functions defined over the sphere and thus can be used for computing view-dependent color emission, without the need for a multi-layer perceptron. Additionally, Plenoxels demonstrated the advantages of linearly interpolating voxels, facilitating the learning of a continuous plenoptic function throughout the volume, akin to NeRF, albeit with discrete data.
Another notable work is DVGO~\cite{sun2022direct}. Each scene is there represented as two dense voxel grids (one for density and one for feature colors) alongside a MLP, for learning view-dependent color.
Similar to Plenoxels, the value of each voxel is linearly interpolated with the 8 nearest voxels, but after applying the activation functions. DVGO also incorporates a preliminary coarse geometry stage to learn the scene's general structure, facilitating adjustments to the bounding box and enabling a more intelligent ray sampling strategy. Despite utilizing lower-resolution models, DVGO achieves high reconstruction quality.
The major drawback of these models is their substantial memory storage requirements.
TensoRF~\cite{chen2022tensorf} addresses this challenge by replacing the dense voxel grid with a planes and vectors decomposition, significantly reducing storage demands while maintaining comparable performance and rendering quality.
Alternatively, Instant-NGP~\cite{mueller2022instant} proposed a multi-resolution voxel grid encoded via a hash function. They define $L$ hash grids of increased resolutions: each entry of each hash grid has $2^T$ parameters and $F$ features. This enables even faster training times and real-time rendering performances while maintaining a compact model; moreover, it demonstrates the efficacy of a multi-resolution approach in enhancing render quality. Inspired by this methodology, K-Planes~\cite{sfk_kplanes_2023} introduces a multi-resolution planar factorization of 3D space, offering reconstruction quality and model compactness similar to the previous methods, but with slightly larger training times. A similar multi-resolution planar factorization has also been proposed in Tri-MipRF~\cite{hu2023tri} for mitigating the aliasing in distant or low-resolution views and blurriness in close-up shots.
Given their properties, we decided to evaluate our paradigm using three different models: a 3D dense voxel grid-based model (DVGO), a decomposed grid-based model (TensoRF), and a multi-resolution hash grid-based model (Instant-NGP).

\subsection{Mixture of Experts and Sparse MoE}
The Mixture of Experts (MoE) paradigm~\cite{jacobs1991adaptive} 
has gained prominence in various machine learning applications. MoE consists of multiple expert networks, each specializing in different regions of the input space, with a meta-network determining the contribution of each expert to the final prediction. 
Building upon the MoE framework, Sparse Mixture of Experts~\cite{shazeer2017outrageously, lepikhin2020scaling, riquelme2021scaling, zhou2022mixture } constitutes a scalable and efficient variant.
At the core of all Sparse MoE algorithms lies an assignment problem between tokens and experts. One approach to tackle this is by approximating the solution with a gating function, which learns to assign input tokens to the most suitable experts. It typically comprises a linear layer, a softmax activation, and a Top-$K$ (where $1 \leq K \leq 2$) operation, aimed at routing the input token to only a subset of the experts. To balance the assignment across all the experts, auxiliary loss functions penalizing unevenly distributed routing are often employed. This sparsity-inducing technique significantly reduces computational overhead while preserving the expressive power of the MoE architecture.

\subsection{MoE and NeRF}

Combining the sparse mixture of experts' paradigms with neural radiance fields presents a non-trivial challenge. In recent research, Switch-NeRF~\cite{zhenxing2022switch} has been introduced as a novel end-to-end large-scale NeRF with learning-based scene decomposition. Inspired by Fedus~\etal~\cite{fedus2022switch}, they propose a full-implicit model with an MLP-based gate comprising 4 layers with 128 neurons and a Top-$1$ function. However, their architecture is ad-hoc, and suffers from extensive training times. 
This poses a challenge for us, as we aim to consider each expert as a black box, taking a point in space and a direction as input and outputting radiance and density values. Moreover, as the mixture comprises experts at different resolutions, we also aim to prioritize routing input tokens toward lower-resolution models and minimize the usage of high-resolution models.

Our work intends to overcome these limitations. We design a novel gate formulation inspired by Fast-NeRFs, that guarantees fast convergence and better performances. We position the gate at the beginning of our MoE pipeline and postpone the multiplication of the gate output with expert predictions as much as possible. This allows our framework to be entirely model-agnostic. Furthermore, our mixture of experts maintains competitive training times, enabling, in the worst case, training models on complex scenes in approximately 1 hour while achieving state-of-the-art accuracy.

\section{Method}\label{sec:method}

In this section, we introduce our model-agnostic ensembling approach. We present in Sec.~\ref{subsec:overview} an overview of our method. We also describe the ray sampling and filtering mechanism in Sec.~\ref{subsec:raysamplfilt}. Then, in Sec.~\ref{subsec:gate}, we discuss the effective construction of a mixture of experts with a gating mechanism in the context of NeRF. Finally, in Sec.~\ref{subsec:smoe}, we present a novel formulation for decomposing the scene into resolution-based parts with an end-to-end training procedure.
\subsection{Overview}\label{subsec:overview}
\begin{figure}[t]
    \centering
    \includegraphics[width=1\textwidth]{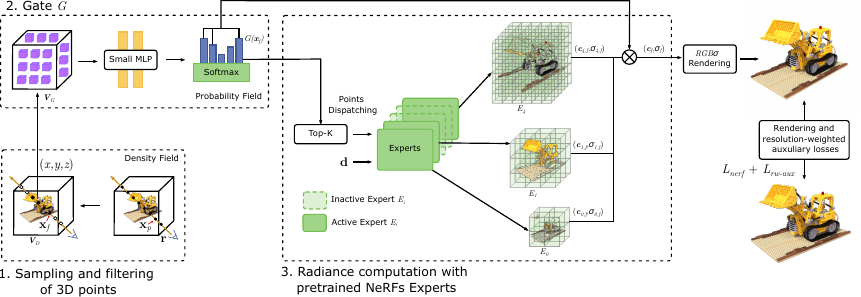}
    \caption{
    A density field is used to compute density values for sampled points along a ray. A filtering step discards low-density points, routed through gating network $G$ for expert assignment. Top-$K$ experts compute radiance and density, aggregating and weighting these values by the corresponding gating probability to get the final values $\mathbf{c}_f$ and $\sigma_f$. Volume rendering equation yields pixel colors, and joint optimization refines our resolution-weighted auxiliary loss, allowing for high-quality and efficient rendering.}
    \label{fig:method}
\end{figure}
Figure~\ref{fig:method} shows a high-level scheme of our method.
After independently training a set of $M$ NeRF models $\{E_{i}\}_{i=1}^{M}$ at different resolutions, we sample $N$ points along a ray $\mathbf{r}$. For each point $\mathbf{x}_p$, a density value $\sigma$ is calculated, and based on this, a filtering step is performed based on the density volume $\mathbf{V}_D$, discarding points in regions with a density below a certain threshold $T$. Initializing the density volume $\mathbf{V}_D$ with the one learned from the lowest resolution model can be helpful (though it can also be learned). Subsequently, the filtered point $\mathbf{x}_f$ is fed to the gate $G$, so we denote with $G(\mathbf{x}_f)$ the probabilities with which the point is assigned to the $M$ experts. Based on these values, each point is routed to the Top-$K$ experts. Each selected expert then computes the radiance $\mathbf{c}_{i,f}$ and density values $\sigma_{i,f}$ for the point, both of which are multiplied by their respective probabilities and summed together to get the final radiance $\mathbf{c}_f$ and density $\sigma_f$. 
Subsequently, the volume rendering Equation~\ref{eq:vreq} is used to aggregate all colors and compute the pixel color for the given ray. Finally, we compute the total loss and jointly optimize the gate and the experts.

In the next sections, we will dive into the gating and sparse mixture of expert modules. 

\subsection{Ray Sampling and Filtering}\label{subsec:raysamplfilt}
The first phase of our method involves learning a coarse and explicit density field starting from the density volume $\mathbf{V}_D$, that we can leverage for skipping empty spaces. Given a ray $\mathbf{r}$ as explained in Sec.~\ref{sec:nerf_basics}, we sample $N$ points along it:
\begin{equation} 
    \mathbf{X}_{p,r} = \sampling(\mathbf{r}) \in \mathbb{R}^{N\times3},
\end{equation}
with $p \in [0,...,N-1]$. We suppress the index $r$ for abuse of notation.
Next, we compute the density for each point $\mathbf{x}_p \in \mathbb{R}^3$ and discard those with negligible density. The density value $\mathbf{\sigma}_{D, p}$ for each point is computed by linearly interpolating $v$ neighboring voxels:

\begin{equation}
    \sigma_{D, p} = \act(\interpolate(\mathbf{x}_p, V_D)) \in \mathbb{R}
\end{equation}
where $\act$ represents a density activation function (such as softplus).
We denote with $\mathbf{x_f \in \mathbb}{R}^{3}$ a remaining point after the filtering operation. 

\subsection{Trainable Gating Model}\label{subsec:gate}
Our gating mechanism incorporates a hybrid architecture: an explicit feature grid $\mathbf{V}_G$ and a shallow MLP. The gating mechanism can also be seen as a probability field. First, we compute per-point features:
\begin{equation}
    \mathbf{feat} = \interpolate(\mathbf{x}_f, \mathbf{V}_G)    \in \mathbb{R}^{C}
\end{equation}
where $C$ represents the number of channels of $\mathbf{V}_G$. Subsequently, we transform each feature into a probability. We compute logits as

\begin{equation}
    \mathbf{logits} = \mlp(\mathbf{feat}) \in \mathbb{R}^{M}
\end{equation}
where $M$ is the number of experts in the mixture of experts. Finally, we apply a per-row softmax. Our gating function can be summarized as:

\begin{equation}
    G(\mathbf{\mathbf{x}_f}) = \softmax(\mlp(\interpolate(\mathbf{x}_f, \mathbf{V}_G))) \in \mathbb{R}^{M}
\end{equation}

\subsection{NeRFs Experts}
As mentioned in Sec.~\ref{subsec:overview}, we employ a set of $M$ pretrained NeRFs models as experts. After feeding as input each point $\mathbf{x}_f$ to the gate, we route them to the $k$ experts with the $k$ highest probabilities $G(\mathbf{x}_f)$. We define this set of indexes of the selected experts as
\begin{equation}
    \mathcal{K} = \argmax_{Top-k}(G(\mathbf{x}_f)).
\end{equation}
Each expert is a Fast-NeRF model that receives the dispatched points and the direction of the ray they lie on as input and outputs a density and a radiance value. We treat each expert as a black box, ensuring our method is inherently model-agnostic. The radiance $\mathbf{c}_f$ and the density $\sigma_f$ for the point $\mathbf{x}_f$ laying on a ray with direction $\mathbf{d}$ can be expressed as the sum of each expert predictions weighted by their probability:

\begin{equation}
    \mathbf{c}_f, \sigma_f = \sum_{i \in \mathcal{K}} E_i(\mathbf{x_f}, \mathbf{d}) \cdot G(\mathbf{x_f})_{i}
\end{equation}

where $E_i$ denotes the $i$-th expert and $G(\mathbf{x_f})_{i}$ the probability for the point to be dispatched to that expert $i$.

Once we obtain radiance and density values for all the points on the ray, we can compute the pixel color with Equation~\ref{eq:vreq}.

\subsection{Resolution-based Routing}\label{subsec:smoe}
To balance the load and prevent the gate from focusing on assigning points to a single expert (typically the one with the highest resolution), we employ a resolution-weighted auxiliary loss. Given $M$ experts, and a batch of points $B$, the auxiliary loss is defined as:

\begin{equation}
    L_{\text{aux}} = \frac{M}{|B|^2}\sum_{i=1}^{M}c_im_i, \qquad  
    m_i = \sum_{\mathbf{x}_f \in B} G(\mathbf{x}_f)_{i},
\end{equation}

where $c_i$ represents the number of inputs dispatched to the expert $i$, and $m_i$ is the sum of all the probabilities for each point in the batch for the expert $i$.
This loss helps balance the workload, ensuring that each expert processes a similar number of points. Ideally, in a perfectly balanced scenario, $L_{aux}$ is expected to be 1, as both $c_i$ and $m_i$ would be $\frac{M}{N}$, resulting in $\sum_{i=1}^{M}c_im_i$ being $\frac{M^2}{N}$. However, in the case of total imbalance, it tends to the number $M$.
We aim to take a step further by assigning as many points as possible to low-resolution experts while discouraging point assignment to high-resolution models. Hence, we introduce some penalty terms $w_i$ associated with each expert. The higher the resolution of the model, the higher the penalty. We define a novel resolution-based auxiliary loss:
\begin{equation}
    L_{\text{rw-aux}} = \frac{M}{|B|^2}\sum_{i=1}^{M}c_im_iw_i. \qquad
\end{equation}

As for the weighting strategy, we opted for the following geometric progression. The weight $w_i$ for the $i$ expert is computed as:

\begin{equation}
    w_i = \exp\left(\frac{\ln M}{M-1}\right)^i,\quad i \in [0,...,M-1].
\end{equation}

The total loss is then defined as:
\begin{equation}
    L_{\text{tot}} = L_{\text{nerf}} + \lambda L_{\text{rw-aux}}.
\end{equation}
As we will see in Sec.~\ref{sec:experiments}, our proposed loss not only further improves the reconstruction quality but also encourages sparsity in higher-resolution models. A comparison with other weighting strategies is proposed in the Supplementary~\ref{sec:penalties}.

\section{Experiments}\label{sec:experiments}

\subsection{Setting}
The code was written in Python 3.8, using PyTorch 1.12, and executed on a single NVIDIA A40 GPU. We tested our technique on the DVGO, TensoRF, and Instant-NGP architecture.
The code from official repositories was used as the starting point. For Instant-NGP, the native implementation in PyTorch \texttt{ngp\_pl} was employed, which exhibits comparable performance and reconstruction quality to the official NVIDIA implementation.
We experimented with various configurations, ranging from a minimum of 3 models to 5 models of different resolutions. The experts are ordered by resolution, such that the $i + 1$-th expert has approximately double the parameters of the $i$-th expert. 
For all configurations and models, $\lambda$ was set to $10^{-3}$. The Gate consists of a grid (dense voxel grid for DVGO, factorized grid for TensoRF, and hash grid for Instant-NGP) and a shallow MLP (2 layers with 64 neurons each) with ReLU activation function. The resolution of the gate is low: $128^3$ for both DVGO and TensoRF, while for Instant-NGP we used $L = 6$. 
The number of iterations is set to $20k$ for all the architectures. 
All other hyperparameters are left as the original implementations. 
We draw experiments with our method using Top-$k$ experts, with $k=1$ and $k=2$. We compare our results versus baselines with comparable resolutions, as well as a Fast-NeRF ensemble (\textit{Ens}) obtained by jointly fine-tuning all models and averaging their predictions. It can be noted that this ensemble can be interpreted as a limit case for our method with $k=M$ and  $G(\mathbf{x}_f)_i = 1/M, \forall i$ similar to the method proposed by~\cite{di2023two}. 

\begin{table}[t]
\caption{Results on DVGO, TensoRF and Instant-NGP with $M=5$. $\|w\|_0$ is expressed as multiple of $10^6$, while for Instant-NGP as a multiple of $10^5$. }
\resizebox{\textwidth}{!}{%
\begin{tabular}{cccccccccccccc}
\hline
\multirow{2}{*}{\textit{Dataset}}                      & \multirow{2}{*}{Metrics}             & \multicolumn{4}{c}{\textit{\textbf{DVGO}}}                                                & \multicolumn{4}{c}{\textit{\textbf{TensoRF}}}                                             & \multicolumn{4}{c}{\textit{\textbf{Instant-NGP}}}                    \\ \cline{3-14} 
                                                       &                                      & \textit{baseline} & \textit{Top-1} & \textit{Top-2} & \multicolumn{1}{c|}{\textit{Ens}}   & \textit{baseline} & \textit{Top-1} & \textit{Top-2} & \multicolumn{1}{c|}{\textit{Ens}}   & \textit{baseline} & \textit{Top-1} & \textit{Top-2} & \textit{Ens}   \\ \hline
\multicolumn{1}{c|}{\multirow{6}{*}{\textit{Blender}}} & \multicolumn{1}{c|}{$\mathbf{PSNR} \uparrow$}   & 33.04             & 33.43          & 33.74          & \multicolumn{1}{c|}{\textbf{33.79}} & 32.98             & 33.68          & \textbf{34.09} & \multicolumn{1}{c|}{34.00}          & 33.35             & 33.56          & 33.83          & \textbf{34.01} \\
\multicolumn{1}{c|}{}                                  & \multicolumn{1}{c|}{$\mathbf{SSIM}\uparrow$}   & 0.961             & 0.964          & \textbf{0.965} & \multicolumn{1}{c|}{\textbf{0.966}} & 0.958             & 0.965          & \textbf{0.968} & \multicolumn{1}{c|}{0.968}          & 0.963             & 0.963          & 0.965          & \textbf{0.966} \\
\multicolumn{1}{c|}{}                                  & \multicolumn{1}{c|}{$\mathbf{LPIPS}\downarrow$}  & 0.026             & 0.024          & \textbf{0.022} & \multicolumn{1}{c|}{\textbf{0.022}} & 0.029             & 0.023          & \textbf{0.021} & \multicolumn{1}{c|}{0.022}          & \textbf{0.025}    & 0.045          & 0.043          & 0.042          \\
\multicolumn{1}{c|}{}                                  & \multicolumn{1}{c|}{$\|w\|_0\downarrow$}    & 99                & \textbf{26}    & 39             & \multicolumn{1}{c|}{97}             & 40                & \textbf{24}    & 33             & \multicolumn{1}{c|}{49}             & 31                & \textbf{17}    & 21             & 26             \\
\multicolumn{1}{c|}{}                                  & \multicolumn{1}{c|}{$\mathbf{GFLOPs}\downarrow$} & 635               & \textbf{499}   & 940            & \multicolumn{1}{c|}{2206}           & 886               & \textbf{732}   & 1344           & \multicolumn{1}{c|}{2214}           & \textbf{46}       & 71             & 123            & 208            \\
\multicolumn{1}{c|}{}                                  & \multicolumn{1}{c|}{$\mathbf{Time}\downarrow$}   & 26'               & \textbf{21'}   & 25'            & \multicolumn{1}{c|}{32'}            & \textbf{44'}      & 69'            & 76'            & \multicolumn{1}{c|}{70'}            & \textbf{11'}      & 32'            & 34'            & 37'            \\ \hline
\multicolumn{1}{c|}{\multirow{6}{*}{\textit{NSVF}}}    & \multicolumn{1}{c|}{$\mathbf{PSNR}\uparrow$}   & 35.21             & 37.12          & 37.59          & \multicolumn{1}{c|}{\textbf{37.68}} & 36.70             & 37.40          & 37.98          & \multicolumn{1}{c|}{\textbf{38.08}} & 36.44             & 36.59          & 37.04          & \textbf{37.31} \\
\multicolumn{1}{c|}{}                                  & \multicolumn{1}{c|}{$\mathbf{SSIM}\uparrow$}   & 0.977             & 0.984          & \textbf{0.986} & \multicolumn{1}{c|}{\textbf{0.986}} & 0.981             & 0.984          & 0.986          & \multicolumn{1}{c|}{\textbf{0.987}} & 0.983             & 0.983          & 0.984          & \textbf{0.985} \\
\multicolumn{1}{c|}{}                                  & \multicolumn{1}{c|}{$\mathbf{LPIPS}\downarrow$}  & 0.015             & 0.009          & 0.008          & \multicolumn{1}{c|}{\textbf{0.007}} & 0.013             & 0.009          & \textbf{0.008} & \multicolumn{1}{c|}{\textbf{0.008}} & \textbf{0.010}    & 0.023          & 0.022          & 0.020          \\
\multicolumn{1}{c|}{}                                  & \multicolumn{1}{c|}{$\|w\|_0\downarrow$}    & 95                & \textbf{27}    & 43             & \multicolumn{1}{c|}{100}            & 42                & \textbf{28}    & 38             & \multicolumn{1}{c|}{54}             & 30                & \textbf{17}    & 20             & 27             \\
\multicolumn{1}{c|}{}                                  & \multicolumn{1}{c|}{$\mathbf{GFLOPs}\downarrow$} & 564               & \textbf{430}   & 811            & \multicolumn{1}{c|}{1903}           & 706               & \textbf{575}   & 1053           & \multicolumn{1}{c|}{1763}           & \textbf{28}       & 52             & 72             & 123            \\
\multicolumn{1}{c|}{}                                  & \multicolumn{1}{c|}{$\mathbf{Time}\downarrow$}   & \textbf{12'}      & 22'            & 25'            & \multicolumn{1}{c|}{30'}            & \textbf{46'}      & 75'            & 75'            & \multicolumn{1}{c|}{75'}            & \textbf{10'}      & 31'            & 33'            & 34'            \\ \hline
\multicolumn{1}{c|}{\multirow{6}{*}{\textit{TaT}}}     & \multicolumn{1}{c|}{$\mathbf{PSNR}\uparrow$}   & 28.93             & 29.14          & 29.27          & \multicolumn{1}{c|}{\textbf{29.37}} & 28.44             & 28.78          & \textbf{29.14} & \multicolumn{1}{c|}{29.11}          & 29.07             & 29.16          & 29.32          & \textbf{29.38} \\
\multicolumn{1}{c|}{}                                  & \multicolumn{1}{c|}{$\mathbf{SSIM}\uparrow$}   & 0.927             & 0.925          & 0.929          & \multicolumn{1}{c|}{\textbf{0.932}} & 0.905             & 0.924          & \textbf{0.929} & \multicolumn{1}{c|}{0.928}          & 0.924             & 0.927          & 0.929          & \textbf{0.930} \\
\multicolumn{1}{c|}{}                                  & \multicolumn{1}{c|}{$\mathbf{LPIPS}\downarrow$}  & 0.107             & 0.108          & 0.105          & \multicolumn{1}{c|}{\textbf{0.103}} & 0                 & 0.106          & \textbf{0.099} & \multicolumn{1}{c|}{0.109}          & \textbf{0.101}    & 0.125          & 0.124          & 0.121          \\
\multicolumn{1}{c|}{}                                  & \multicolumn{1}{c|}{$\|w\|_0\downarrow$}    & 74                & \textbf{16}    & 26             & \multicolumn{1}{c|}{65}             & \textbf{7}        & 9              & 15             & \multicolumn{1}{c|}{80}             & 88                & \textbf{37}    & 47             & 70             \\
\multicolumn{1}{c|}{}                                  & \multicolumn{1}{c|}{$\mathbf{GFLOPs}\downarrow$} & 2666              & \textbf{1626}  & 3066           & \multicolumn{1}{c|}{7198}           & 3567              & \textbf{2791}  & 5126           & \multicolumn{1}{c|}{9146}           & \textbf{211}      & 229            & 531            & 1003           \\
\multicolumn{1}{c|}{}                                  & \multicolumn{1}{c|}{$\mathbf{Time}\downarrow$}   & \textbf{22'}      & 25'            & 28'            & \multicolumn{1}{c|}{38'}            & 72'               & \textbf{70'}   & 78'            & \multicolumn{1}{c|}{101'}           & \textbf{14'}      & 37'            & 39'            & 45'            \\ \hline
\multicolumn{1}{c|}{\multirow{6}{*}{\textit{LLFF}}}    & \multicolumn{1}{c|}{$\mathbf{PSNR}\uparrow$}   & 26.24             & 26.43          & 26.62          & \multicolumn{1}{c|}{\textbf{26.65}} & 26.71             & 26.73          & 27.09          & \multicolumn{1}{c|}{\textbf{27.10}} & 24.97             & 24.90          & \textbf{25.17} & 25.19          \\
\multicolumn{1}{c|}{}                                  & \multicolumn{1}{c|}{$\mathbf{SSIM}\uparrow$}   & 0.831             & 0.832          & 0.839          & \multicolumn{1}{c|}{\textbf{0.843}} & 0.835             & 0.836          & 0.862          & \multicolumn{1}{c|}{\textbf{0.864}} & 0.764             & 0.763          & \textbf{0.777} & 0.778          \\
\multicolumn{1}{c|}{}                                  & \multicolumn{1}{c|}{$\mathbf{LPIPS}\downarrow$}  & 0.136             & 0.115          & 0.111          & \multicolumn{1}{c|}{\textbf{0.107}} & 0.114             & 0.111          & \textbf{0.101} & \multicolumn{1}{c|}{\textbf{0.101}} & \textbf{0.128}    & 0.239          & 0.237          & 0.234          \\
\multicolumn{1}{c|}{}                                  & \multicolumn{1}{c|}{$\|w\|_0\downarrow$} & 62                & \textbf{26}    & 40             & \multicolumn{1}{c|}{113}            & 19                & \textbf{11}    & 16             & \multicolumn{1}{c|}{23}             & 152               & \textbf{68}    & 75             & 148            \\
\multicolumn{1}{c|}{}                                  & \multicolumn{1}{c|}{$\mathbf{GFLOPs}\downarrow$} & 1678              & \textbf{1514}  & 2508           & \multicolumn{1}{c|}{4972}           & 4542              & \textbf{3226}  & 5921           & \multicolumn{1}{c|}{13522}          & 573               & \textbf{887}   & 1391           & 2712           \\
\multicolumn{1}{c|}{}                                  & \multicolumn{1}{c|}{$\mathbf{Time}\downarrow$}   & \textbf{24'}      & 28'            & 32'            & \multicolumn{1}{c|}{36'}            & 58'               & \textbf{49'}   & 57'            & \multicolumn{1}{c|}{68'}            & \textbf{24'}      & 25'            & 46'            & 42'            \\ \hline
\end{tabular}%
}
\label{tab:performance}
\end{table}

\subsection{Metrics and Datasets}
For each test, image-quality metrics such as PSNR, SSIM~\cite{wang2004image}, and LPIPS~\cite{zhang2018unreasonable} (computed on AlexNet~\cite{krizhevsky2012imagenet}) are presented. Additionally, we report the number of non-zero parameters as $\|w_0\|$, the average GFLOPs required by each model to render the images of the test set, and total training times.  We present results obtained across four major datasets, namely: Synthetic-NeRF~\cite{mildenhall2020nerf}, Neural Sparse Voxel Field Dataset (NSVF)~\cite{liu2020neural}, TanksAndTemple~\cite{knapitsch2017tanks}, and Local Light Field Fusion Dataset (LLFF)~\cite{mildenhall2019local}.

\subsection{Quantitative Results}
\begin{figure}[t]
    \centering
    \begin{subfigure}[b]{0.32\textwidth}
        \includegraphics[width=\textwidth, height=3.15cm]{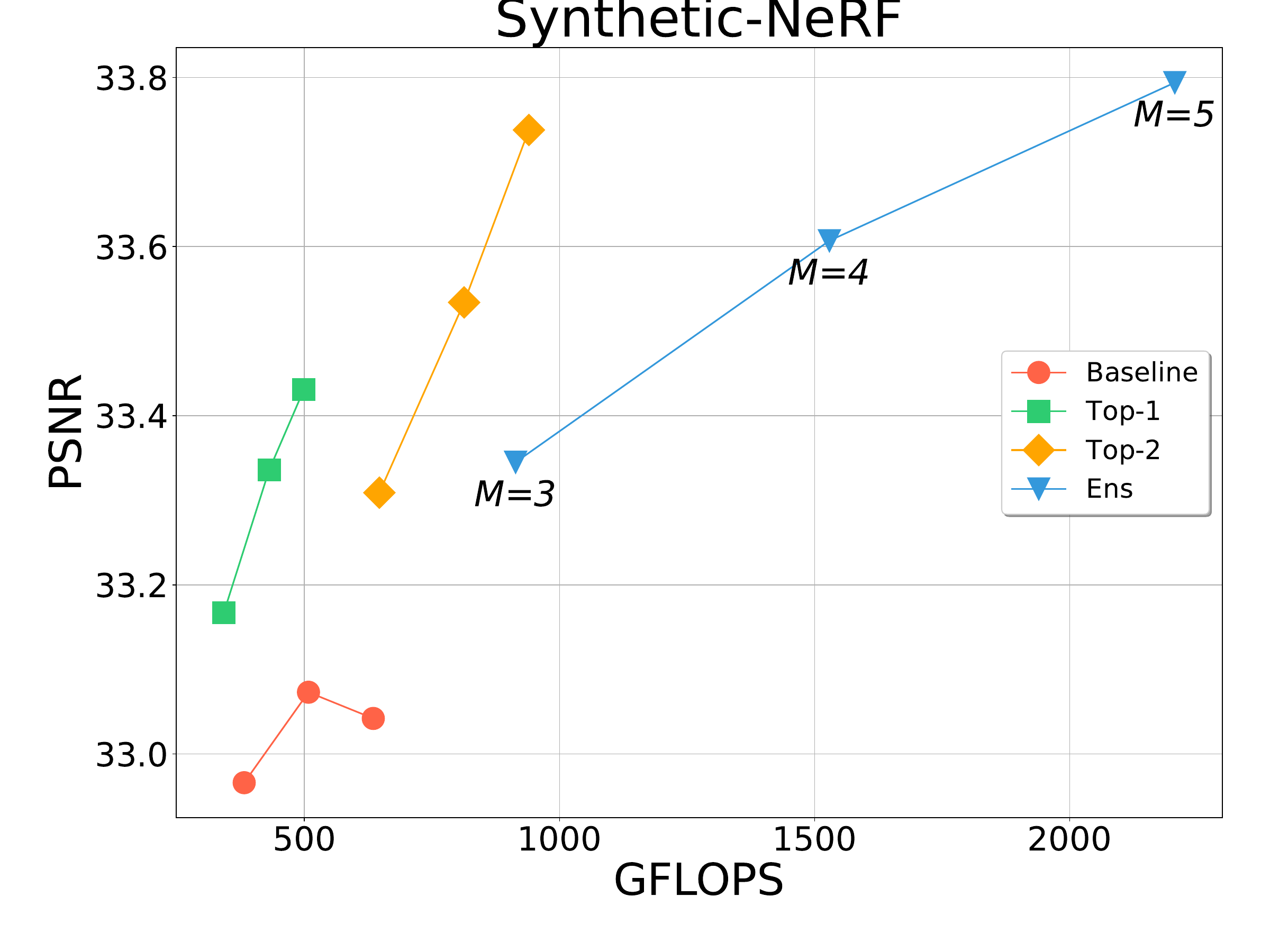}
   
    \end{subfigure}
    \hfill
    \begin{subfigure}[b]{0.32\textwidth}
        \includegraphics[width=\textwidth, height=3.15cm]{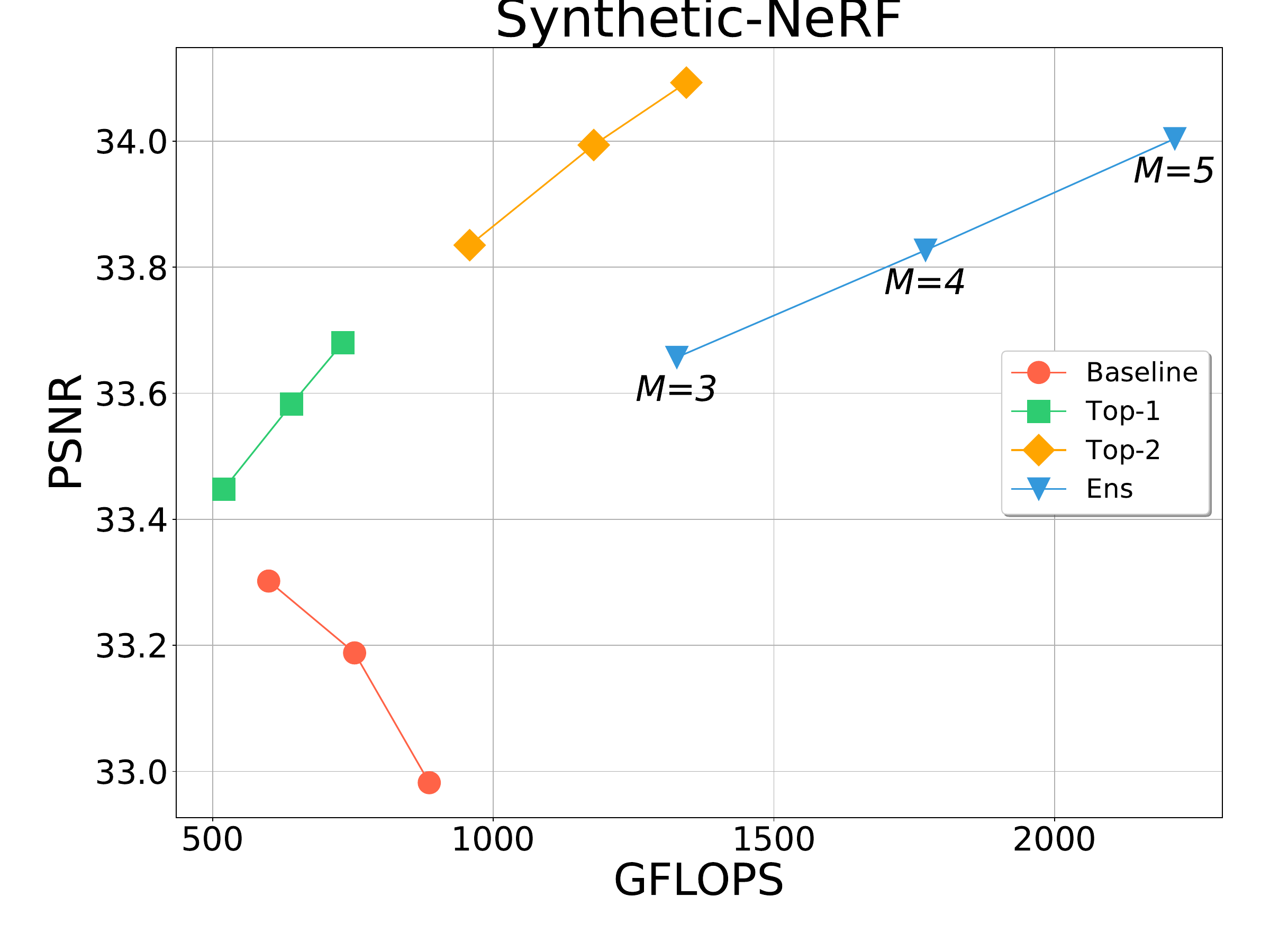}

    \end{subfigure}
    \hfill
    \begin{subfigure}[b]{0.32\textwidth}
        \includegraphics[width=\textwidth, height=3.15cm]{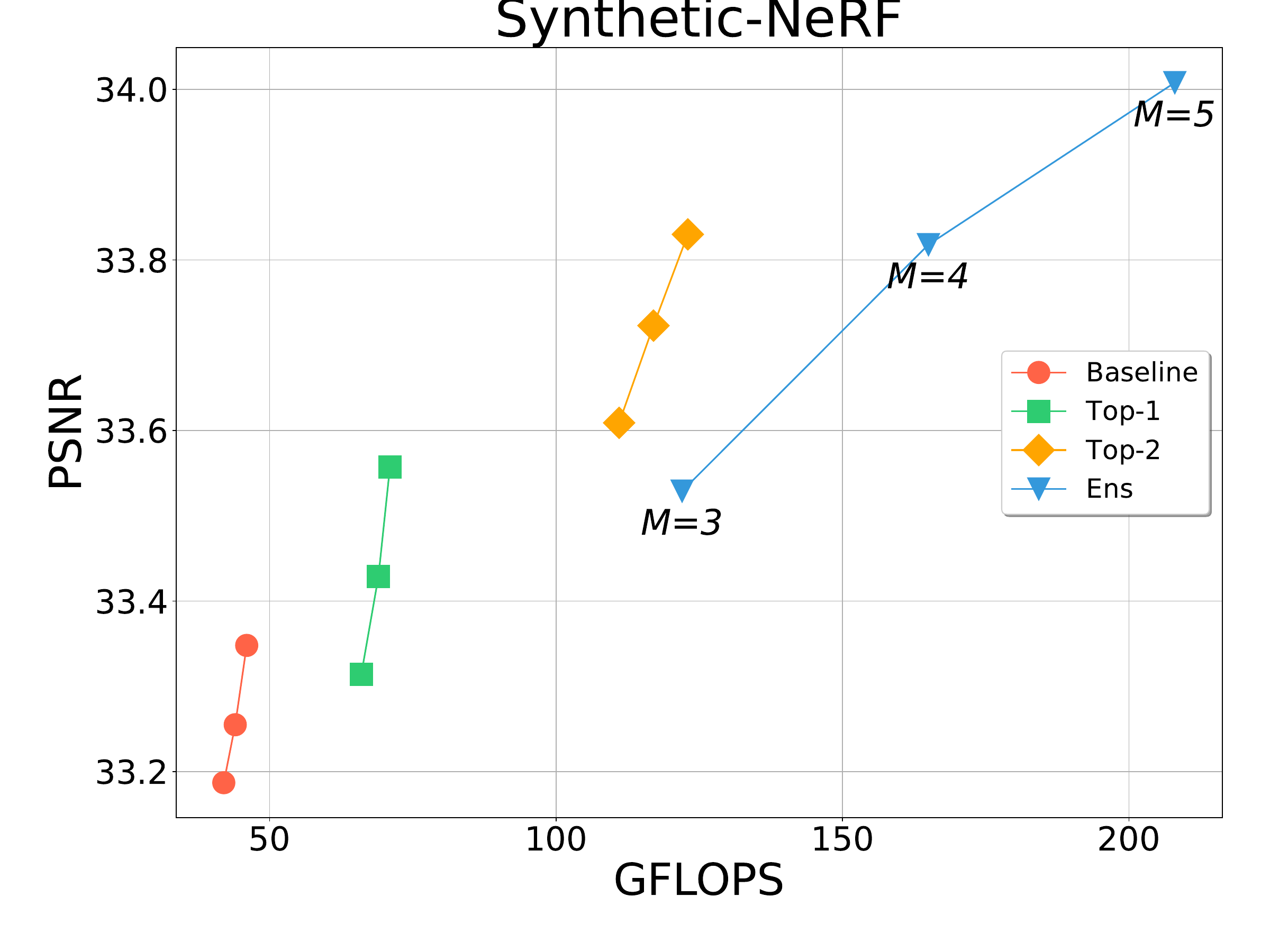}

    \end{subfigure}
    \hfill

    \medskip

        \begin{subfigure}[b]{0.32\textwidth}
        \includegraphics[width=\textwidth, height=3.15cm]{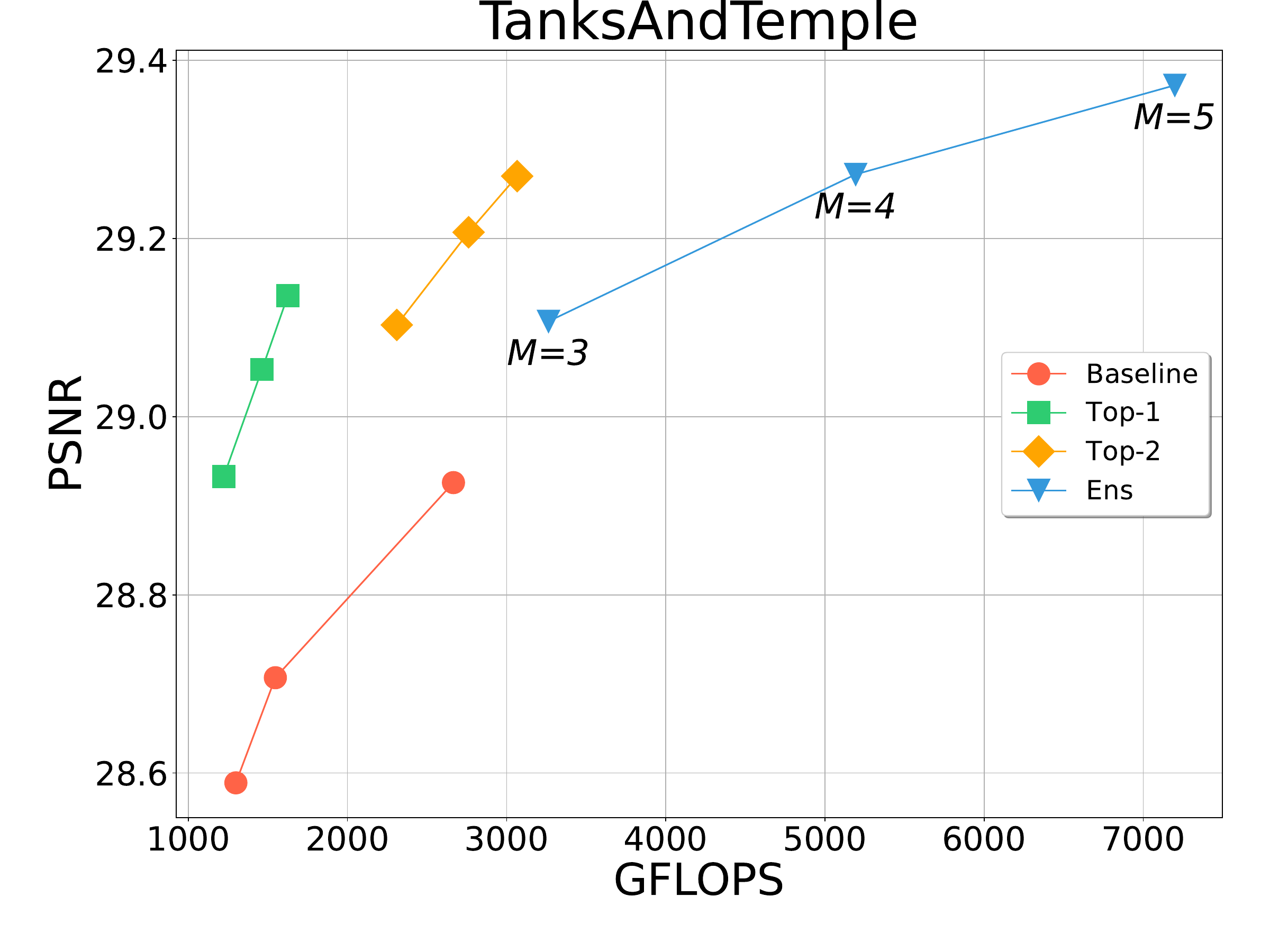}
        \caption{DVGO}
    \end{subfigure}
    \hfill
    \begin{subfigure}[b]{0.32\textwidth}
        \includegraphics[width=\textwidth, height=3.15cm]{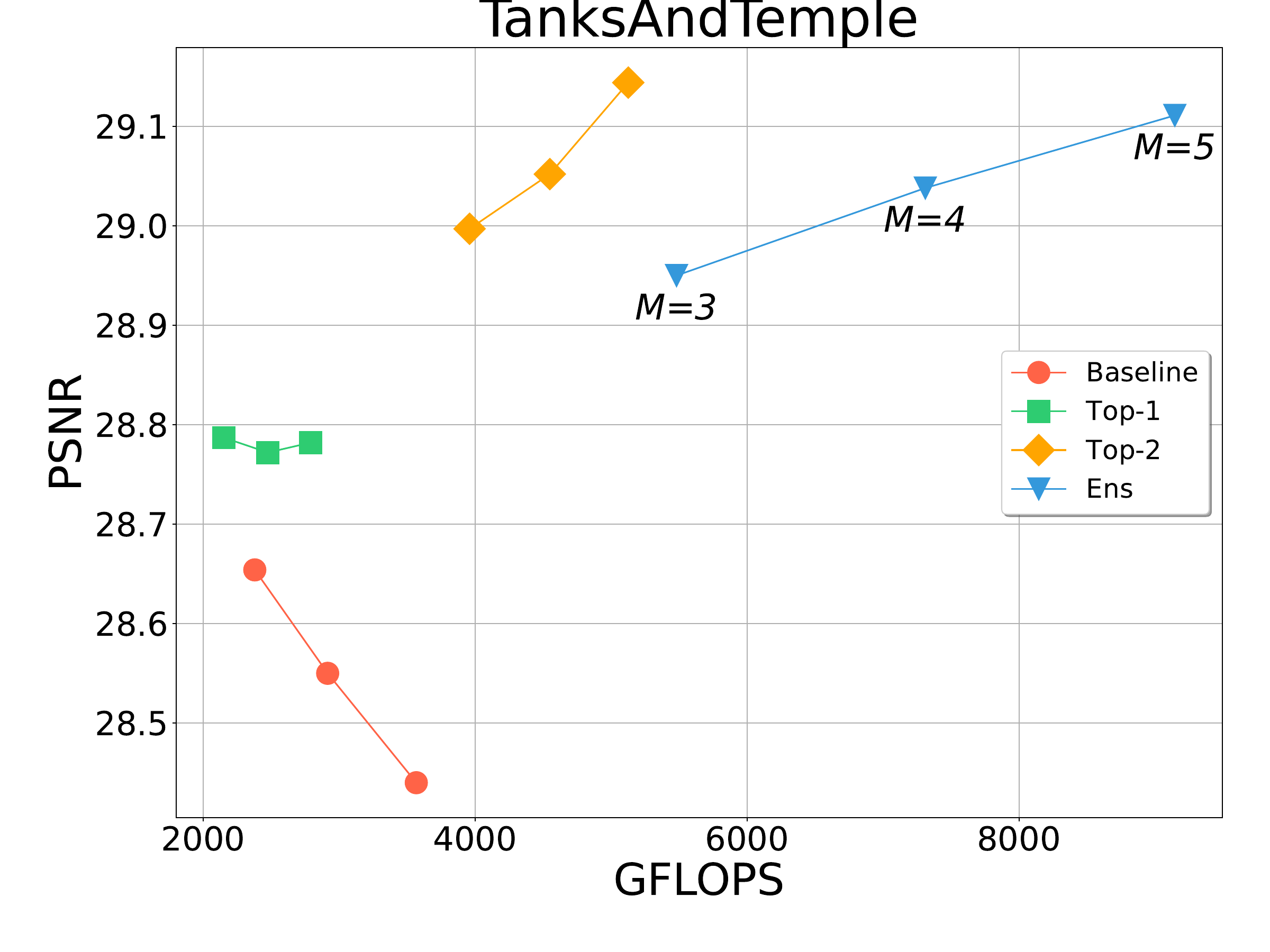}
        \caption{TensoRF}
    \end{subfigure}
    \hfill
    \begin{subfigure}[b]{0.32\textwidth}
        \includegraphics[width=\textwidth, height=3.15cm]{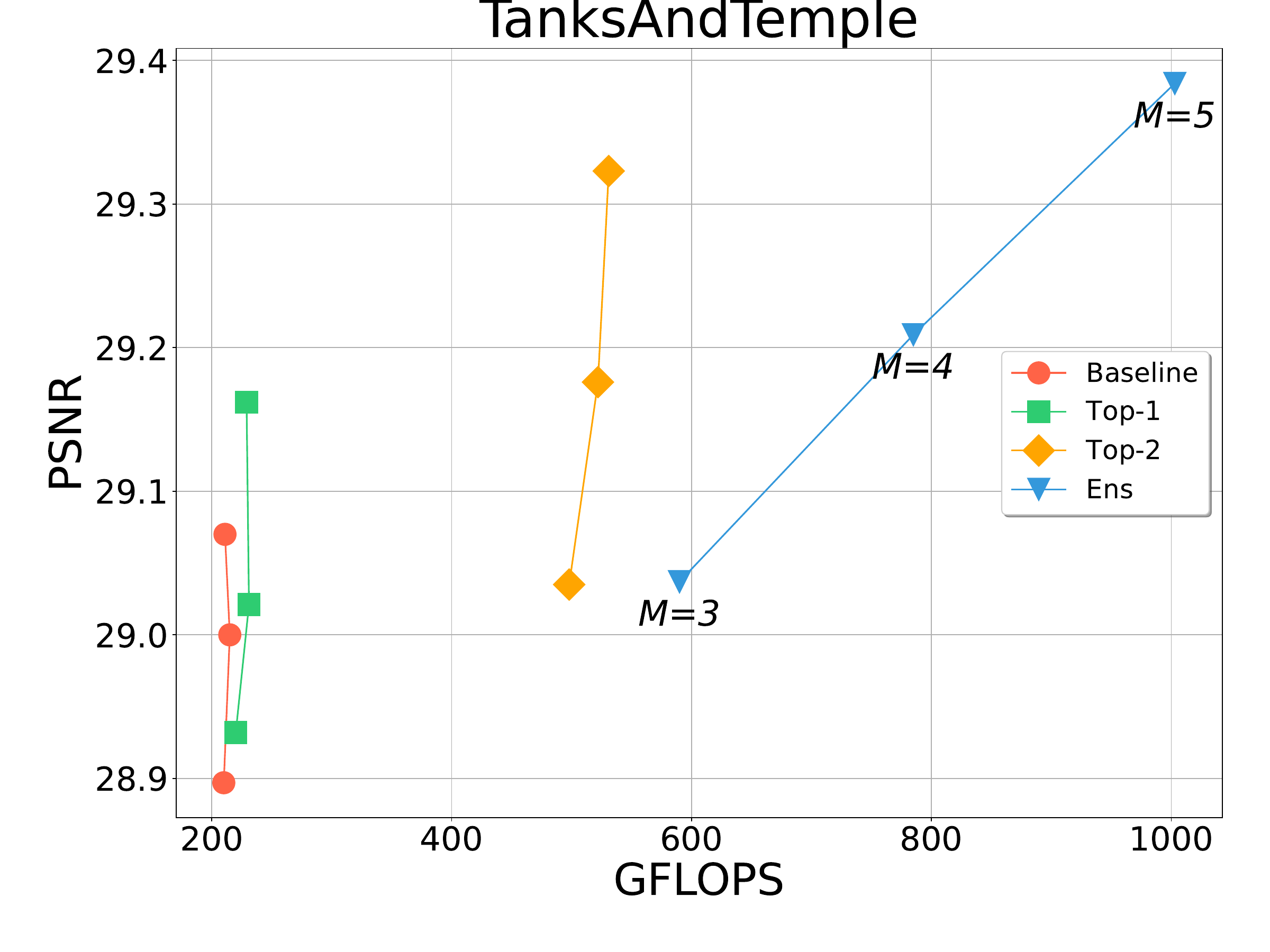}
        \caption{Instant-NGP}
    \end{subfigure}
    \hfill

    \caption{PSNR/GFLOPs plots for Synthetic-NeRF and TanksAndTemple. The remaining datasets show similar results.}
    \label{fig:performance}

\end{figure}

The main experimental results in terms of the metrics defined above are shown in Table~\ref{tab:performance}. Here, we present the results obtained using a mixture of $M=5$ experts; further configurations can be found in the supplementary Sec. \ref{sec:add-results}.
Our experiments reveal that our MoE provides a significant rendering quality improvement with respect to the baseline with no or limited impact in terms of computational cost. The increase in rendering quality is notable in Synthetic NeRF (up to 1 dB) and NSVF (up to 1.3 dB). 
Although more moderate, improvements are still evident, even in more challenging datasets such as TanksAndTemple and LLFF, with about 0.5 dB gain. 
While Top-$1$ can already achieve state-of-the-art render quality, the Top-$2$ strategy further enhances image quality, reaching levels comparable to the ensemble but with much greater efficiency (about half the average FLOPs per rendering) and using significantly fewer parameters. This observation is also illustrated in Figure \ref{fig:performance} 
, where FLOPS/PSNR plots are shown (each marker in every curve refers to the cases with $M=3,4,5$ experts respectively).
Training times are on average are longer but still acceptable (in the worst-case scenario, around 1h is required to train our MoE). 
\fra{However, in the case of Top-1, they can be faster than baselines. This is because of our resolution methods, which tend to favor low-resolution models.}
Based on these analyses, the Top-$2$ strategy strikes an excellent quality/cost trade-off. The ensembling configuration, using all experts, represents an upper bound in terms of image quality while being the worst case in computational terms.

\subsection{Qualitative Results}
\begin{figure}[h]
\centering
\includegraphics[width=0.8\textwidth]{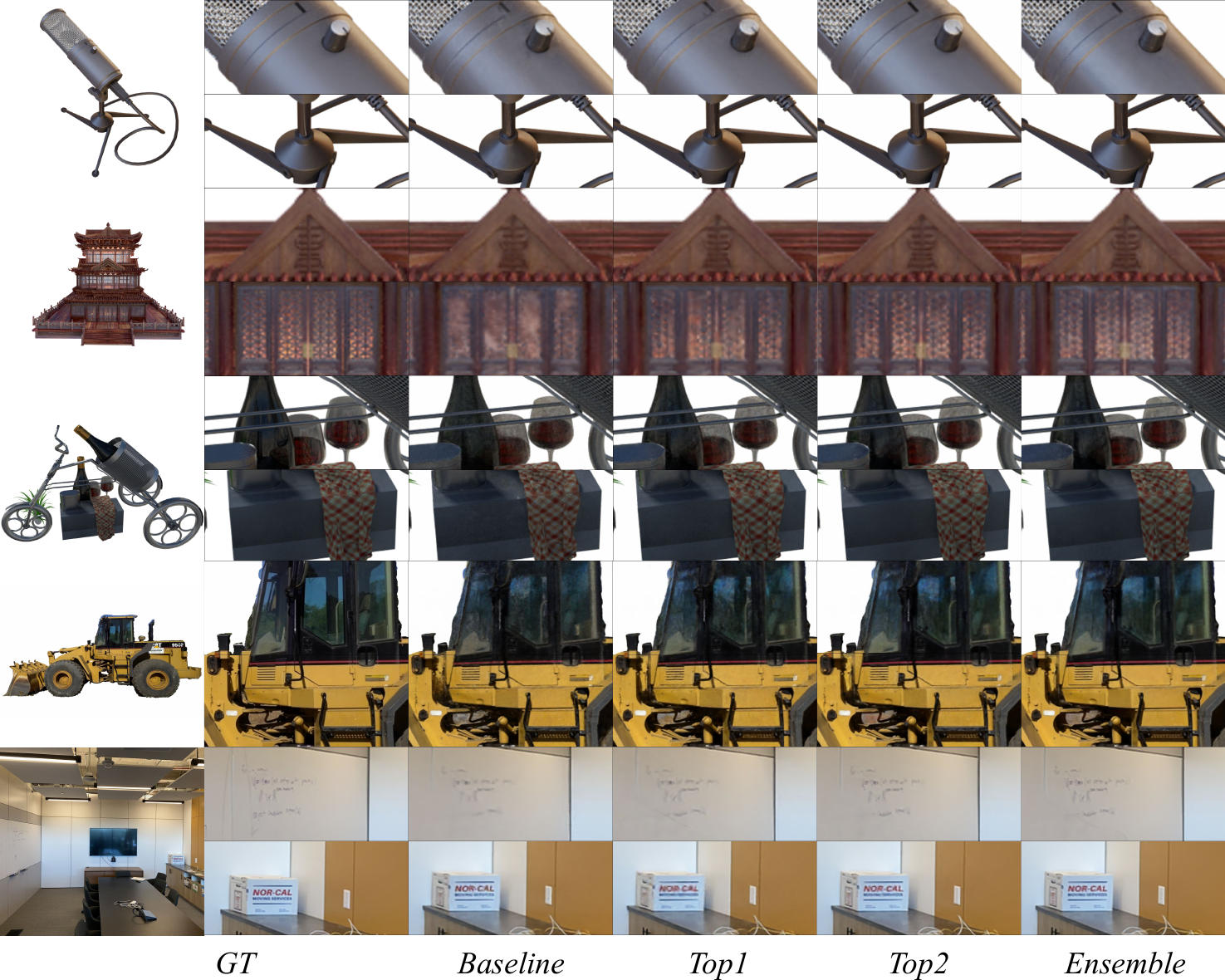}
\caption{Qualitative results on some of the scenes of each dataset. From left to right: ground truths, baselines, Top-1, Top-2 and Ensemble. Each model has the same parameters and has been trained for the same number of iterations (DVGO).}
\label{fig:qualitative}
\end{figure}

Figure \ref{fig:qualitative} shows visual comparisons among the baseline, ensemble, Top-$1$, and Top-$2$.
From the selected image crops, one can appreciate that our method ensures superior reconstruction quality compared to the baselines, effectively reproducing sharper details while reducing noise on texture-less spots. This is particularly notable when examining elements such as the window decorations in the Palace scene or the text on the box above the desk in the Room scene. Additionally, surfaces such as the semi-transparent glass in the Wineholder or the Caterpillar scene appear less noisy and more faithful to the original. This improvement can be attributed to the synergy among different resolution models: low-resolution models excel at representing lower frequencies, thereby introducing less noise, while high-resolution models can focus solely on high-frequency components.
Additionally, it is important to notice that there is no significant difference between the ensemble and Top-$2$.

\fra{
\subsection{Comparison with Naive Methods}
As shown in Figure~\ref{fig:fig1}, our method leads to significantly higher quality reconstructions at a greatly reduced computational cost. Decreasing the step size yields the least noticeable improvement while incurring a substantial computational expense, as each sampled point requires an evaluation by the color decoder. Our method achieves high-quality reconstructions with the same step size as the baseline models. Increasing the MLP parameters can boost accuracy, but again, the computational costs become substantial compared to our method. Similarly, increasing resolution results in way inferior PSNR with respect to our method at a comparable computational cost. Our method also allows scaling to higher resolutions, while baseline models tend to introduce noise and artifacts, leading to a decrease in reconstruction quality as the resolution increases.
}
\subsection{Gate Visualization and Scene Decomposition}
\begin{figure}[h]
\centering
\includegraphics[width=0.8\textwidth]{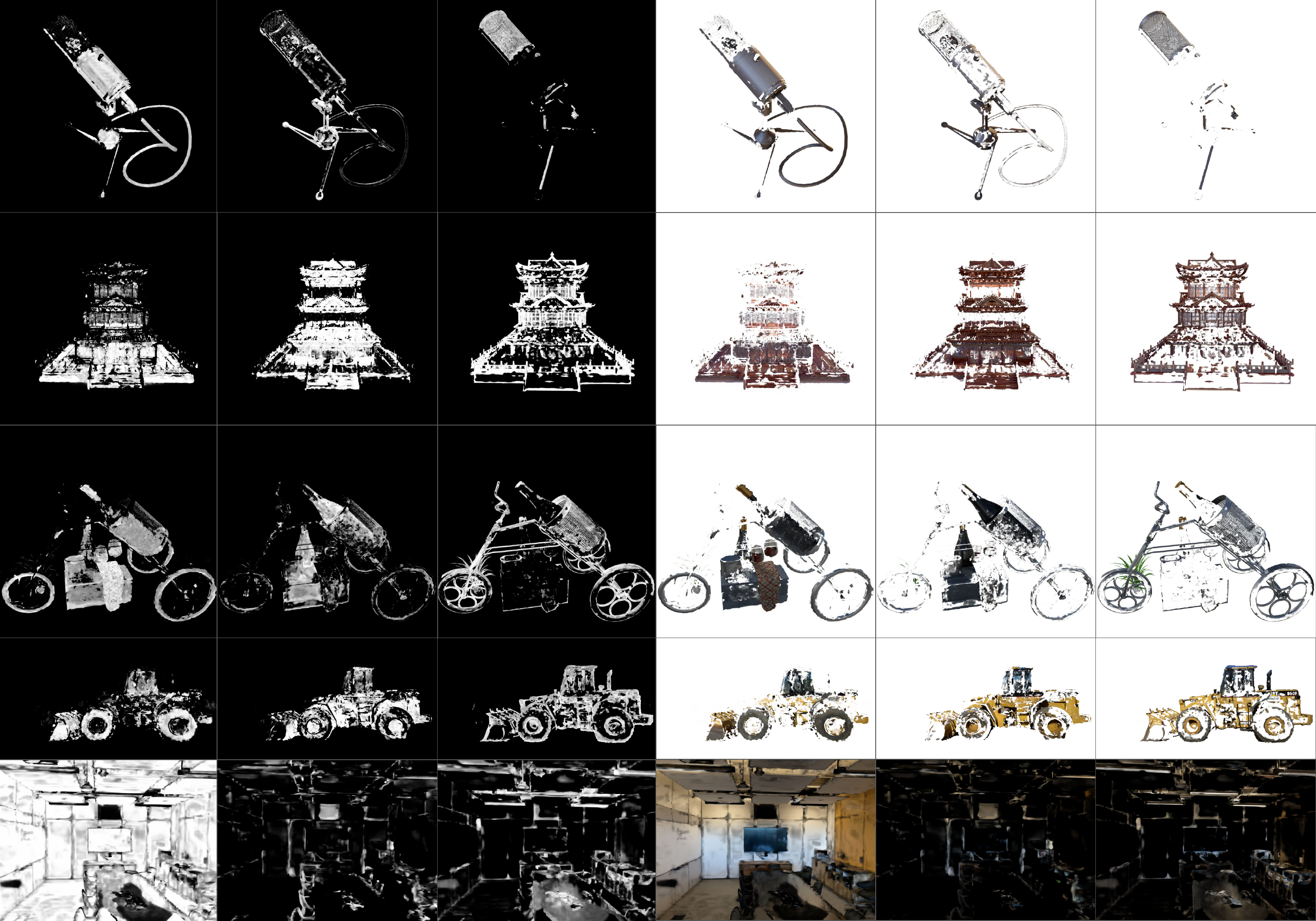}
\caption{Gate (gray-scale images) and per-model output visualizations with 3 experts and resolution-based routing (DVGO). Images are ordered by increasing resolutions.}
\label{fig:gate}
\end{figure}

In Figure~\ref{fig:gate}, we visualize the gate (probability field in grayscale) and each expert output in the case $M=3$.
On the right side of the figure, one can appreciate per-model renders and experts' specialization. It is worth noting how higher-resolution experts render high-frequency details. This is particularly evident in the \textit{Mic} scene.

\subsection{Ablation}

\begin{figure*}[h]
  \centering
  \begin{minipage}[b]{0.4\textwidth}
    \centering
    \resizebox{\textwidth}{!}{%
\begin{tabular}{ccccc}
\hline
\textbf{Res} & \textbf{Gate Type}   & $\mathbf{PSNR}$  & $\|w\|_0$ & $\mathbf{GFLOPs}$ \\ \hline
$128^3$ & Ours        & 36.80  & 26 & 476    \\
$256^3$ & Ours        & 36.77 & 33 & 601    \\
$300^3$ & Ours        & 36.79 & 42 & 677    \\
-   & Linear      & 36.25 & 24  & 444    \\
-   & Switch-NeRF & 36.44 & 25  & 1095   \\ \hline
\end{tabular}%
}
   \captionof{table}{Comparison of different gate resolutions and formulations on Lego scene and DVGO.\label{tab:abl_gate_resolutions}}
  \end{minipage}
    \hfill
  \begin{minipage}[b]{0.55\textwidth}
    \centering
    \resizebox{\textwidth}{!}{%
        \includegraphics{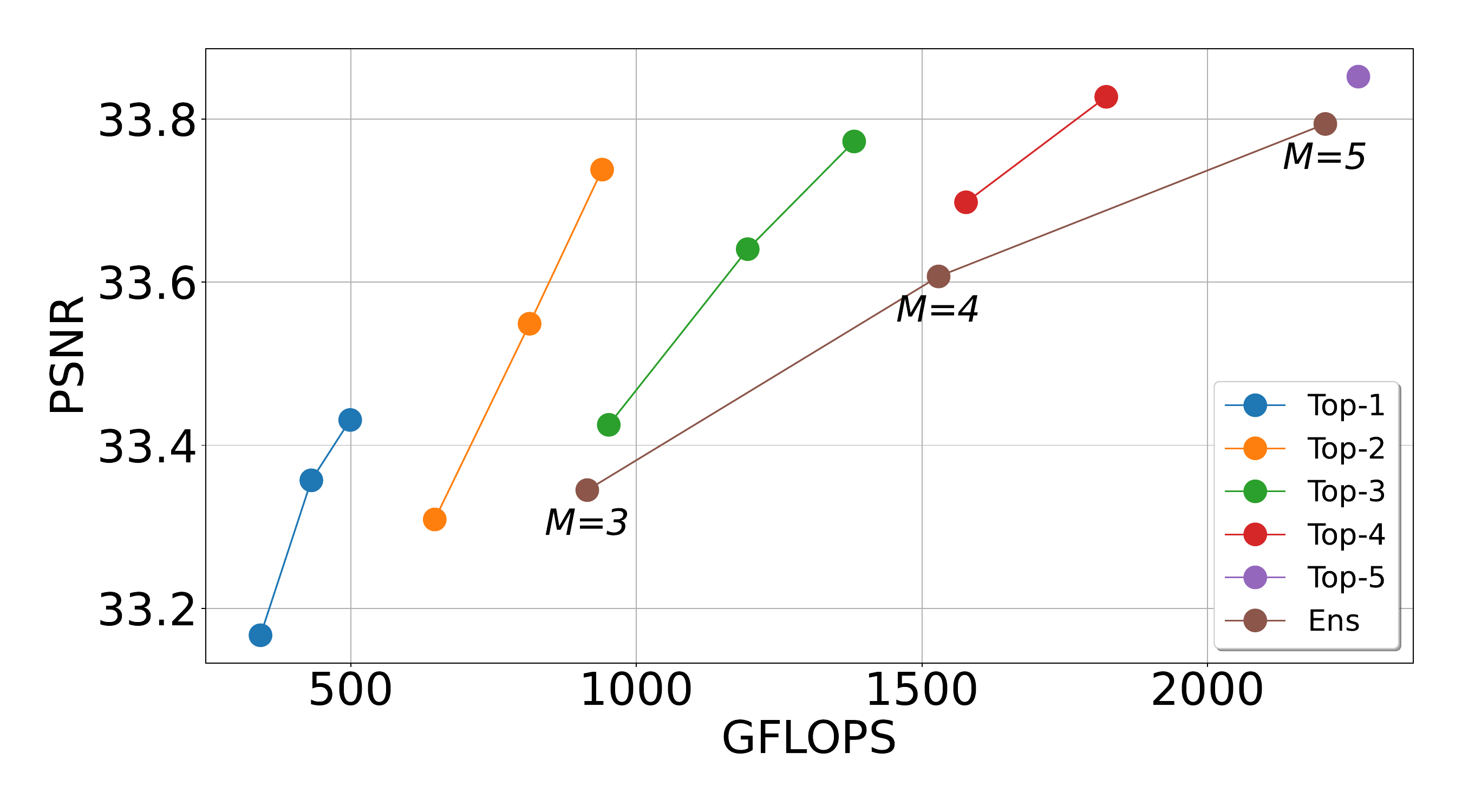}
    }
    \caption{Comparison of Gating Functions (Top-1 to Top-5) on Synthetic-NeRF with DVGO.\label{fig:abl_topk_comparisons}}
  \end{minipage}
\end{figure*}
Here we investigate how different design choices of our model affect the performance and rendering quality. These choices include the gate resolution, different gate formulations, and the number of experts in the Top-$k$.

Here we investigate how different design choices of our model affect the performance and rendering quality. These choices include the gate resolution, different gate formulations, and the number of experts in the Top-$k$.

\subsubsection{Gate Resolution and Gate Formulation.}
We show in Table~\ref{tab:abl_gate_resolutions} that a low-resolution gate is sufficient to achieve high-quality rendering. Interestingly, as the gate resolution increases, there is a slight decrease in rendering quality during testing, coupled with an increase in required FLOPs per image rendering and number of parameters.
We also compare various gating strategies, including linear gating, the configuration proposed by Switch-NeRF, and our gate formulation. Through experimentation, it becomes evident that our approach outperforms others in terms of both render quality and performance, achieving performances comparable to a linear gating function.

\subsubsection{Why Top-2?}
Here we investigate the performance and rendering quality trends with varying values of $k$. In Figure~\ref{fig:abl_topk_comparisons}, we can see that while $k = 1$ can lead to a significant increase in quality with comparable performance to baseline models, the Top-2 further enhances quality at the expense of increased (but still acceptable) computation. The Top-3, Top-4, and Top-5 provide marginal quality improvements at a significantly higher computational cost. Hence, we consider the Top-2 the optimal balance between performance and quality.

\section{Limitations}
\fra{ Our study present some limitations. First of all, a pre-training phase, on which each model is trained \textit{independently} is required for achieving good results. Training end-to-end, without a pre-training phase, can lead to reconstruction quality that is noticeably inferior to baselines. Pre-training models at different resolutions allows for diversified architectures, making it easier for the gate to learn the decomposition.
The training is also strongly dependent on the auxiliary loss. Different values of $\lambda$ can significantly influence performances and load balancing. We conducted a sweep to identify a value that works well across many scenarios, but it may not be suitable for different datasets.  
Another limitation is the overhead introduced by the MoE. Each input token is first interpolated with the gate's grid, decoded into probabilities and routed to the chosen expert. These operations can significantly slow-down the rendering process, leading to higher training and inference times. Although considerable effort was devoted to developing the most efficient gate possible, our MoE is still slower than the respective baselines. }

\section{Conclusions}
In this paper, we introduced a model-agnostic framework for enhancing the rendering of Fast-NeRF models. Our formulation of the Gate reduces computational costs in both training and inference phases while ensuring better quality compared to a traditional gate. Additionally, the introduction of an auxiliary loss with res penalty allows for increased utilization of low-resolution models, reducing the number of active parameters and promoting sparsity in higher-resolution models.
Our results demonstrate how this approach can significantly improve reconstruction quality while considering performance metrics. Specifically, we show that a Top-2 strategy strikes a good balance between performance and quality.

\section*{Acknowledgements}
This work was partially funded by Hi!PARIS Center on Data Analytics and Artificial Intelligence.

\newpage
\section{Additional Details}

Our MoE is reducible to a Sparsely-Gated Mixture of Experts with full capacity. Since the number of points sampled is reasonable, we did not find necessary to impose a limit to the capacity factor of each model. We also introduce different-resolution experts in the MoE, promoting the routing of tokens towards lower-resolution experts.
Regarding DVGO and TensoRF, we utilized models at various grid resolutions, while keeping all other hyperparameters consistent with the original implementations. The selected resolutions are $160^3, 200^3, 256^3, 320^3,$ and $384^3$.
For Instant-NGP, we trained models while varying the $L$ parameter across the following values: $6, 8, 10, 16,$ and $24$.
Experts within the MoE are arranged in increasing resolution order.

\subsection{$\|w\|_0$ Computation}
We define $\|w\|_0$ as the count of non-null parameters in the Mixture of Experts. As our MoE operates as a Sparsely-Gated MoE, only a subset of the Neural Radiance Fields models are effectively employed during rendering. Moreover, our resolution-based routing mechanism has the potential to significantly reduce this parameter.

To compute the parameters, we freeze the MoE and render all the images in the test set. Subsequently, we calculate the loss in the usual manner and compute gradients with respect to the MoE's parameters. $||w||_0$ then represents the count of parameters with non-null gradients:
\begin{equation}
    \|w\|_0 := \|\frac{\partial L_{\text{nerf}}}{\partial W_{\text{MoE}} \neq 0 } \|_0
\end{equation}
This can be associated with gradient-magnitude pruning. We leave this idea as a future project. 

\subsection{Instant-NGP Implementation}
The implementation for Instant-NGP closely resembles that proposed in Figure \ref{fig:method}, albeit with a slight modification: the gate comprises a tiny Instant-NGP model. This gate model consists of a multi-resolution hash grid with $L=6$, which computes per-point features. These features are then linearly interpolated with the 8 nearest voxels, similar to DVGO and TensoRF, before being concatenated together. Following this concatenation, they undergo transformation into probabilities using the same methodology as in the main method. Additionally, the MLP remains shallow, comprising 2 layers with 64 neurons each, activated by ReLU functions.
\subsection{Ensembling NeRFs' Outputs}
\label{sec:theory}
\begin{figure}[]
\centering
\includegraphics[width=0.5\textwidth]{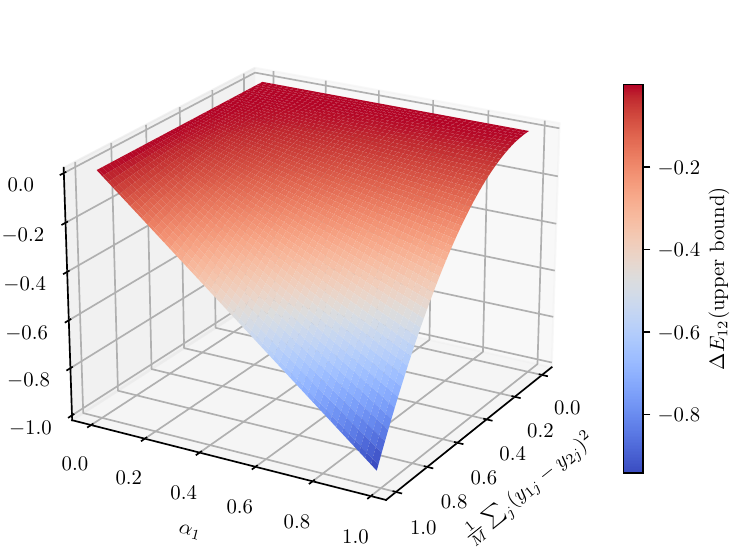}
  \caption{Upper bound for $\Delta E_{12}$ such that \eqref{eq:cond3} is satisfied.}
  \label{fig:boundary}
\end{figure}
\begin{figure}[]
\centering
\includegraphics[width=0.5\textwidth]{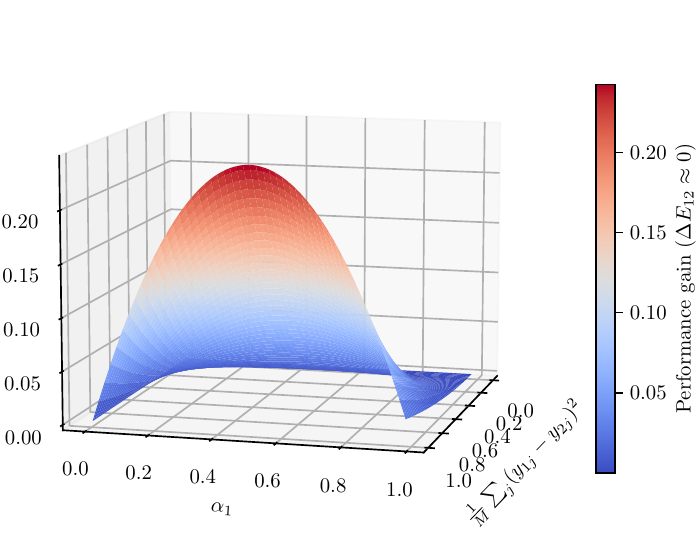}
  \caption{Estimated performance gain as a function of $\alpha_1$ and of the difference in output between the two NeRFs.}
  \label{fig:aetalpha}
\end{figure}

To validate our Mixture of Experts, we conducted a comparison with an ensemble of Neural Radiance Fields, where the output is determined by averaging the predictions of each NeRF model. Starting from a set of $M$ pre-trained models, an ensemble is built and is then fine-tuned.
Averaging the outputs of NeRFs can notably enhance the quality of reconstruction.
Let us assume we have two NeRFs, having as output respectively $\boldsymbol{y}_{1}$ and $\boldsymbol{y}_{2}$, while the ground truth is $\hat{\boldsymbol{y}}$. We can write an equality of their performance, in terms of mean-squared error:
\begin{align}
    E_1 &= \Delta E_{12} + E_2\nonumber\\
    \label{eq:cond1}
    \frac{1}{M}\sum_j (y_{1j}-\hat{y})^2 &= \Delta E_{12} + \frac{1}{M}\sum_j (y_{2j}-\hat{y})^2
\end{align}
where $\Delta E_{12}$ is the error gap between the two models and $M$ is the number of outputs on which the error is averaged. From \eqref{eq:cond1} we can easily write
\begin{equation}
    \frac{1}{M}\sum_j (y_{2j}^2 - y_{1j}^2)= \Delta E_{12} + \frac{2}{M}\sum_j\hat{y}(y_{2j}-y_{1j}).
\end{equation}
Through ensembling, we desire the error $E_{\text{ens}}$ to be lower than either of the two other models. Without loss of generality, let us consider the case such that it is lower than the error of the first model:
\begin{align}
    E^{\text{ens}} &< E_1 \nonumber\\
    \frac{1}{M}\sum_j (\alpha_1 y_{1j} + \alpha_2 y_{2j} - \hat{y})^2 &< \frac{1}{M}\sum_j (y_{1j} - \hat{y})^2\label{eq:cond2}
\end{align}
where $\alpha_1,\!\alpha_2$ are two weighting factors for the two NeRF's outputs. If we expand \eqref{eq:cond2} we obtain
\begin{align}
    \frac{1}{M}\sum_j (\alpha_1 y_{1j} + &\alpha_2 y_{2j})^2 - y_{1j}^2 < \nonumber\\
    &\frac{2}{M}\sum_j \hat{y}\left[ \alpha_2 y_{2j} - (1-\alpha_1) y_{1j} \right ].\label{eq:int1}
\end{align}
In order to plug \eqref{eq:cond1} in \eqref{eq:int1}, we need to impose $\alpha_2 = 1-\alpha_1$. Hence, we have
\begin{align}
    \frac{1}{M}\sum_j &\left [ \alpha_1 y_{1j} + (1-\alpha_1) y_{2j}\right ]^2 - y_{1j}^2 < \nonumber\\
    &(1-\alpha_1)\left [ \Delta E_{12} + \frac{1}{M}\sum_j (y_{2j}-\hat{y})^2 \right ]\label{eq:int2}.
\end{align}
By expanding and simplifying \eqref{eq:int2}, we obtain
\begin{equation}
    \frac{1}{M}\sum_j (\alpha_1^2-\alpha_1)(y_{1j} - y_{2j})^2 < (1-\alpha_1)\Delta E_{12}.\label{eq:cond3}
\end{equation}
Fig.~\ref{fig:boundary} pictures the upper bound dictated by $\Delta E_{12}$ such that \eqref{eq:cond3} is verified. Given that for \eqref{eq:cond1} we never stated which of the two models has the lowest error, evidently if $\Delta E_{12}>0$, for $\alpha_1<1$ \eqref{eq:cond3} is always verified. However, this is not anymore the case when $\Delta E_{12}<0$. In order to maximize the performance of the ensemble, we need to settle to the critical point of \eqref{eq:cond3} (global maximum) with respect to $\alpha_{1}$, which is
\begin{equation}
    \alpha_{1} = \frac{1-\Delta E_{12}}{2}.
\end{equation}
If we assume $\Delta E_{12}$ being distributed as a random variable with zero means (this is a realistic assumption as we never stated which of the two models performs the best), evidently the best solution is for $\alpha_1 = \frac{1}{2}$: Fig.~\ref{fig:aetalpha} graphically evidences this.

\section{Algorithm}
\label{sec:pseudocode}
We present an intuitive pseudo-algorithm outlining our Mixture of Experts framework. Our native implementation in PyTorch closely adheres to this algorithm.
\begin{algorithm}[t]
\caption{Sparse MoE Trainining}
\begin{algorithmic}[1]
\Require $D$: dataset, $res$: resolution, $l$: number of pre-training iterations, $m$: number of training iterations, $M$: number of experts in the mixture, $k$: number of top points to select, $\lambda$: resolution-based aux-loss penalties, $W$: aux-loss weights

\State $experts \gets $ create\_experts($res$)
\State $experts \gets$ pre\_train($experts$, $l$) 
\State $gate \gets$ create\_gate()
\State $moe \gets$ build\_moe($experts$, $gate$)

\For{$i \gets 1$ \textbf{to} $m$}
    \State $\mathbf{o}$, $\mathbf{d}$, $\mathbf{rgb} \gets$ batch($D$)
    \State $\mathbf{x} \gets$ sample\_and\_filter\_points($\mathbf{o}$, $\mathbf{d}$)
    \State $G(\mathbf{x}) \gets$ $gate(\mathbf{x})$
    \State $topk\_idx$, $topk\_vals \gets$ top-k($G(\mathbf{x}), k$)
    \State $\mathbf{\sigma_s}$, $\mathbf{c_s}$ $\gets$ array()
    \For{$j \gets 1$ \textbf{to} $M$}
        \State $mask \gets topk\_idx == j$
            \State $\mathbf{\sigma}$, $\mathbf{c} \gets$ $experts[j](\mathbf{x}[mask], \mathbf{d})$
        \State $\mathbf{\sigma_s}[j] \gets \sigma \cdot topk\_vals[mask]$
        \State ${\mathbf{c_s}[j]} \gets \mathbf{c} \cdot topk\_vals[mask]$
    \EndFor
    
    \State $\mathbf{\sigma_f} \gets$ sum($ \mathbf{\sigma}_s$)
    \State $\mathbf{c_f} \gets$ sum($\mathbf{c_s}$)
    \State $\mathbf{colors} \gets$ volume\_rendering($\mathbf{\sigma_f}$, $\mathbf{c_f}$)
    \State $L_{\text{tot}} \gets  L_{\text{nerf}}(rgb, colors) + \lambda L_{\text{rw-aux}}(P(\mathbf{x}), W)$
    \State optimize($moe$)
\EndFor
\end{algorithmic}
\end{algorithm}
The initial phase involves independently training $M$ models at different resolutions for $m$ iterations. Given that we utilize Fast NeRFs, this phase tends to be rapid.
Following this, we construct the Sparsely-Gated Mixture of Experts as outlined in lines 1-4 of the algorithm. The training of the MoE commences at line 4:
Given a batch of rays, typically represented by a triplet (origin, direction, and ground truth color), points along the ray are sampled and appropriately filtered. Subsequently, for each point, a probability distribution is computed, indicating the confidence of the gate in assigning each point to each expert (line 8). Based on this probability, the points are delivered to the top-k experts, who calculate, for each point, the density and radiance, which are then multiplied by their corresponding probability value (lines 9-16). The density and radiance values are aggregated, yielding a single array of density and radiance (line 20). Finally, we proceed as usual by computing the pixel color using the volume rendering equation, calculating the loss, and optimizing both the gate and the experts.
Our algorithm is tailored for a single GPU, but it can be readily extended to a multi-GPU environment, as the computation of the experts' output can be parallelized.

\section{Resolution Penalties}
\label{sec:penalties}
We experimented with different resolution-based penalties, namely: linear, geometric progression, and quadratic, defined as follows:
\begin{equation}
    \text{linear}:  w_{i+1} = w_i + k
\end{equation}

\begin{equation}
    \text{geom progr.}: w_i = \exp\left(\frac{\ln M}{M-1}\right)^i
\end{equation}
\begin{equation}
    \text{quadratic}: w_{i+1} = 2 \cdot w_{i}
\end{equation}
where $i \in [0,...,M-1]$ and $k = 1$.
Considering the results (Tab.~\ref{tab:abl_res}), we opted for the geometric progression as it provides slightly better results. It's worth noting how introducing such penalties further improves render quality compared to a standard auxiliary loss without resolution penalty (indicated as "none").

\begin{figure}[h]
\centering
\includegraphics[width=0.9\textwidth]{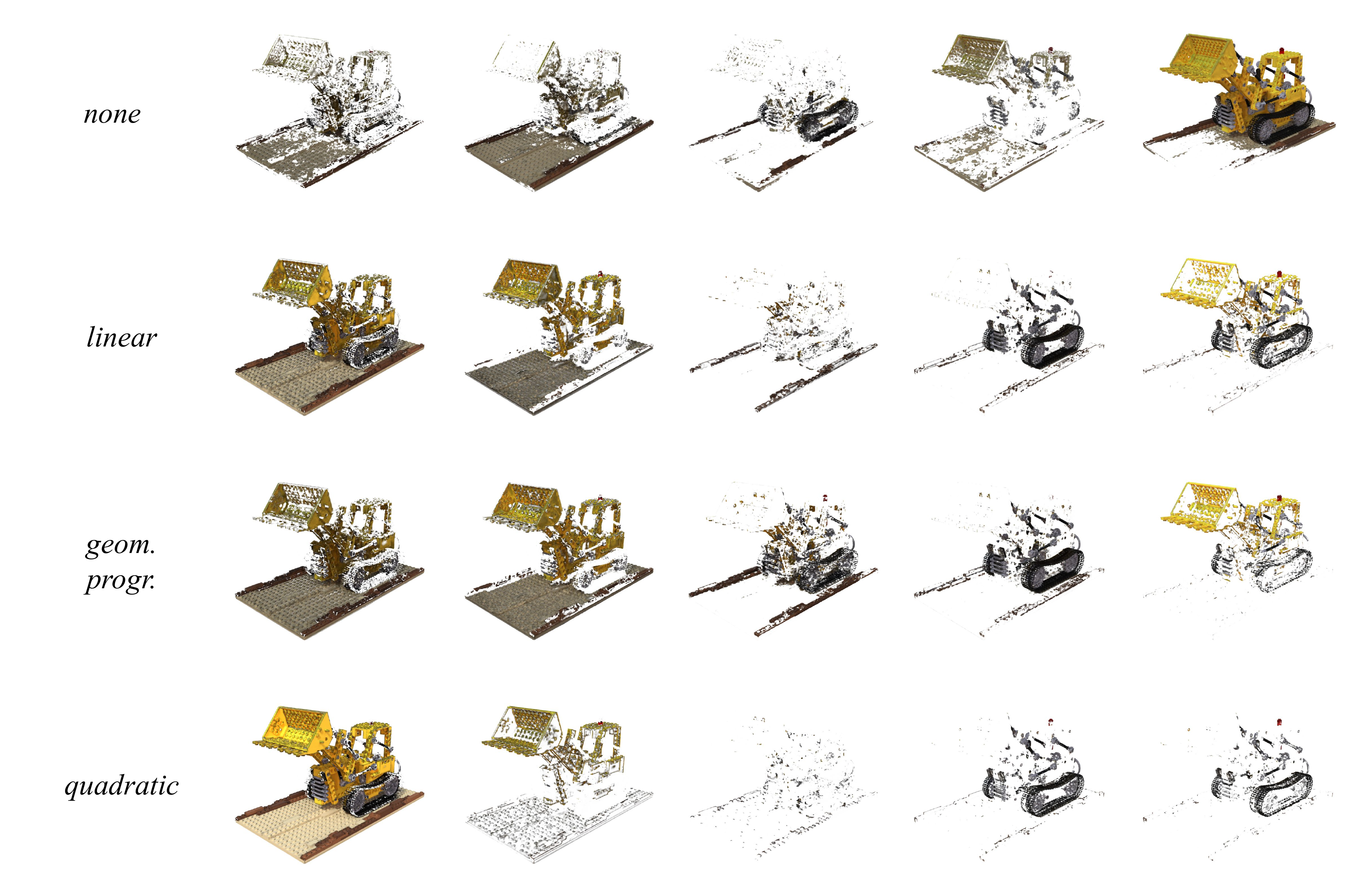}
  \caption{Per-experts outputs with a MoE of 5 NeRFs and top-2 with different resolution penalties. Experts' output are ordered by increasing resolution. }
  \label{fig:res_maps}
\end{figure}

In Fig.~\ref{fig:res_maps}, we provide a qualitative evaluation of experts' specialization. With our strategy, we aim to utilize high-resolution models as sparingly as possible.  
\begin{table}[h]
\centering
\caption{Comparison among different penalty strategies}
\label{tab:abl_res}
\resizebox{0.5\textwidth}{!}{%
\begin{tabular}{c|cc|cc}
\hline
\multirow{2}{*}{Strategy} & \multicolumn{2}{c|}{\textit{Top-1}} & \multicolumn{2}{c}{\textit{Top-2}} \\ 
                          & PSNR  & $\|w\|_0$  & PSNR   & $\|w\|_0$  \\ \hline
\textit{none}             & 33.41            & 32               & 33.61           & 56               \\
\textit{geom. progr.}     & \textbf{33.43}   & \textbf{26}      & \textbf{33.74}  & \textbf{39}      \\
\textit{quadratic}        & 33.31            & 20               & 33.64           & 37               \\
\textit{linear}           & 33.41            & 32               & 33.63           & 36               \\ \hline
\end{tabular}%
}
\end{table}

\section{Visualizing Experts Specialization}
\begin{figure}[ht!]
    \centering
    \includegraphics[width=\textwidth,height=0.9\textheight,keepaspectratio]{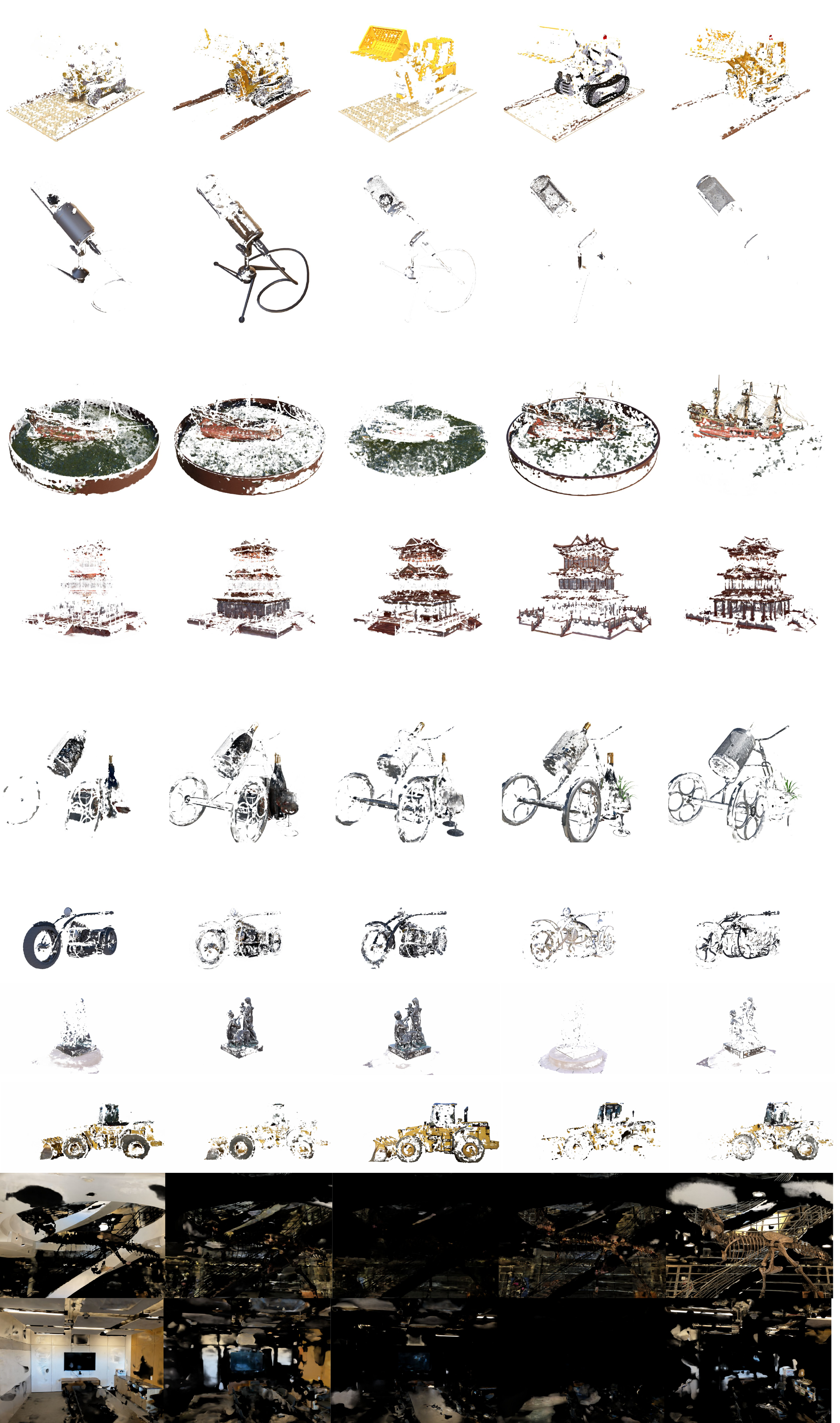}
    \caption{Per-expert output with Top-1 and 5 experts. Outputs are ordered from left to right by increasing experts' resolution.}
    \label{fig:top1-visual}
\end{figure}

\begin{figure}[ht!]
    \centering
    \includegraphics[width=\textwidth,height=0.9\textheight,keepaspectratio]{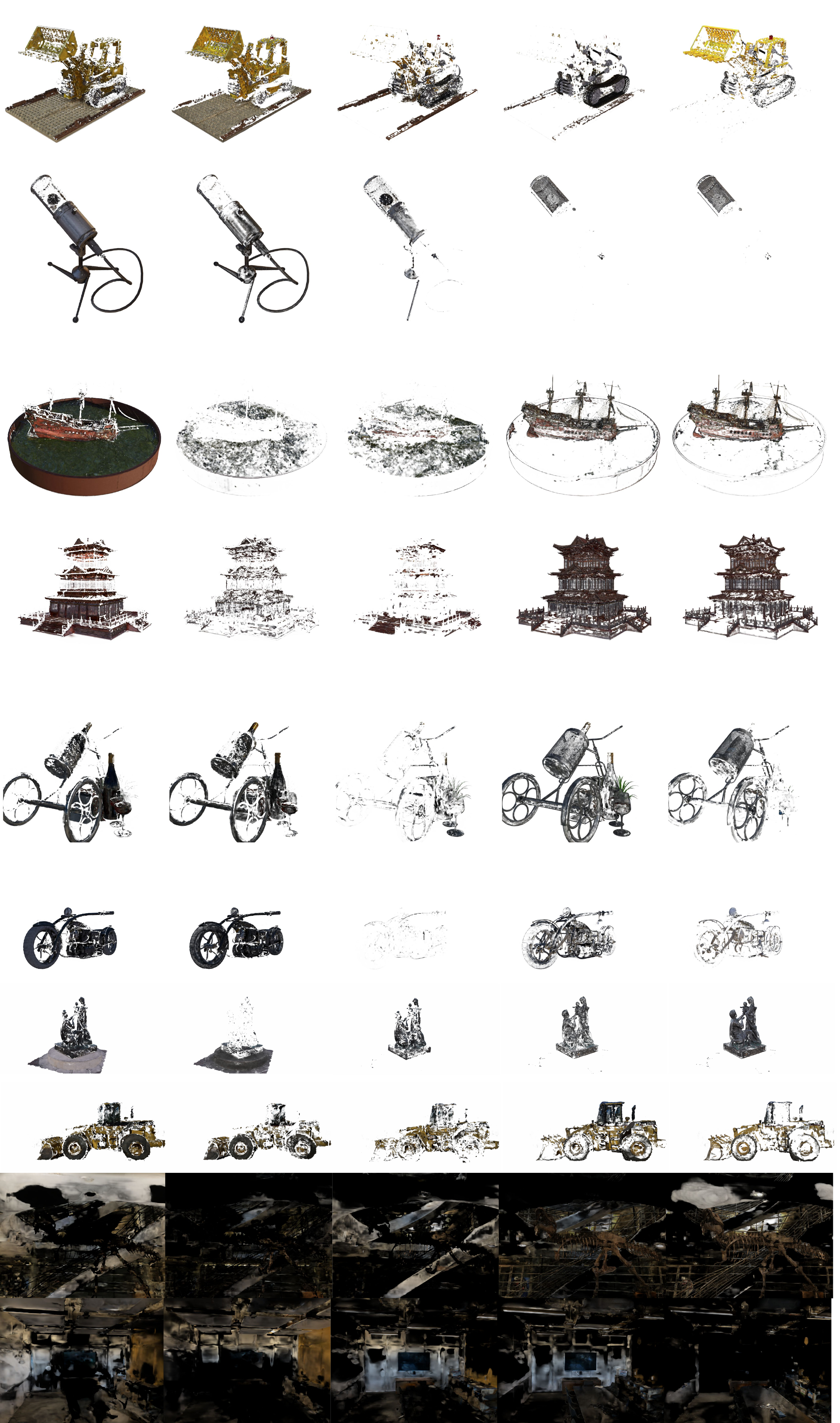}
    \caption{Per-expert output with Top-2 and 5 experts. Outputs are ordered from left to right by increasing experts' resolution.}
    \label{fig:top2-visual}
\end{figure}

In Fig.~\ref{fig:top1-visual} and~\ref{fig:top2-visual} we show the output for each expert in the MoE with $M=5$, with respectively Top-$1$ and Top-$2$ function and resolution penalty. 

\section{Comparison with SoTA Methods}
Although our goal was to focus on fast models and improve reconstruction quality at a lower computational cost compared to their baseline counterparts, our method achieves comparable (and often superior) accuracy to explicit models. Table \ref{tab:vs-sota} presents a comparison with three state-of-the-art implicit models: Mip-NeRF~\cite{barron2021mip}, Mip-NeRF 360~\cite{barron2022mip} and Zip-NeRF~\cite{barron2023zip}. Training times refers to a \textit{single GPU configuration} (NVIDIA A40). Mip-NeRF was trained for $1 000 000$ iterations; Mip-NeRF 360 and Zip-NeRF for $200 000$ iterations.

\begin{table}[t]
\centering
\caption{Comparison among boosted Fast-NeRF and implicit SoTA methods on Synthetic-NeRF dataset.}
\label{tab:vs-sota}
\resizebox{0.5\textwidth}{!}{%
\begin{tabular}{lccc}
\hline
                                              &  \textbf{PSNR}$\uparrow$ &  \textbf{SSIM} $\uparrow$ &  \textbf{Time} $\downarrow$\\ \hline
DVGO \textsubscript{Top-1}    & 33.43                                     & 0.964                                     & 21'                                         \\
TensoRF \textsubscript{Top-1} & 33.68                                     & 0.965                                     & 69'                                          \\
iNGP \textsubscript{Top-1}    & 33.56                                     & 0.963                                     & 32'                                         \\ \hline
DVGO \textsubscript{Top-2}    & 33.74                                     & 0.965                                     & 25'                                         \\
TensoRF \textsubscript{Top-2} & 34.09                                     & 0.968                                     & 76'                                         \\
iNGP \textsubscript{Top-2}    & 33.83                                     & 0.965                                     & 34'                                         \\ \hline
Mip-NeRF                                      & 33.09                                     & 0.961                                     & 16h                                         \\
Mip-NeRF360                                   & 32.96                                     & 0.960                                     & 5h                                          \\
Zip-NeRF                                      & 33.69                                     & 0.973                                     & 7h                                          \\ \hline
\end{tabular}%
}
\end{table}

\section{Additional Results}
\label{sec:add-results}

\begin{figure}[ht!]
    \centering
    \begin{subfigure}[b]{0.24\textwidth}
        \includegraphics[width=\textwidth]{figures/dvgo_Synthetic-NeRF_flops_final.pdf}
   
    \end{subfigure}
    \hfill
    \begin{subfigure}[b]{0.24\textwidth}
        \includegraphics[width=\textwidth]{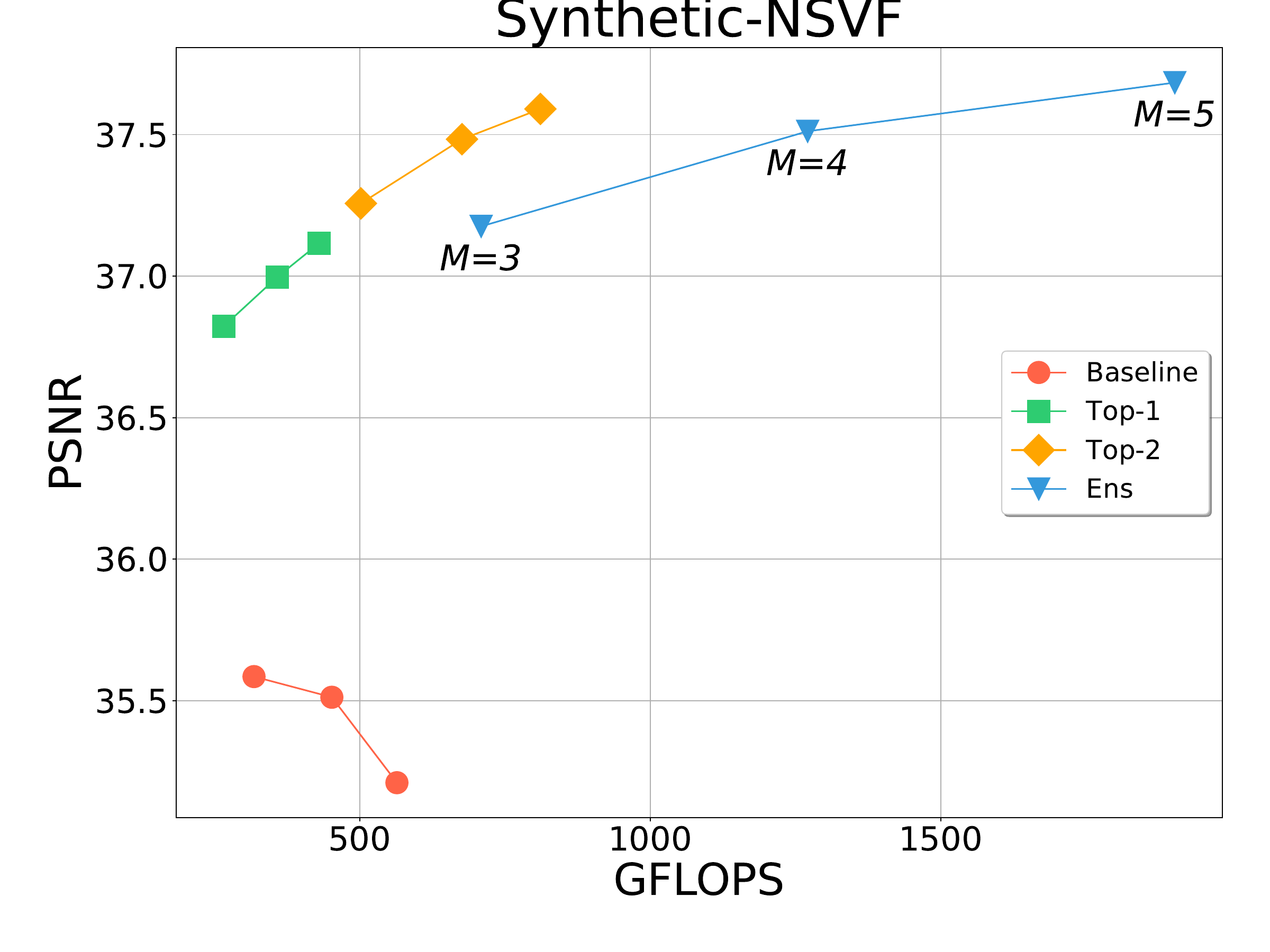}

    \end{subfigure}
    \hfill
        \begin{subfigure}[b]{0.24\textwidth}
        \includegraphics[width=\textwidth]{figures/dvgo_TanksAndTemple_flops_final.pdf}

    \end{subfigure}
    \hfill
    \begin{subfigure}[b]{0.24\textwidth}
        \includegraphics[width=\textwidth]{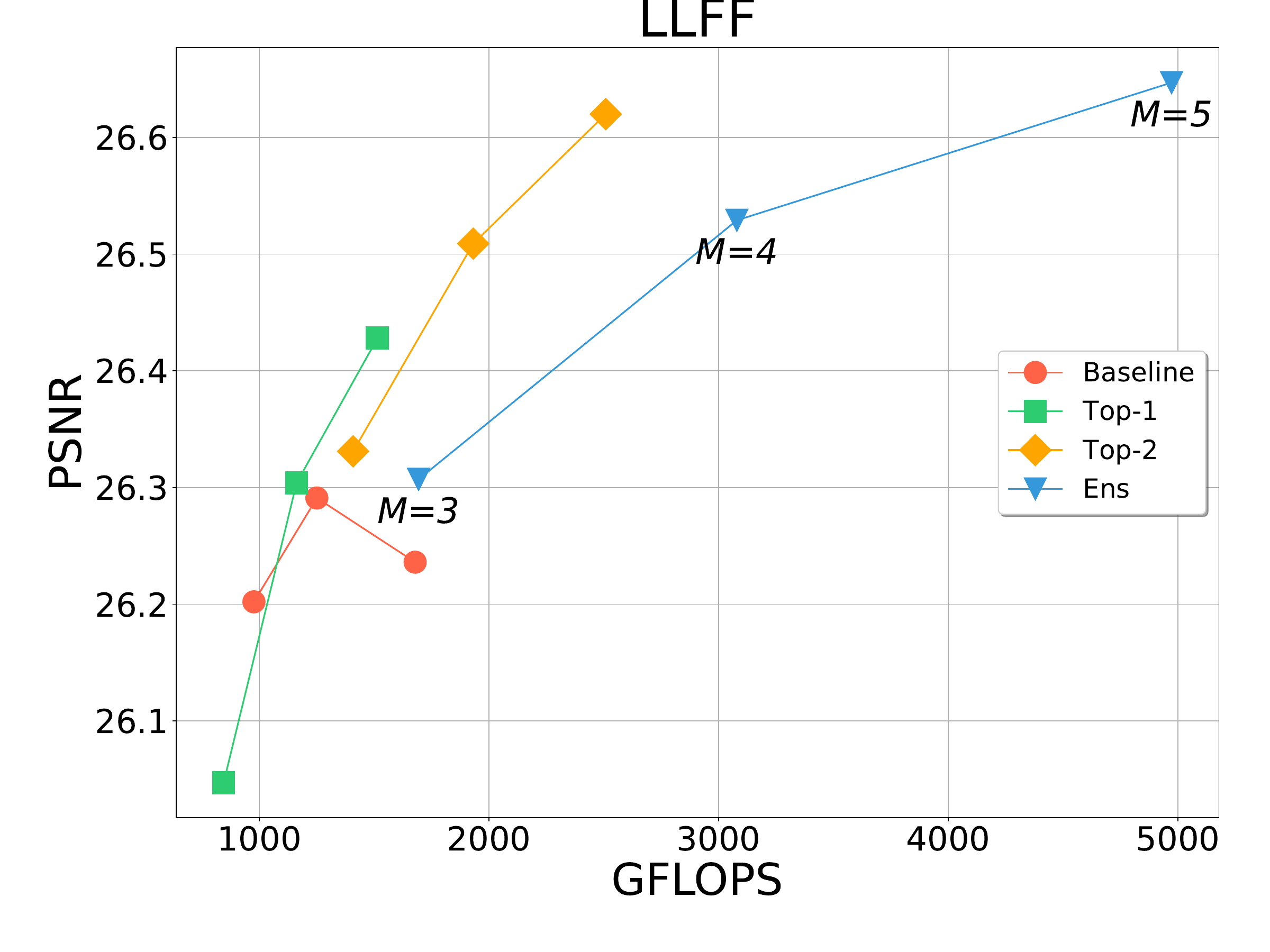}

    \end{subfigure}

    \medskip

        \begin{subfigure}[b]{0.24\textwidth}
        \includegraphics[width=\textwidth]{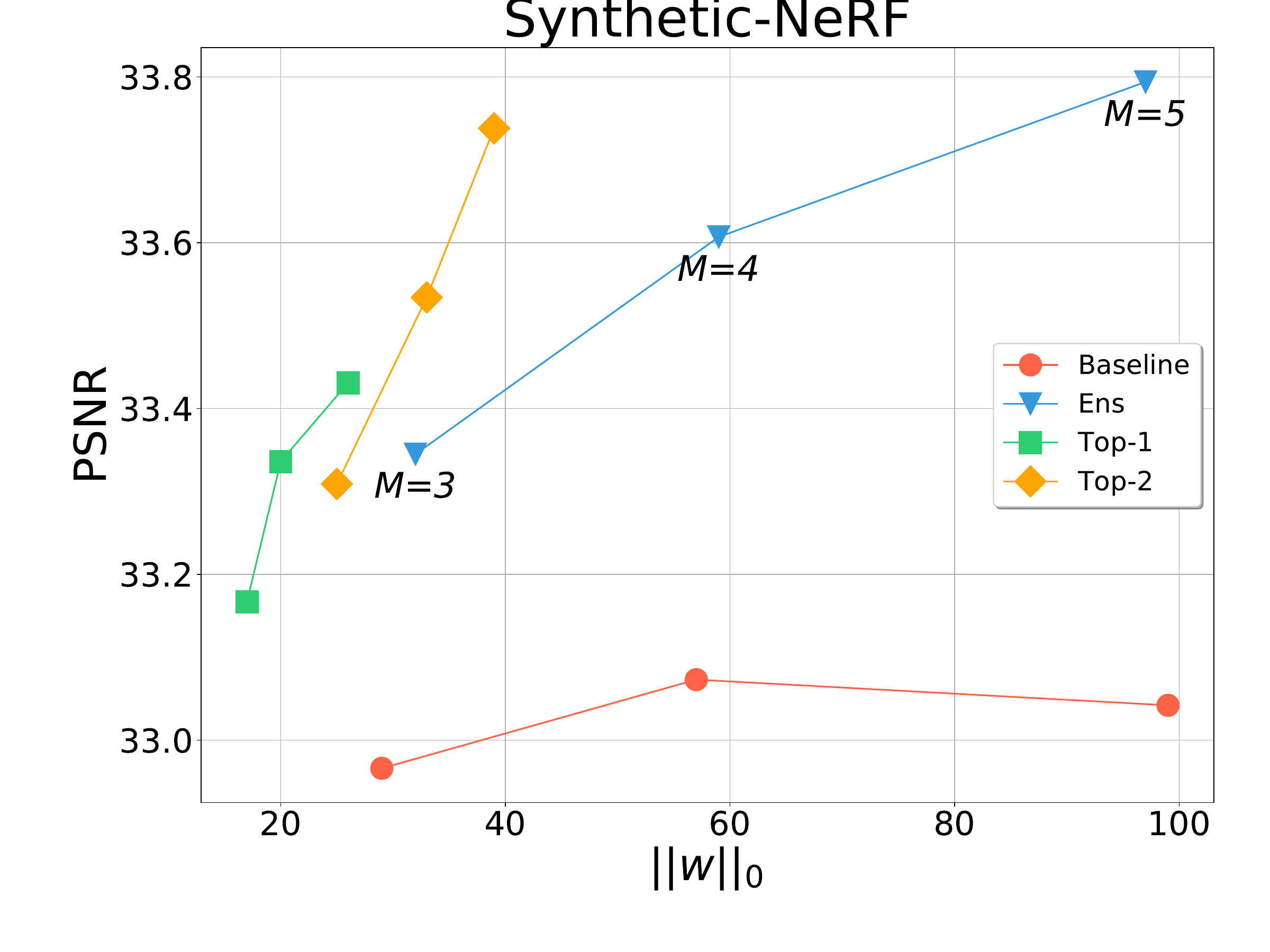}
   
    \end{subfigure}
    \hfill
    \begin{subfigure}[b]{0.24\textwidth}
        \includegraphics[width=\textwidth]{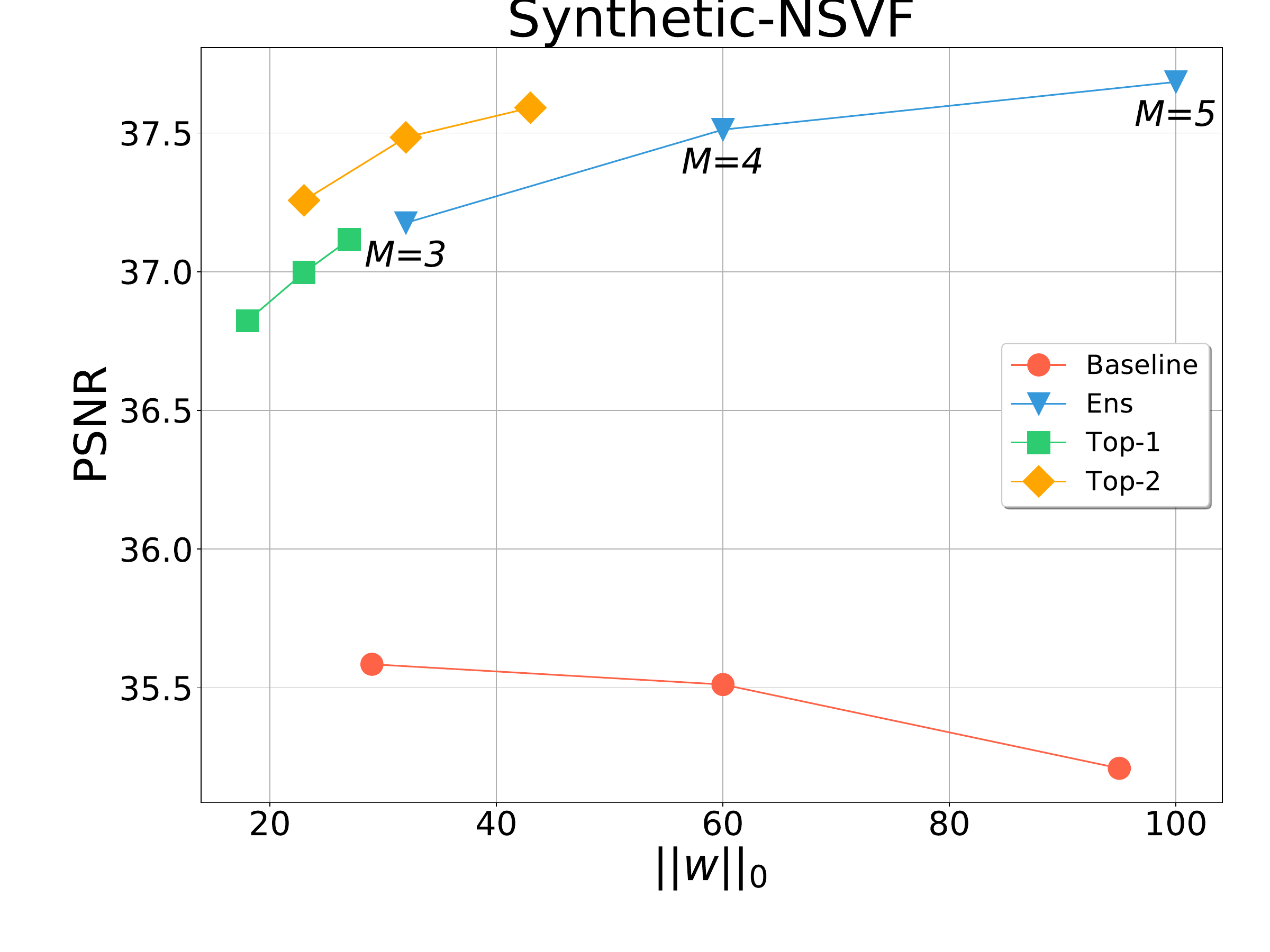}

    \end{subfigure}
    \hfill
        \begin{subfigure}[b]{0.24\textwidth}
        \includegraphics[width=\textwidth]{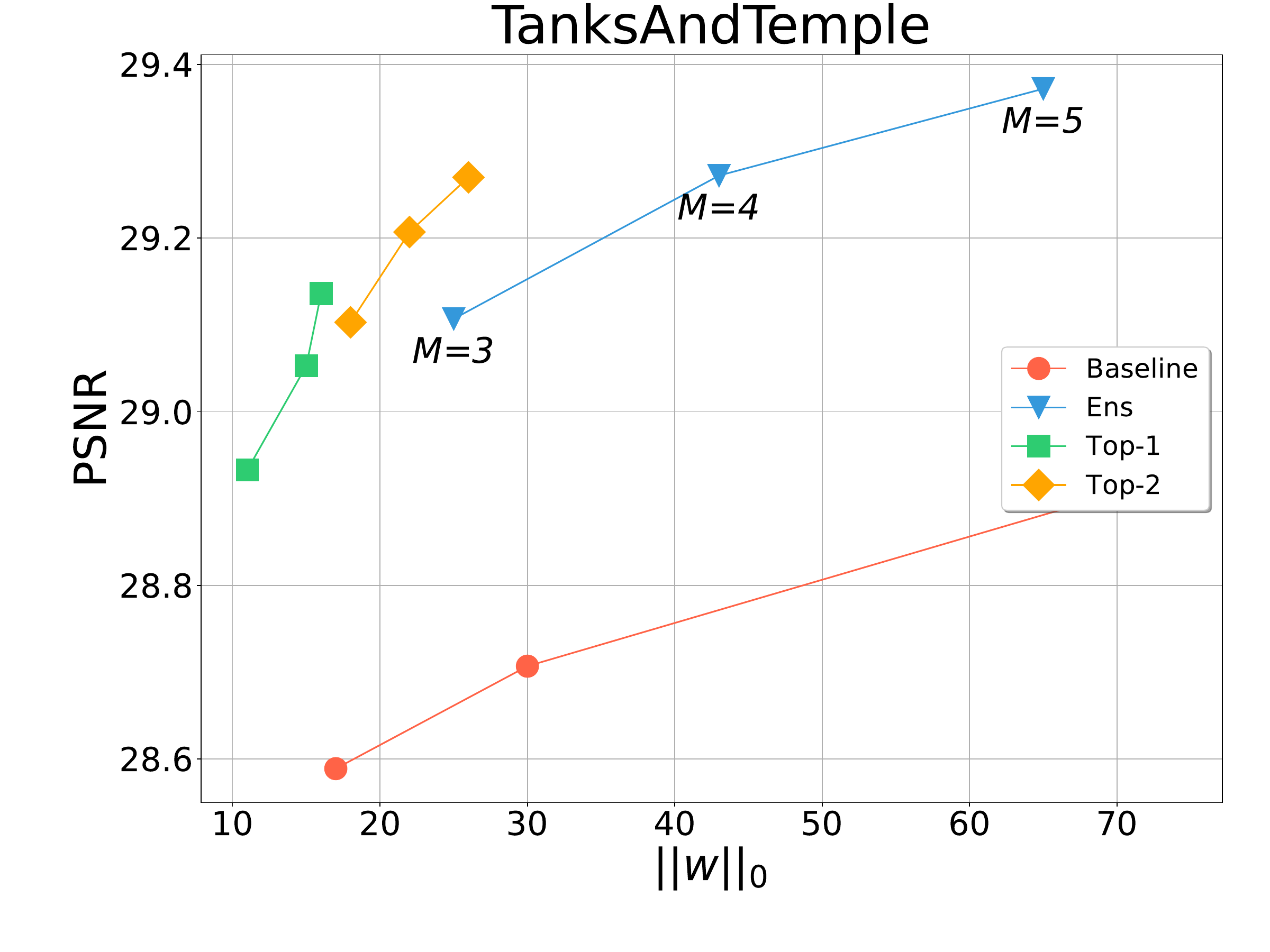}

    \end{subfigure}
    \hfill
    \begin{subfigure}[b]{0.24\textwidth}
        \includegraphics[width=\textwidth]{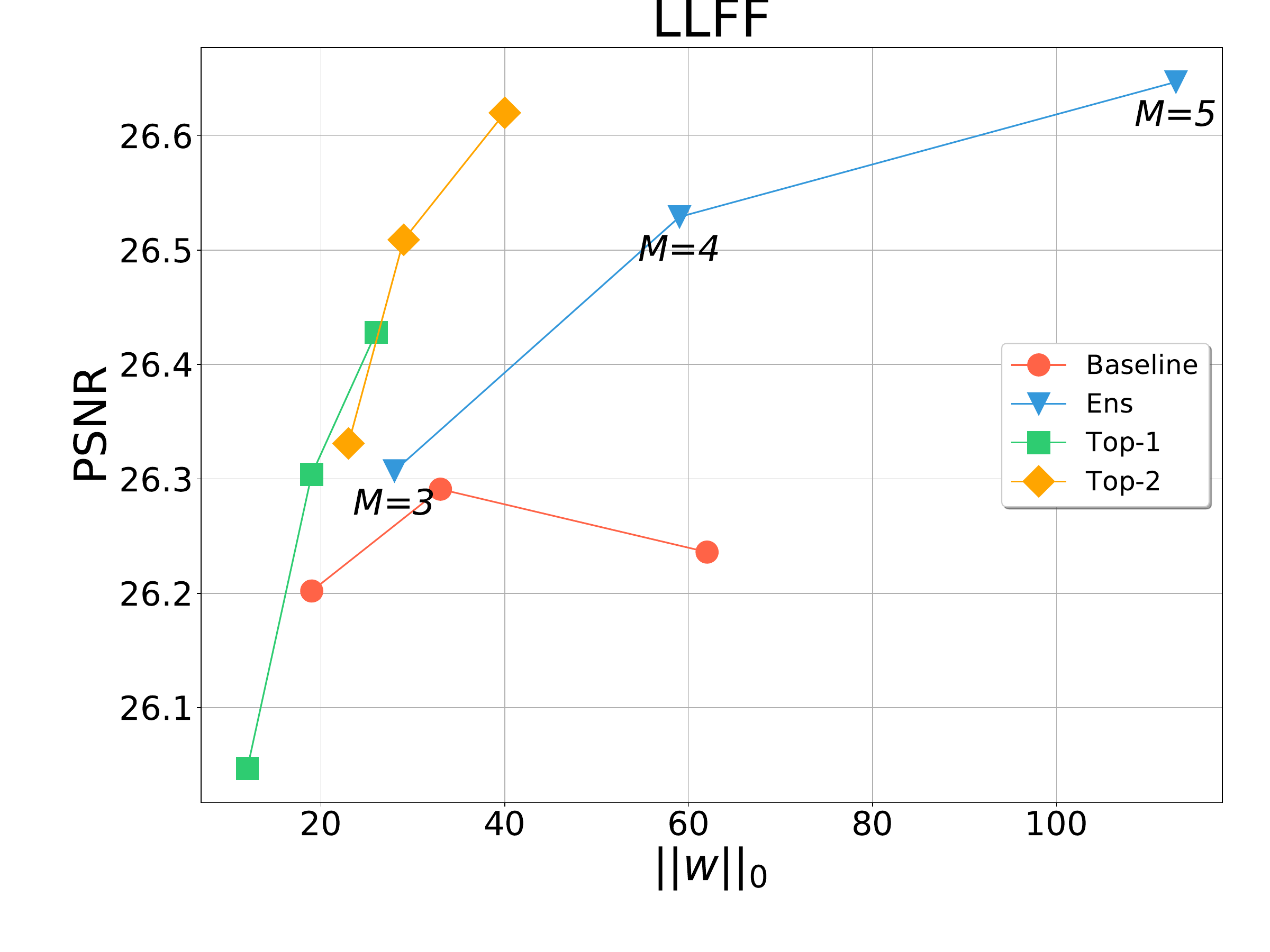}

    \end{subfigure}

    \caption{PSNR/GFLOPs and PSNR/$||w||_0$ plots for DVGO aggregated results.}
    \label{fig:dvgo_aggr}

\end{figure}

\begin{figure}[h]
    \centering
    \begin{subfigure}[b]{0.24\textwidth}
        \includegraphics[width=\textwidth]{figures/tensorf_Synthetic-NeRF_flops_final.pdf}
   
    \end{subfigure}
    \hfill
    \begin{subfigure}[b]{0.24\textwidth}
        \includegraphics[width=\textwidth]{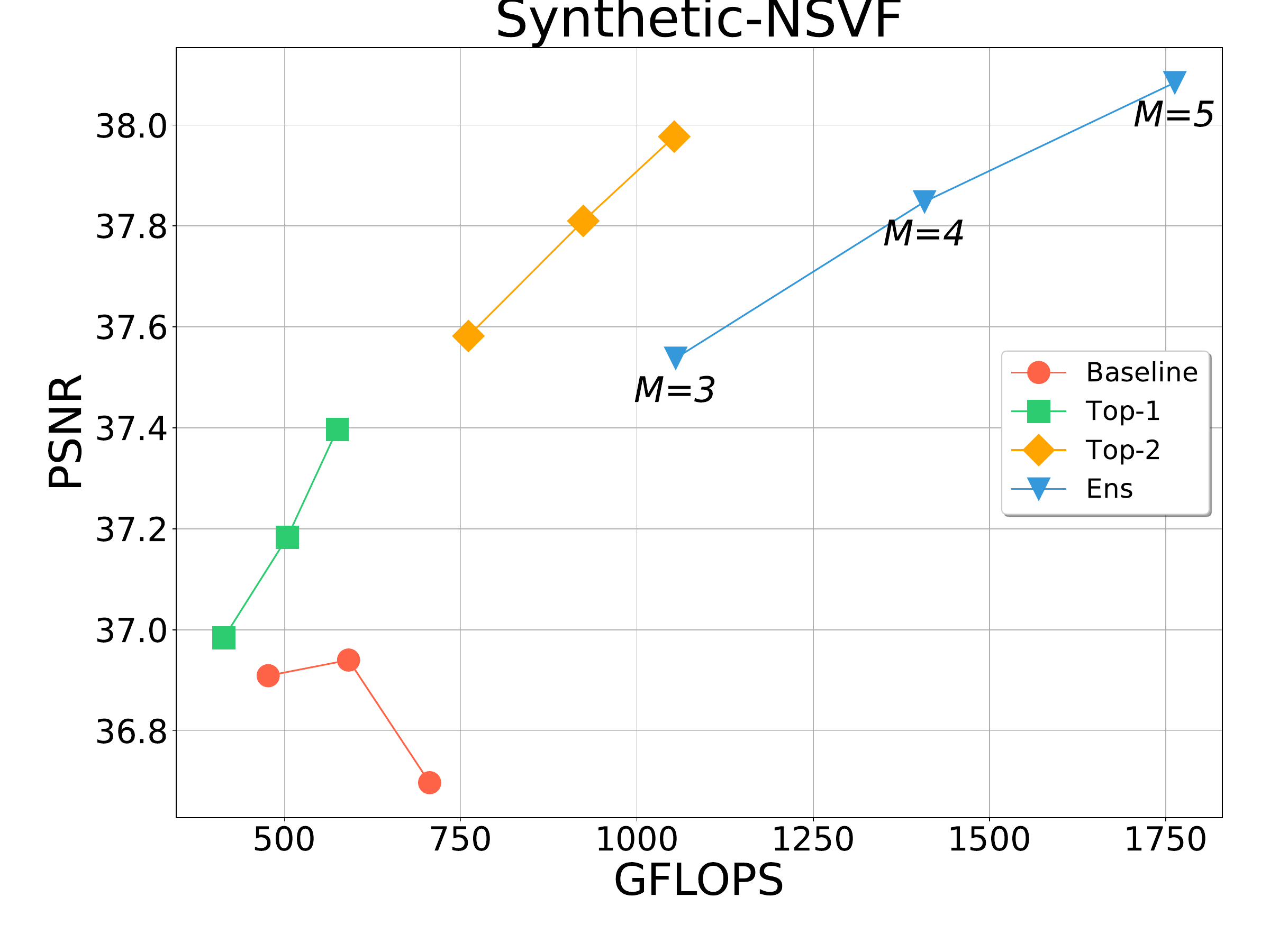}

    \end{subfigure}
    \hfill
        \begin{subfigure}[b]{0.24\textwidth}
        \includegraphics[width=\textwidth]{figures/tensorf_TanksAndTemple_flops_final.pdf}

    \end{subfigure}
    \hfill
    \begin{subfigure}[b]{0.24\textwidth}
        \includegraphics[width=\textwidth]{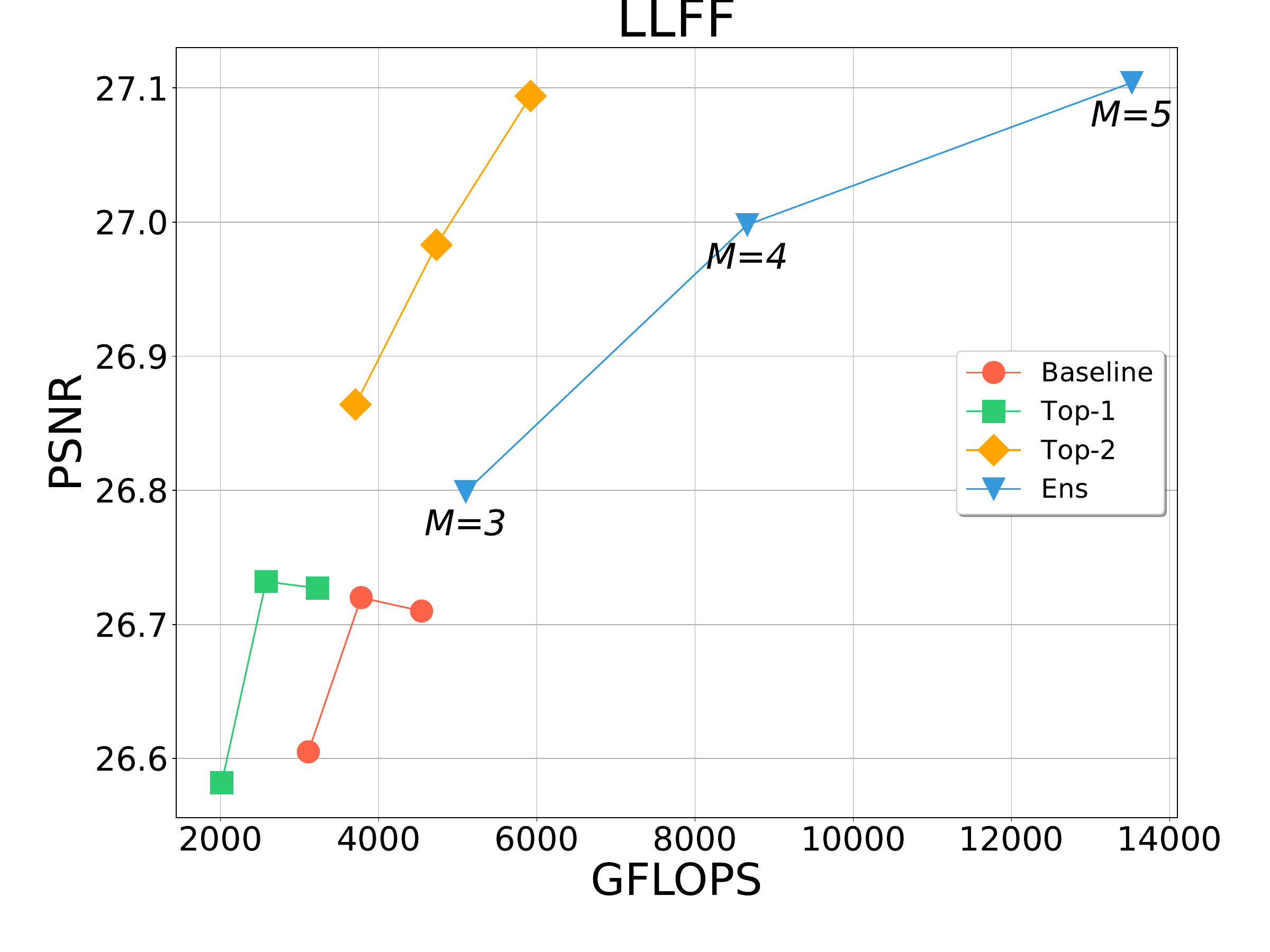}

    \end{subfigure}

    \medskip

        \begin{subfigure}[b]{0.24\textwidth}
        \includegraphics[width=\textwidth]{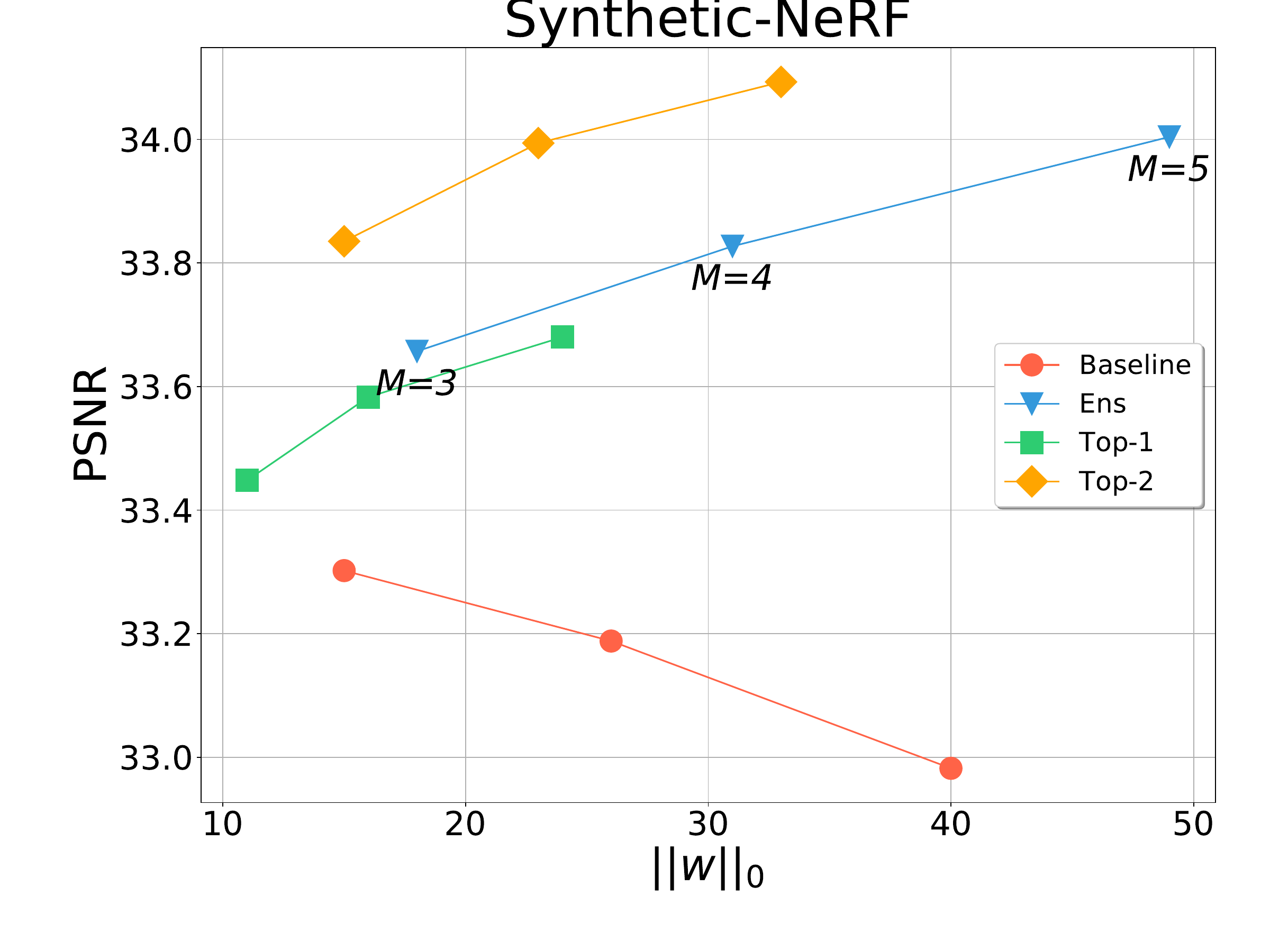}
   
    \end{subfigure}
    \hfill
    \begin{subfigure}[b]{0.24\textwidth}
        \includegraphics[width=\textwidth]{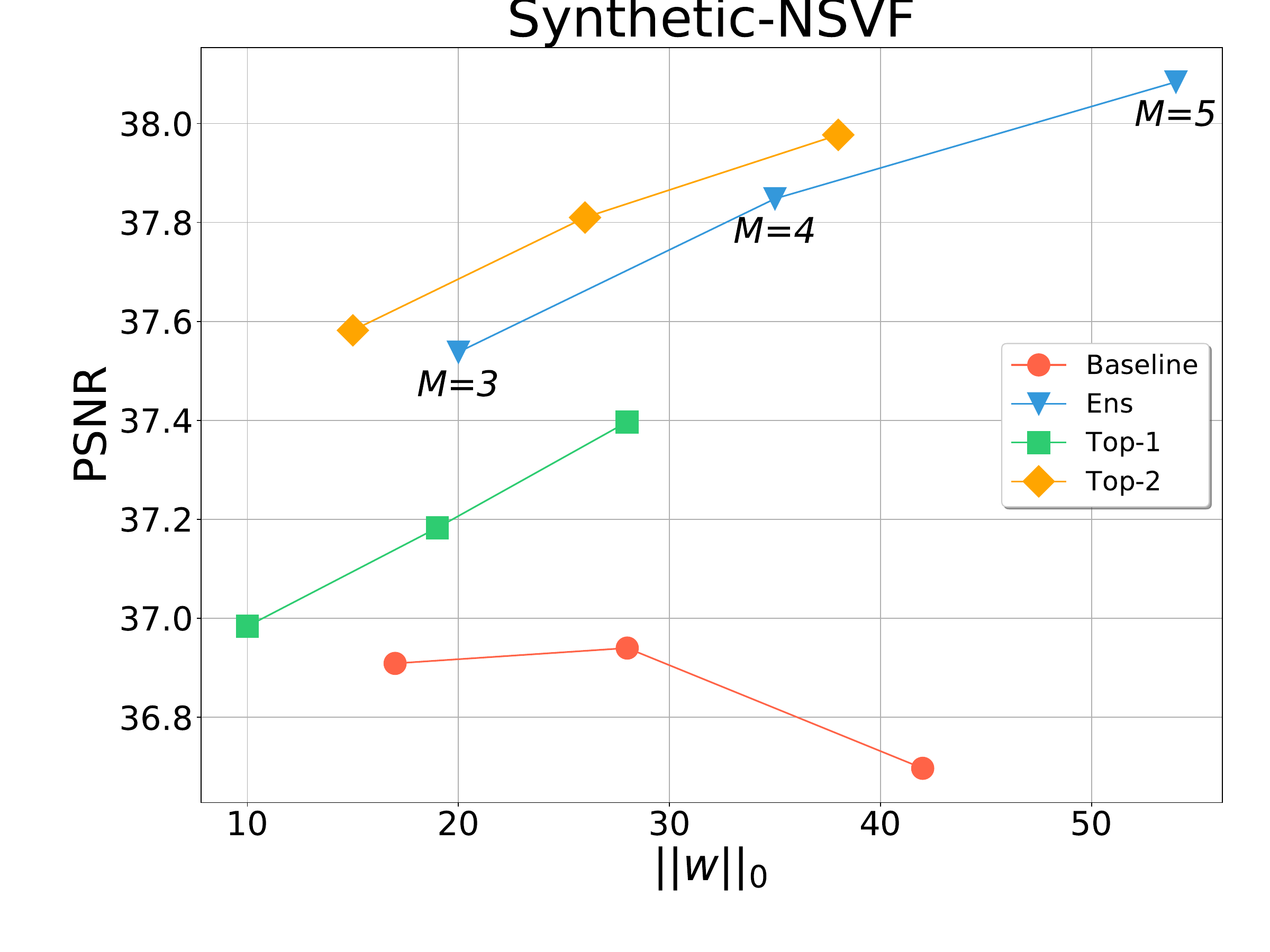}

    \end{subfigure}
    \hfill
        \begin{subfigure}[b]{0.24\textwidth}
        \includegraphics[width=\textwidth]{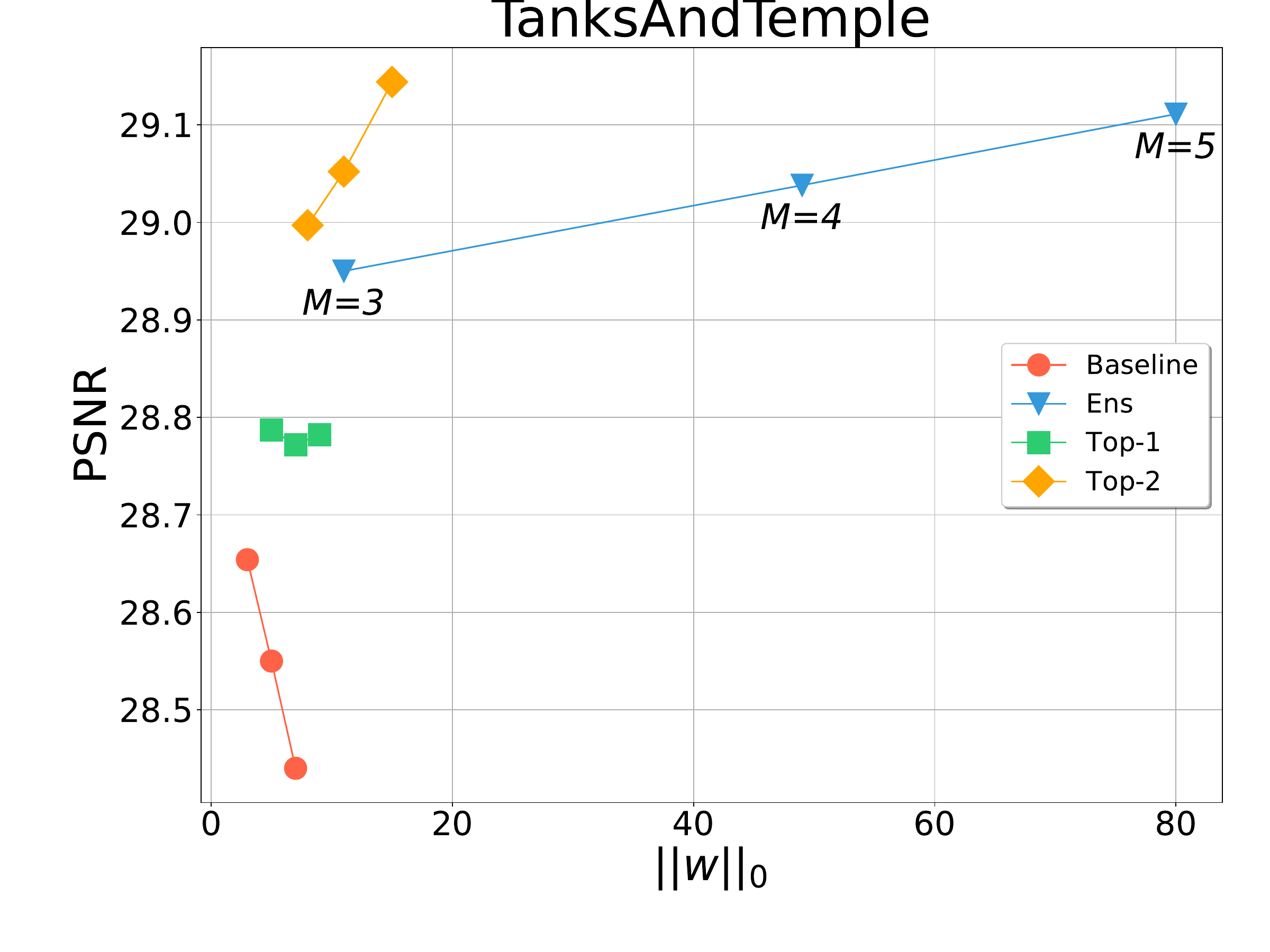}

    \end{subfigure}
    \hfill
    \begin{subfigure}[b]{0.24\textwidth}
        \includegraphics[width=\textwidth]{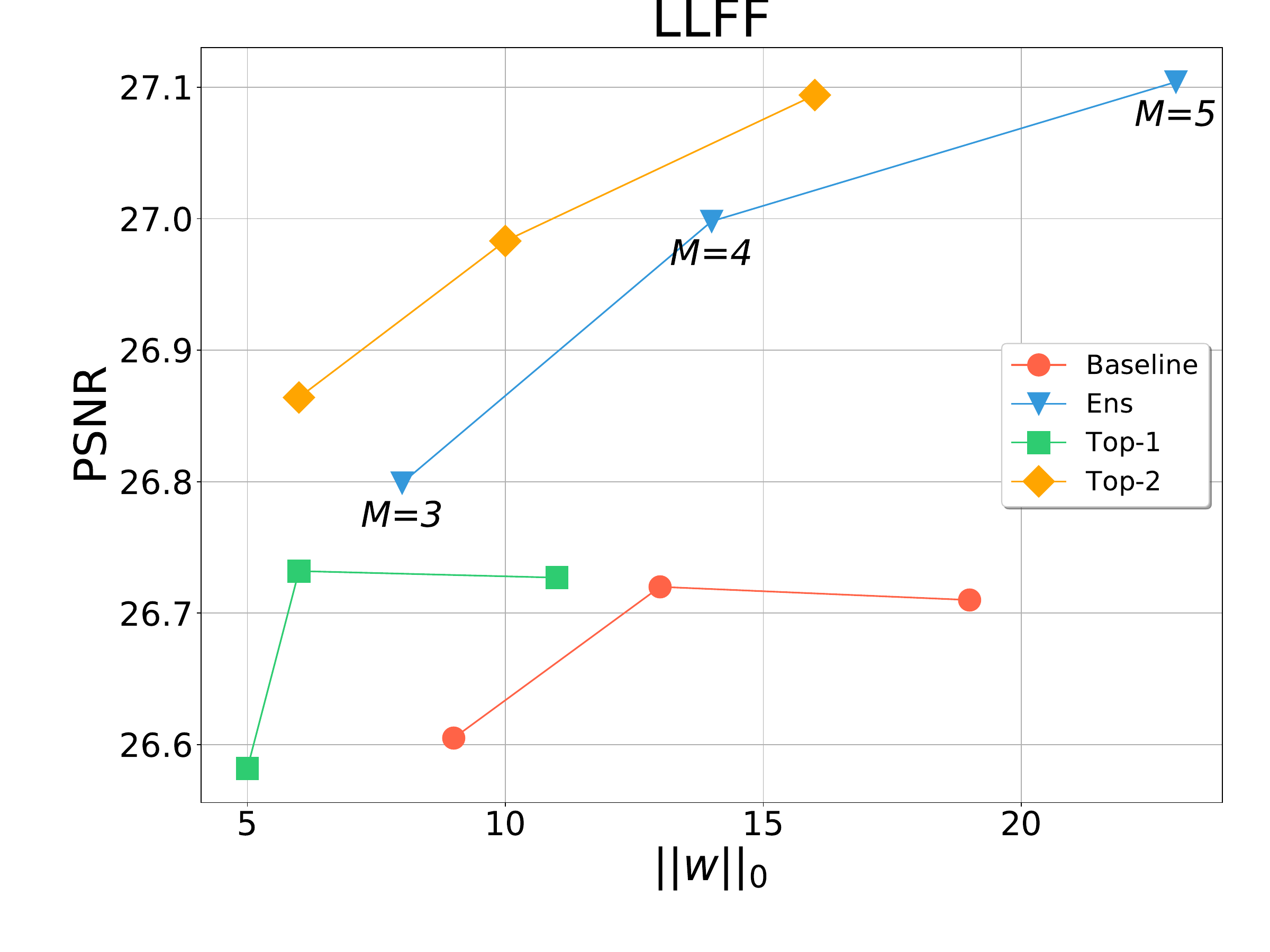}

    \end{subfigure}

    \caption{PSNR/GFLOPs and PSNR/$||w||_0$ plots for aggregated results with TensoRF.}
    \label{fig:tensorf_aggr}

\end{figure}

\begin{figure}[h]
    \centering
    \begin{subfigure}[b]{0.24\textwidth}
        \includegraphics[width=\textwidth]{figures/ngp_Synthetic-NeRF_flops_final.pdf}
   
    \end{subfigure}
    \hfill
    \begin{subfigure}[b]{0.24\textwidth}
        \includegraphics[width=\textwidth]{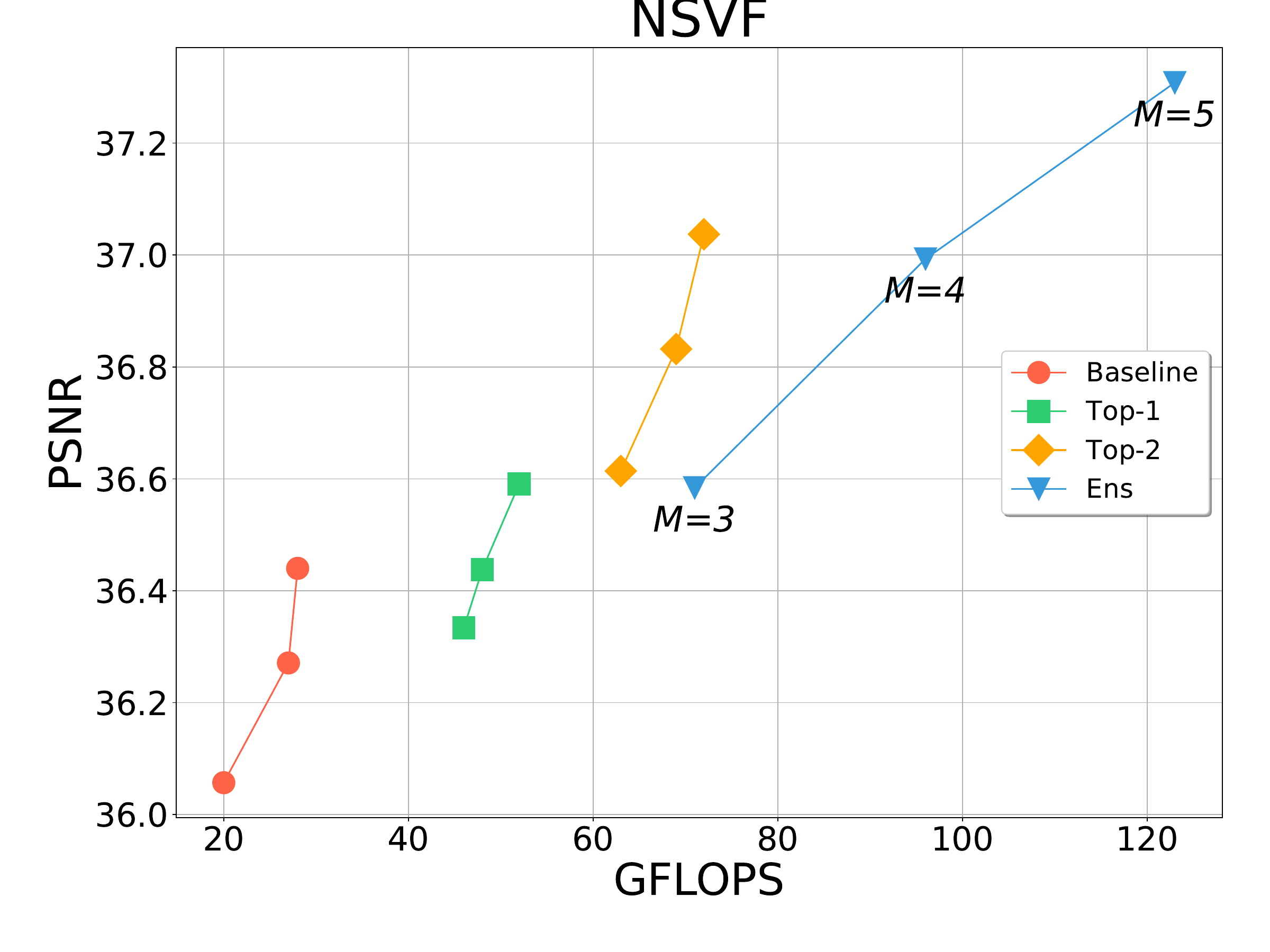}

    \end{subfigure}
    \hfill
        \begin{subfigure}[b]{0.24\textwidth}
        \includegraphics[width=\textwidth]{figures/ngp_TanksAndTemple_flops_final.pdf}

    \end{subfigure}
    \hfill
    \begin{subfigure}[b]{0.24\textwidth}
        \includegraphics[width=\textwidth]{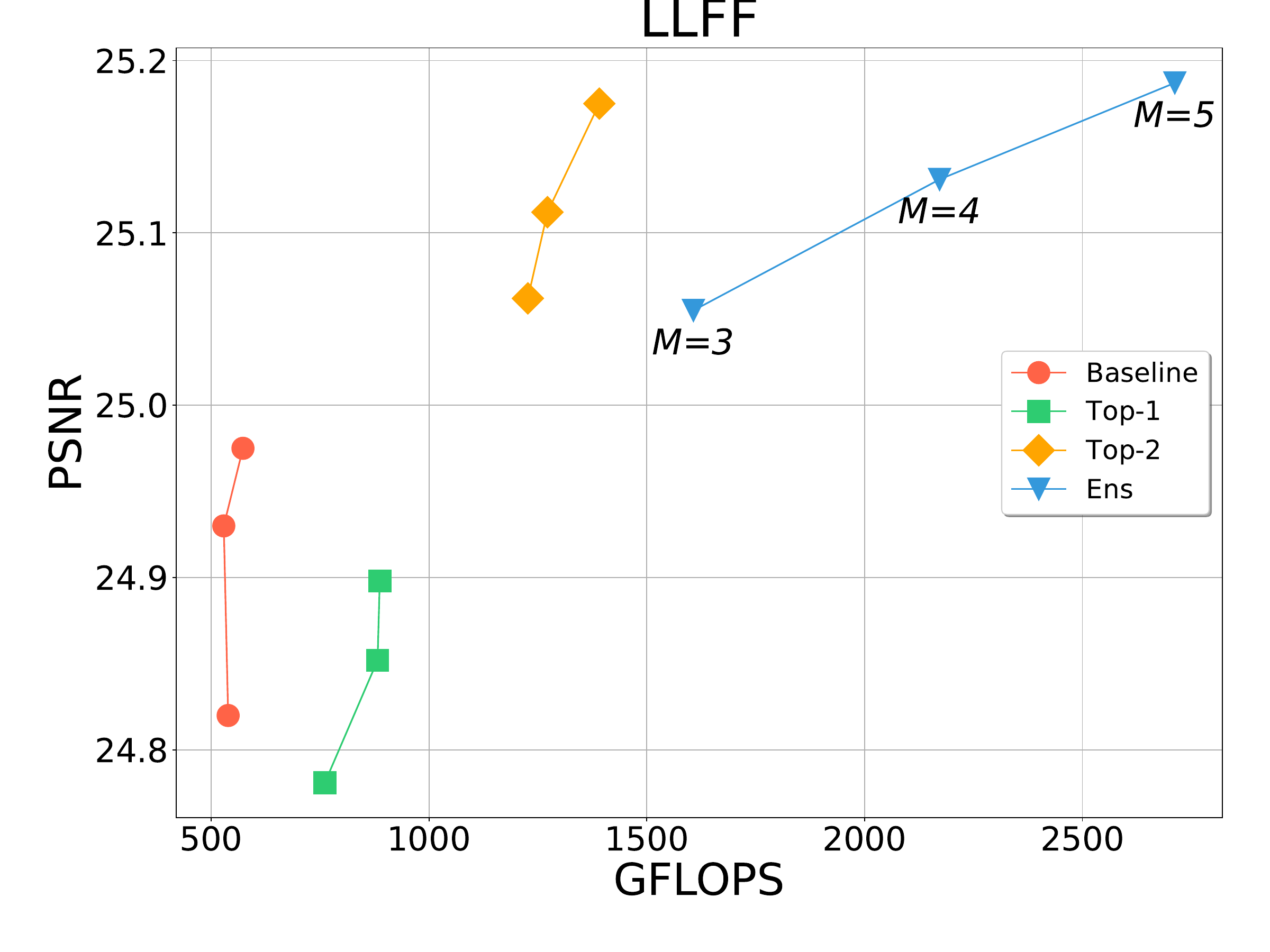}

    \end{subfigure}

    \medskip

        \begin{subfigure}[b]{0.24\textwidth}
        \includegraphics[width=\textwidth]{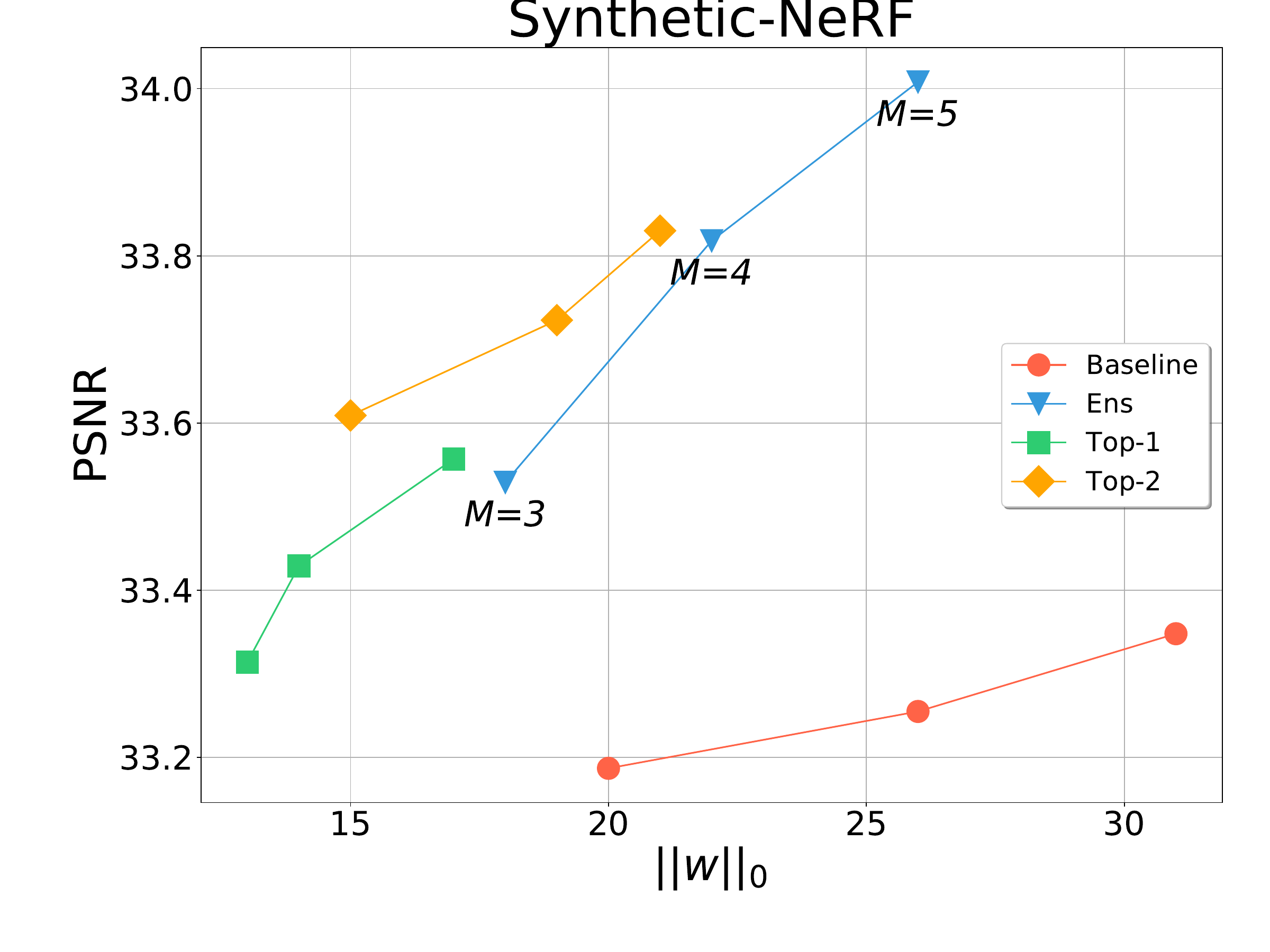}
   
    \end{subfigure}
    \hfill
    \begin{subfigure}[b]{0.24\textwidth}
        \includegraphics[width=\textwidth]{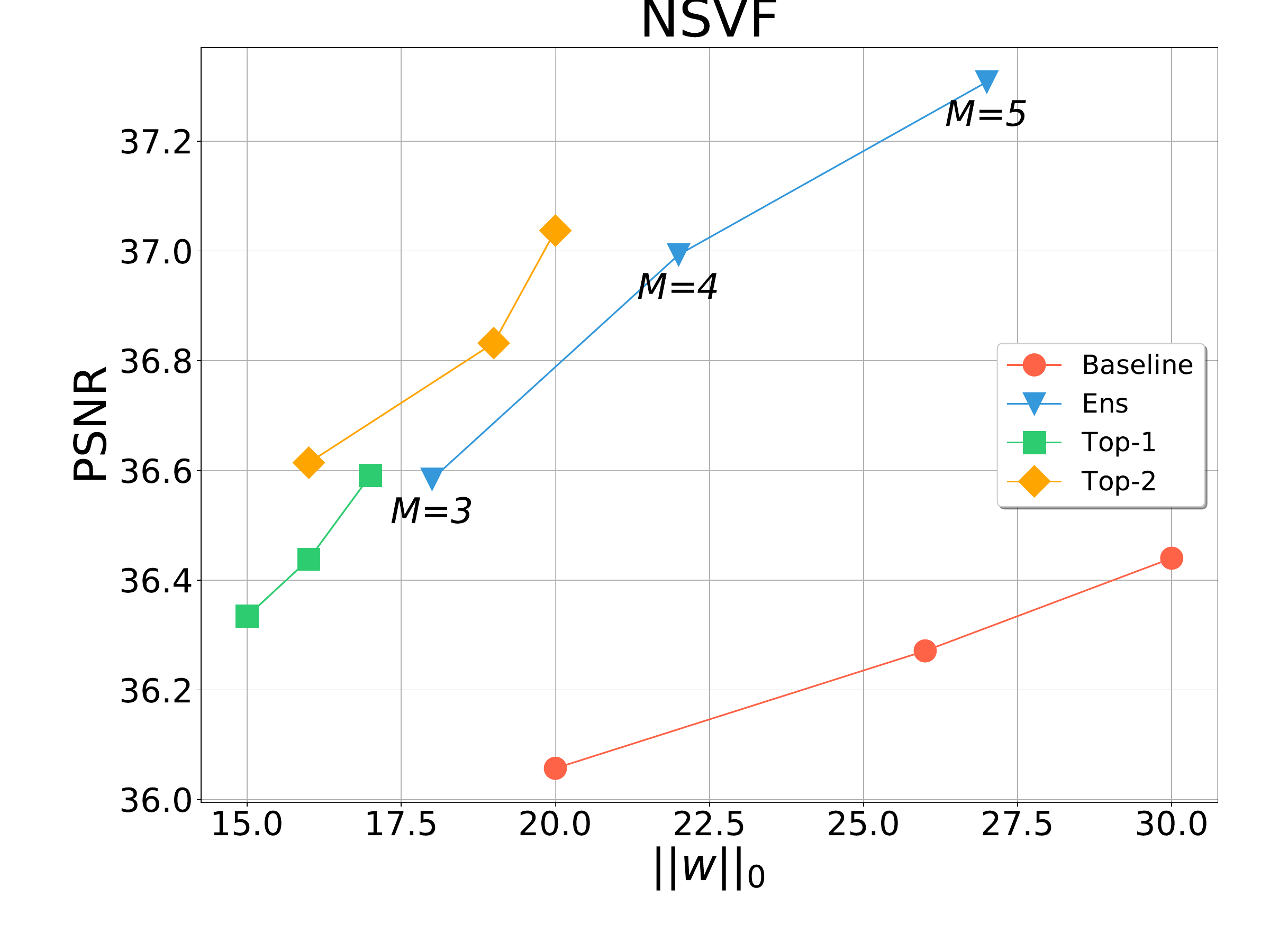}

    \end{subfigure}
    \hfill
        \begin{subfigure}[b]{0.24\textwidth}
        \includegraphics[width=\textwidth]{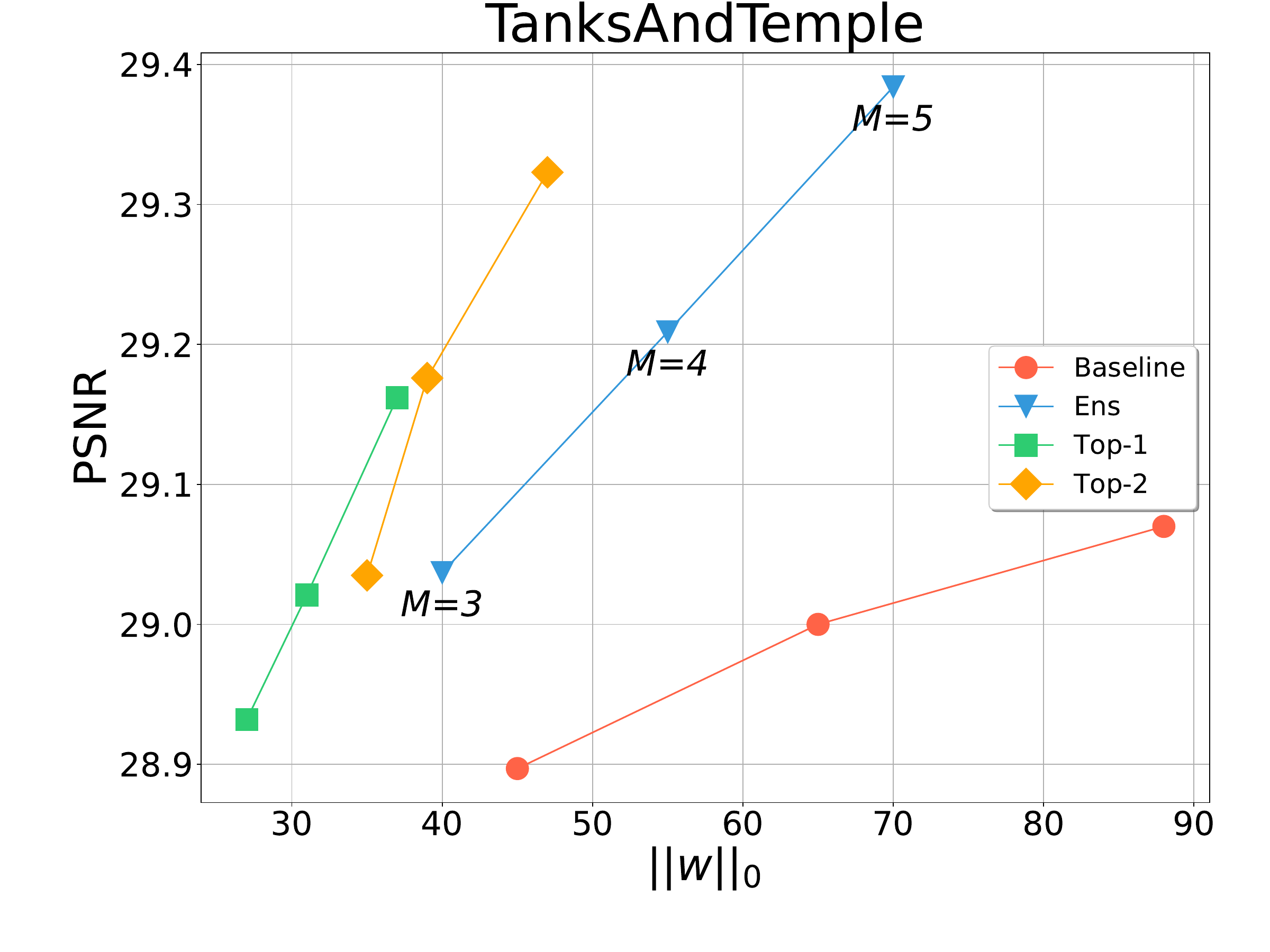}

    \end{subfigure}
    \hfill
    \begin{subfigure}[b]{0.24\textwidth}
        \includegraphics[width=\textwidth]{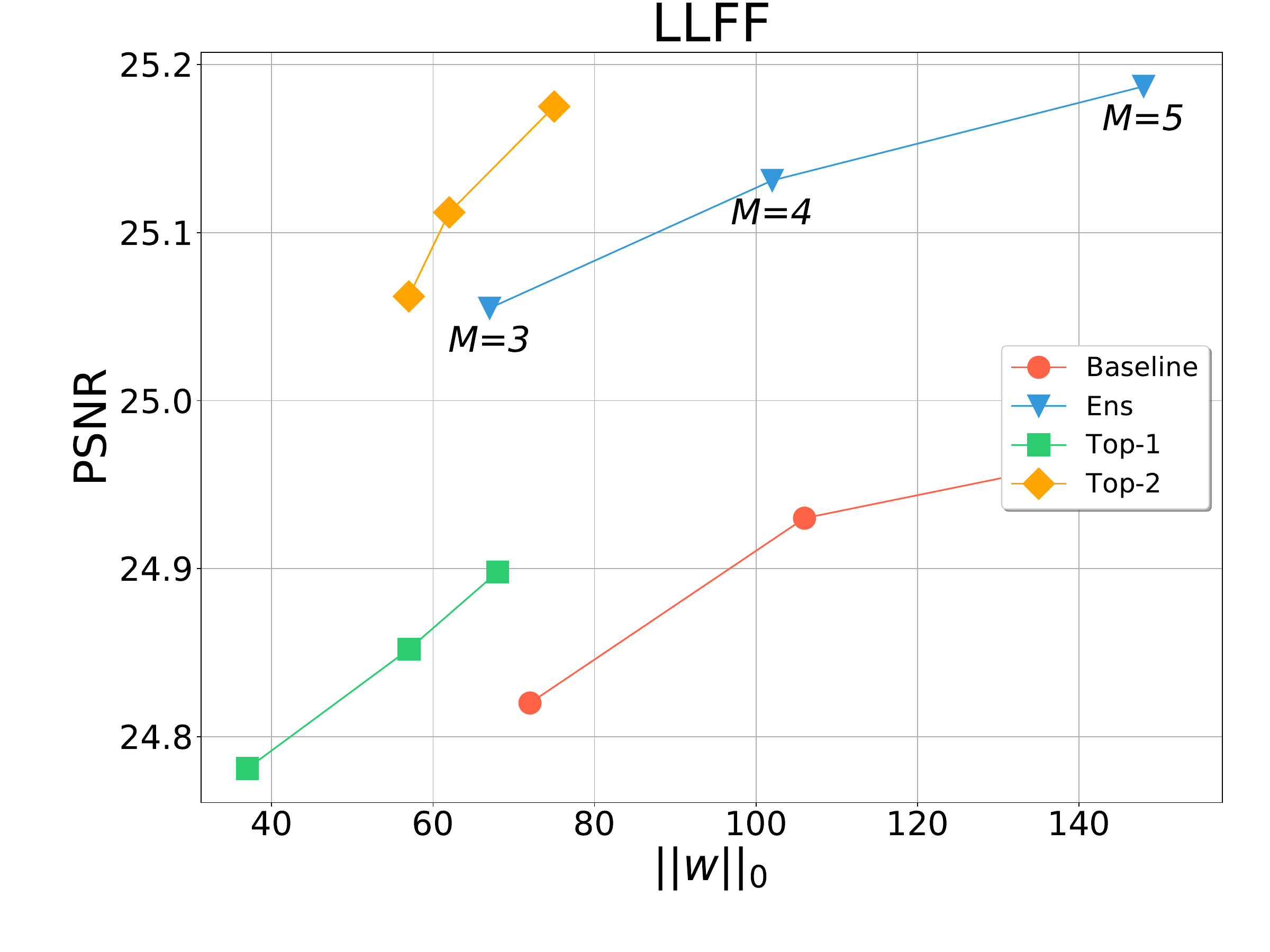}

    \end{subfigure}

    \caption{PSNR/GFLOPs and PSNR/$||w||_0$ plots for aggregated results with Instant-NGP.}
    \label{fig:ngp_aggr}

\end{figure}

In this section, we present additional results for DVGO, TensoRF, and Instant-NGP with varying values of $M={3, 4, 5}$.
In Table~\ref{tab:agg-dvgo} and Figure~\ref{fig:dvgo_aggr}, we provide aggregated results for all the tested datasets on DVGO, including plots of PSNR/GLOPS and PSNR/$\|w\|_0$.
Similarly, in Table~\ref{tab:agg-tensorf} and Figure~\ref{fig:tensorf_aggr}, we present the results for TensoRF. The results for Instant-NGP are shown in Table~\ref{tab:agg-ngp} and Figure~\ref{fig:ngp_aggr}.

\begin{table}[H]
\caption{Aggregated results on DVGO}
\label{tab:agg-dvgo}
\resizebox{\textwidth}{!}{%
\begin{tabular}{cccccc|cccc|cccc}
\hline
\multirow{2}{*}{\textit{Dataset}}                      & \multirow{2}{*}{\textit{Metrics}}    & \multicolumn{4}{c|}{\textit{M=3}}                                  & \multicolumn{4}{c|}{\textit{M=4}}                                  & \multicolumn{4}{c}{\textit{M=5}}                                   \\ \cline{3-14} 
                                                       &                                      & \textit{baseline} & \textit{Top-1} & \textit{Top-2} & \textit{Ens} & \textit{baseline} & \textit{Top-1} & \textit{Top-2} & \textit{Ens} & \textit{baseline} & \textit{Top-1} & \textit{Top-2} & \textit{Ens} \\ \hline
\multicolumn{1}{c|}{\multirow{5}{*}{\textit{Blender}}} & \multicolumn{1}{c|}{\textbf{PSNR}$~\uparrow$}   & 32.97             & 33.17          & 33.31          & 33.35        & 33.07             & 33.34          & 33.53          & 33.61        & 33.04             & 33.43          & 33.74          & 33.79        \\
\multicolumn{1}{c|}{}                                  & \multicolumn{1}{c|}{\textbf{SSIM}~$\uparrow$}   & 0.962             & 0.962          & 0.964          & 0.964        & 0.963             & 0.963          & 0.965          & 0.965        & 0.963          & 0.964          & 0.965          & 0.966        \\
\multicolumn{1}{c|}{}                                  & \multicolumn{1}{c|}{\textbf{LPIPS}~$\downarrow$}  & 0.026             & 0.026          & 0.025          & 0.025        & 0.025             & 0.025          & 0.024          & 0.023        & 0.026             & 0.024          & 0.022          & 0.022        \\
\multicolumn{1}{c|}{}                                  & \multicolumn{1}{c|}{$\|w\|_0\downarrow$} & 29                & 17             & 25             & 32           & 57                & 20             & 33             & 59           & 99                & 26             & 39             & 97           \\
\multicolumn{1}{c|}{}                                  & \multicolumn{1}{c|}{\textbf{GFLOPs}~$\downarrow$} & 382               & 342            & 647            & 914          & 508               & 431            & 813            & 1529         & 635               & 499            & 940            & 2206         \\
\multicolumn{1}{l|}{}                                  & \multicolumn{1}{c|}{\textbf{time}~$\downarrow$}   & 10'               & 12'            & 15'            & 15'          & 13'               & 16'            & 19'            & 23'          & 26'               & 20'            & 24'            & 32'          \\ \hline
\multicolumn{1}{c|}{\multirow{5}{*}{\textit{NSVF}}}                  & \multicolumn{1}{c|}{\textbf{PSNR}$~\uparrow$}   & 35.59             & 36.82          & 37.26          & 37.18        & 35.51             & 37.00          & 37.48          & 37.51        & 35.21             & 37.12          & 37.59          & 37.68        \\
\multicolumn{1}{c|}{\textit{}}                         & \multicolumn{1}{c|}{\textbf{SSIM}~$\uparrow$}   & 0.979             & 0.982          & 0.984          & 0.984        & 0.979             & 0.983          & 0.985          & 0.985        & 0.977             & 0.984          & 0.986          & 0.986        \\
\multicolumn{1}{c|}{\textit{}}                         & \multicolumn{1}{c|}{\textbf{LPIPS}~$\downarrow$}  & 0.014             & 0.010          & 0.009          & 0.009        & 0.014             & 0.010          & 0.008          & 0.008        & 0.015             & 0.009          & 0.008          & 0.007        \\
\multicolumn{1}{c|}{\textit{}}                         & \multicolumn{1}{c|}{$\|w\|_0\downarrow$} & 29                & 18             & 23             & 32           & 60                & 23             & 32             & 60           & 95                & 27             & 43             & 100          \\
\multicolumn{1}{c|}{\textit{}}                         & \multicolumn{1}{c|}{\textbf{GFLOPs}~$\downarrow$} & 318               & 266            & 502            & 709          & 452               & 358            & 676            & 1271         & 564               & 430            & 811            & 1903         \\
\multicolumn{1}{l|}{}                                  & \multicolumn{1}{c|}{\textbf{time}~$\downarrow$}   & 7'                & 13'            & 16'            & 16'          & 22'               & 17'            & 21'            & 25'          & 12'               & 21'            & 25'            & 35'          \\ \hline
\multicolumn{1}{c|}{\multirow{5}{*}{\textit{TaT}}}                      & \multicolumn{1}{c|}{\textbf{PSNR}$~\uparrow$}   & 28.59             & 28.93          & 29.10          & 29.11        & 28.71             & 29.05          & 29.21          & 29.27        & 28.93             & 29.14          & 29.27          & 29.37        \\
\multicolumn{1}{c|}{\textit{}}                         & \multicolumn{1}{c|}{\textbf{SSIM}~$\uparrow$}   & 0.919             & 0.921          & 0.925          & 0.926        & 0.921             & 0.923          & 0.927          & 0.930        & 0.927             & 0.925          & 0.929          & 0.932        \\
\multicolumn{1}{c|}{\textit{}}                         & \multicolumn{1}{c|}{\textbf{LPIPS}~$\downarrow$}  & 0.131             & 0.115          & 0.113          & 0.113        & 0.121             & 0.110          & 0.108          & 0.107        & 0.107             & 0.108          & 0.105          & 0.103        \\
\multicolumn{1}{c|}{\textit{}}                         & \multicolumn{1}{c|}{$\|w\|_0\downarrow$} & 17                & 11             & 18             & 25           & 30                & 15             & 22             & 43           & 74                & 16             & 26             & 65           \\
\multicolumn{1}{c|}{\textit{}}                         & \multicolumn{1}{c|}{\textbf{GFLOPs}~$\downarrow$} & 1299              & 1223           & 2310           & 3263         & 1547              & 1463           & 2761           & 5193         & 2666              & 1626           & 3066           & 7198         \\
\multicolumn{1}{l|}{}                                  & \multicolumn{1}{c|}{\textbf{time}~$\downarrow$}   & 10'               & 15'            & 18'            & 18'          & 18'               & 19'            & 23'            & 27'          & 22'               & 25'            & 28'            & 38'          \\ \hline
\multicolumn{1}{c|}{\multirow{5}{*}{\textit{LLFF}}}                     & \multicolumn{1}{c|}{\textbf{PSNR}$~\uparrow$}   & 26.20             & 26.05          & 26.33          & 26.31        & 26.29             & 26.30          & 26.51          & 26.53        & 26.24             & 26.43          & 26.62          & 26.65        \\
\multicolumn{1}{c|}{}                                  & \multicolumn{1}{c|}{\textbf{SSIM}~$\uparrow$}   & 0.832             & 0.823          & 0.831          & 0.834        & 0.832             & 0.830          & 0.837          & 0.841        & 0.831             & 0.832          & 0.839          & 0.843        \\
\multicolumn{1}{c|}{}                                  & \multicolumn{1}{c|}{\textbf{LPIPS}~$\downarrow$}  & 0.141             & 0.130          & 0.120          & 0.123        & 0.137             & 0.117          & 0.116          & 0.112        & 0.136             & 0.115          & 0.111          & 0.107        \\
\multicolumn{1}{c|}{}                                  & \multicolumn{1}{c|}{$\|w\|_0\downarrow$} & 19                & 12             & 23             & 28           & 33                & 19             & 29             & 59           & 62                & 26             & 40             & 113          \\
\multicolumn{1}{c|}{}                                  & \multicolumn{1}{c|}{\textbf{GFLOPs}~$\downarrow$} & 976               & 844            & 1408           & 1693         & 1250              & 1162           & 1931           & 3079         & 1678              & 1514           & 2508           & 4972         \\
\multicolumn{1}{l|}{}                                  & \multicolumn{1}{c|}{\textbf{time}~$\downarrow$}   & 12'               & 14'            & 16'            & 13'          & 16'               & 20'            & 23'            & 23'          & 24'               & 28'            & 32'            & 36'          \\ \hline
\end{tabular}%
}
\end{table}

\begin{table}[H]
\caption{Aggregated results on TensoRF}
\label{tab:agg-tensorf}
\resizebox{\textwidth}{!}{%
\begin{tabular}{cccccc|cccc|cccc}
\hline
\multirow{2}{*}{\textit{Dataset}}                      & \multirow{2}{*}{\textit{Metrics}}    & \multicolumn{4}{c|}{\textit{M=3}}                                  & \multicolumn{4}{c|}{\textit{M=4}}                                  & \multicolumn{4}{c}{\textit{M=5}}                                   \\ \cline{3-14} 
                                                       &                                      & \textit{baseline} & \textit{Top-1} & \textit{Top-2} & \textit{Ens} & \textit{baseline} & \textit{Top-1} & \textit{Top-2} & \textit{Ens} & \textit{baseline} & \textit{Top-1} & \textit{Top-2} & \textit{Ens} \\ \hline
\multicolumn{1}{c|}{\multirow{5}{*}{\textit{Blender}}} & \multicolumn{1}{c|}{\textbf{PSNR}$~\uparrow$}   & 33.30             & 33.45          & 33.83          & 33.66        & 33.19             & 33.58          & 33.99          & 33.83        & 32.98             & 33.68          & 34.09          & 34.00        \\
\multicolumn{1}{c|}{}                                  & \multicolumn{1}{c|}{\textbf{SSIM}~$\uparrow$}   & 0.962             & 0.965          & 0.967          & 0.966        & 0.962             & 0.966          & 0.968          & 0.967        & 0.958             & 0.965          & 0.968          & 0.968        \\
\multicolumn{1}{c|}{}                                  & \multicolumn{1}{c|}{\textbf{LPIPS}~$\downarrow$}  & 0.027             & 0.025          & 0.024          & 0.024        & 0.027             & 0.024          & 0.022          & 0.023        & 0.029             & 0.024          & 0.021          & 0.022        \\
\multicolumn{1}{c|}{}                                  & \multicolumn{1}{c|}{$\|w\|_0\downarrow$} & 15                & 11             & 15             & 18           & 26                & 16             & 23             & 31           & 40                & 24             & 33             & 49           \\
\multicolumn{1}{c|}{}                                  & \multicolumn{1}{c|}{\textbf{GFLOPs}~$\downarrow$} & 600               & 520            & 958            & 1327         & 753               & 641            & 1179           & 1770         & 886               & 732            & 1344           & 2214         \\
\multicolumn{1}{l|}{}                                  & \multicolumn{1}{c|}{\textbf{time}~$\downarrow$}   & 29'               & 40'            & 44'            & 40'          & 36'               & 55'            & 61'            & 53'          & 44'               & 69'            & 76'            & 70'          \\ \hline
\multicolumn{1}{c|}{\multirow{5}{*}{\textit{NSVF}}}               & \multicolumn{1}{c|}{\textbf{PSNR}$~\uparrow$}   & 36.91             & 36.98          & 37.58          & 37.54        & 36.95             & 37.18          & 37.81          & 37.85        & 36.70             & 37.40          & 37.98          & 38.08        \\
\multicolumn{1}{c|}{\textit{}}                         & \multicolumn{1}{c|}{\textbf{SSIM}~$\uparrow$}   & 0.982             & 0.983          & 0.985          & 0.985        & 0.981             & 0.983          & 0.986          & 0.986        & 0.981             & 0.984          & 0.986          & 0.987        \\
\multicolumn{1}{c|}{\textit{}}                         & \multicolumn{1}{c|}{\textbf{LPIPS}~$\downarrow$}  & 0.011             & 0.010          & 0.009          & 0.009        & 0.012             & 0.010          & 0.008          & 0.008        & 0.013             & 0.009          & 0.008          & 0.008        \\
\multicolumn{1}{c|}{\textit{}}                         & \multicolumn{1}{c|}{$\|w\|_0\downarrow$} & 17                & 10             & 15             & 20           & 28                & 19             & 26             & 35           & 42                & 28             & 38             & 54           \\
\multicolumn{1}{c|}{\textit{}}                         & \multicolumn{1}{c|}{\textbf{GFLOPs}~$\downarrow$} & 477               & 414            & 761            & 1055         & 591               & 504            & 924            & 1408         & 706               & 575            & 1053           & 1763         \\
\multicolumn{1}{l|}{}                                  & \multicolumn{1}{c|}{\textbf{time}~$\downarrow$}   & 29'               & 45'            & 44'            & 40'          & 36'               & 58'            & 60'            & 56'          & 46'               & 75'            & 75'            & 75'          \\ \hline
\multicolumn{1}{c|}{\multirow{5}{*}{\textit{TaT}}}                      & \multicolumn{1}{c|}{\textbf{PSNR}$~\uparrow$}   & 28.65             & 28.79          & 29.00          & 28.95        & 28.55             & 28.77          & 29.05          & 29.04        & 28.44             & 28.78          & 29.14          & 29.11        \\
\multicolumn{1}{c|}{\textit{}}                         & \multicolumn{1}{c|}{\textbf{SSIM}~$\uparrow$}   & 0.906             & 0.922          & 0.924          & 0.925        & 0.906             & 0.924          & 0.927          & 0.927        & 0.905             & 0.924          & 0.929          & 0.928        \\
\multicolumn{1}{c|}{\textit{}}                         & \multicolumn{1}{c|}{\textbf{LPIPS}~$\downarrow$}  & 0.136             & 0.112          & 0.111          & 0.113        & 0.130             & 0.106          & 0.104          & 0.113        & 0.123             & 0.106          & 0.099          & 0.109        \\
\multicolumn{1}{c|}{\textit{}}                         & \multicolumn{1}{c|}{$\|w\|_0\downarrow$} & 3                 & 5              & 8              & 11           & 5                 & 7              & 11             & 49           & 7                 & 9              & 15             & 80           \\
\multicolumn{1}{c|}{\textit{}}                         & \multicolumn{1}{c|}{\textbf{GFLOPs}~$\downarrow$} & 2379              & 2151           & 3959           & 5481         & 2915              & 2475           & 4549           & 7311         & 3567              & 2791           & 5126           & 9146         \\
\multicolumn{1}{l|}{}                                  & \multicolumn{1}{c|}{\textbf{time}~$\downarrow$}   & 40'               & 41'            & 43'            & 44'          & 54'               & 53'            & 59'            & 72'          & 72'               & 70'            & 78'            & 101'         \\ \hline
\multicolumn{1}{c|}{\multirow{5}{*}{\textit{LLFF}}}                     & \multicolumn{1}{c|}{\textbf{PSNR}$~\uparrow$}   & 26.60             & 26.58          & 26.86          & 26.80        & 26.72             & 26.73          & 26.98          & 27.00        & 26.71             & 26.73          & 27.09          & 27.10        \\
\multicolumn{1}{c|}{}                                  & \multicolumn{1}{c|}{\textbf{SSIM}~$\uparrow$}   & 0.834             & 0.835          & 0.841          & 0.840        & 0.836             & 0.839          & 0.843          & 0.843        & 0.835             & 0.836          & 0.862          & 0.864        \\
\multicolumn{1}{c|}{}                                  & \multicolumn{1}{c|}{\textbf{LPIPS}~$\downarrow$}  & 0.127             & 0.117          & 0.110          & 0.116        & 0.118             & 0.112          & 0.105          & 0.107        & 0.114             & 0.111          & 0.101          & 0.101        \\
\multicolumn{1}{c|}{}                                  & \multicolumn{1}{c|}{$\|w\|_0\downarrow$} & 9                 & 5              & 6              & 8            & 13                & 6              & 10             & 14           & 19                & 11             & 16             & 23           \\
\multicolumn{1}{c|}{}                                  & \multicolumn{1}{c|}{\textbf{GFLOPs}~$\downarrow$} & 3111              & 2016           & 3707           & 5100         & 3779              & 2576           & 4731           & 8660         & 4542              & 3226           & 5921           & 13522        \\
\multicolumn{1}{l|}{}                                  & \multicolumn{1}{c|}{\textbf{time}~$\downarrow$}   & 36'               & 27'            & 31'            & 31'          & 49'               & 37'            & 43'            & 49'          & 58'               & 49'            & 57'            & 68'          \\ \hline
\end{tabular}%
}
\end{table}

\begin{table}[H]
\caption{Aggregated results on Instant-NGP}
\label{tab:agg-ngp}
\resizebox{\textwidth}{!}{%
\begin{tabular}{cccccc|cccc|cccc}
\hline
\multirow{2}{*}{\textit{Dataset}}                      & \multirow{2}{*}{\textit{Metrics}}    & \multicolumn{4}{c|}{\textit{M=3}}                                  & \multicolumn{4}{c|}{\textit{M=4}}                                  & \multicolumn{4}{c}{\textit{M=5}}                                   \\ \cline{3-14} 
                                                       &                                      & \textit{baseline} & \textit{Top-1} & \textit{Top-2} & \textit{Ens} & \textit{baseline} & \textit{Top-1} & \textit{Top-2} & \textit{Ens} & \textit{baseline} & \textit{Top-1} & \textit{Top-2} & \textit{Ens} \\ \hline
\multicolumn{1}{c|}{\multirow{5}{*}{\textit{Blender}}} & \multicolumn{1}{c|}{\textbf{PSNR}$~\uparrow$}   & 33.19             & 33.31          & 33.61          & 33.53        & 33.25             & 33.43          & 33.72          & 33.82        & 33.35             & 33.56          & 33.83          & 34.01        \\
\multicolumn{1}{c|}{}                                  & \multicolumn{1}{c|}{\textbf{SSIM}~$\uparrow$}   & 0.961             & 0.962          & 0.962          & 0.963        & 0.962             & 0.963          & 0.964          & 0.964        & 0.963             & 0.963          & 0.965          & 0.966        \\
\multicolumn{1}{c|}{}                                  & \multicolumn{1}{c|}{\textbf{LPIPS}~$\downarrow$}  & 0.027             & 0.049          & 0.047          & 0.047        & 0.026             & 0.046          & 0.045          & 0.044        & 0.025             & 0.045          & 0.043          & 0.042        \\
\multicolumn{1}{c|}{}                                  & \multicolumn{1}{c|}{$\|w\|_0\downarrow$} & 20                & 13             & 15             & 18           & 26                & 14             & 19             & 22           & 31                & 17             & 21             & 26           \\
\multicolumn{1}{c|}{}                                  & \multicolumn{1}{c|}{\textbf{GFLOPs}~$\downarrow$} & 42                & 66             & 111            & 122          & 44                & 69             & 117            & 165          & 46                & 71             & 123            & 208          \\
\multicolumn{1}{l|}{}                                  & \multicolumn{1}{c|}{\textbf{time}~$\downarrow$}   & 7'                & 23'            & 26'            & 27'          & 9'                & 27'            & 30'            & 30'          & 11'               & 32'            & 34'            & 37'          \\ \hline
\multicolumn{1}{c|}{\multirow{5}{*}{\textit{NSVF}}}             & \multicolumn{1}{c|}{\textbf{PSNR}$~\uparrow$}   & 36.06             & 36.33          & 36.61          & 36.58        & 36.27             & 36.44          & 36.83          & 36.99        & 36.44             & 36.59          & 37.04          & 37.31        \\
\multicolumn{1}{c|}{\textit{}}                         & \multicolumn{1}{c|}{\textbf{SSIM}~$\uparrow$}   & 0.981             & 0.981          & 0.982          & 0.982        & 0.982             & 0.983          & 0.983          & 0.984        & 0.983             & 0.983          & 0.984          & 0.985        \\
\multicolumn{1}{c|}{\textit{}}                         & \multicolumn{1}{c|}{\textbf{LPIPS}~$\downarrow$}  & 0.012             & 0.024          & 0.024          & 0.023        & 0.011             & 0.023          & 0.023          & 0.022        & 0.010             & 0.023          & 0.022          & 0.020        \\
\multicolumn{1}{c|}{\textit{}}                         & \multicolumn{1}{c|}{$\|w\|_0\downarrow$} & 20                & 15             & 16             & 18           & 26                & 16             & 19             & 22           & 30                & 17             & 20             & 27           \\
\multicolumn{1}{c|}{\textit{}}                         & \multicolumn{1}{c|}{\textbf{GFLOPs}~$\downarrow$} & 20                & 46             & 63             & 71           & 27                & 48             & 69             & 96           & 28                & 52             & 72             & 123          \\
\multicolumn{1}{l|}{}                                  & \multicolumn{1}{c|}{\textbf{time}~$\downarrow$}   & 7'                & 24'            & 25'            & 27'          & 8'                & 28'            & 29'            & 32'          & 10'               & 31'            & 33'            & 34'          \\ \hline
\multicolumn{1}{c|}{\multirow{5}{*}{\textit{TaT}}}                      & \multicolumn{1}{c|}{\textbf{PSNR}$~\uparrow$}   & 28.90             & 28.93          & 29.04          & 29.04        & 29.00             & 29.02          & 29.18          & 29.21        & 29.07             & 29.16          & 29.32          & 29.38        \\
\multicolumn{1}{c|}{\textit{}}                         & \multicolumn{1}{c|}{\textbf{SSIM}~$\uparrow$}   & 0.919             & 0.920          & 0.918          & 0.922        & 0.922             & 0.924          & 0.928          & 0.929        & 0.924             & 0.927          & 0.929          & 0.930        \\
\multicolumn{1}{c|}{\textit{}}                         & \multicolumn{1}{c|}{\textbf{LPIPS}~$\downarrow$}  & 0.107             & 0.131          & 0.130          & 0.129        & 0.105             & 0.127          & 0.125          & 0.124        & 0.101             & 0.125          & 0.124          & 0.121        \\
\multicolumn{1}{c|}{\textit{}}                         & \multicolumn{1}{c|}{$\|w\|_0\downarrow$} & 45                & 27             & 35             & 40           & 65                & 31             & 39             & 55           & 88                & 37             & 47             & 70           \\
\multicolumn{1}{c|}{\textit{}}                         & \multicolumn{1}{c|}{\textbf{GFLOPs}~$\downarrow$} & 210               & 220            & 498            & 590          & 215               & 231            & 522            & 785          & 211               & 229            & 531            & 1003         \\
\multicolumn{1}{l|}{}                                  & \multicolumn{1}{c|}{\textbf{time}~$\downarrow$}   & 9'                & 30'            & 30'            & 32'          & 11'               & 33'            & 34'            & 34'          & 14'               & 37'            & 39'            & 45'          \\ \hline
\multicolumn{1}{c|}{\multirow{5}{*}{\textit{LLFF}}}                     & \multicolumn{1}{c|}{\textbf{PSNR}$~\uparrow$}   & 24.82             & 24.78          & 25.06          & 25.06        & 24.93             & 24.85          & 25.11          & 25.13        & 24.97             & 24.90          & 25.17          & 25.19        \\
\multicolumn{1}{c|}{}                                  & \multicolumn{1}{c|}{\textbf{SSIM}~$\uparrow$}   & 0.764             & 0.762          & 0.772          & 0.772        & 0.763             & 0.765          & 0.777          & 0.776        & 0.764             & 0.763          & 0.777          & 0.778        \\
\multicolumn{1}{c|}{}                                  & \multicolumn{1}{c|}{\textbf{LPIPS}~$\downarrow$}  & 0.134             & 0.244          & 0.241          & 0.240        & 0.130             & 0.243          & 0.238          & 0.237        & 0.128             & 0.239          & 0.237          & 0.234        \\
\multicolumn{1}{c|}{}                                  & \multicolumn{1}{c|}{$\|w\|_0\downarrow$} & 72                & 37             & 57             & 67           & 106               & 57             & 62             & 102          & 152               & 68             & 75             & 148          \\
\multicolumn{1}{c|}{}                                  & \multicolumn{1}{c|}{\textbf{GFLOPs}~$\downarrow$} & 539               & 761            & 1227           & 1607         & 529               & 882            & 1272           & 2172         & 573               & 887            & 1391           & 2712         \\
\multicolumn{1}{l|}{}                                  & \multicolumn{1}{c|}{\textbf{time}~$\downarrow$}   & 14'               & 20'            & 22'            & 22'          & 18'               & 28'            & 30'            & 29'          & 24'               & 25'            & 46'            & 42'          \\ \hline
\end{tabular}%
}
\end{table}

\begin{table}[H]
\caption{DVGO results on Synthetic-NeRF}
\label{tab:dvgo_1}
\resizebox{\textwidth}{!}{%
\begin{tabular}{ccccccc|ccccc|ccccc}
\hline
\multirow{2}{*}{\textit{Method}}                        & \multirow{2}{*}{\textit{Scene}}            & \multicolumn{5}{c|}{\textit{M=3}}                                                & \multicolumn{5}{c|}{\textit{M=4}}                                                & \multicolumn{5}{c}{\textit{M=5}}                                                 \\ \cline{3-17} 
                                                        &                                            & \textbf{PSNR}~$\ $  & \textbf{SSIM}~$\ $  & \textbf{LPIPS}~$\  $ & $\|w\||_0\  $ & \textbf{GFLOPs} & \textbf{PSNR}~$\ $  & \textbf{SSIM}~$\ $  & \textbf{LPIPS}~$\  $ & $\|w\||_0\  $ & \textbf{GFLOPs} & \textbf{PSNR}~$\ $  & \textbf{SSIM}~$\ $  & \textbf{LPIPS}~$\  $ & $\|w\||_0\  $ & \textbf{GFLOPs} \\ \hline
\multicolumn{1}{c|}{\multirow{9}{*}{\textit{Baseline}}} & \multicolumn{1}{c|}{\textit{Lego}}         & \textit{35.99} & 0.981          & 0.009          & 9           & 224             & 35.85          & 0.980          & 0.009          & 13          & 324             & 35.52          & 0.978          & 0.010          & 26          & 929             \\
\multicolumn{1}{c|}{}                                   & \multicolumn{1}{c|}{\textit{Mic}}          & 35.47          & 0.990          & 0.006          & 7           & 85              & 36.24          & 0.991          & 0.005          & 9           & 127             & 36.68          & 0.992          & 0.005          & 18          & 845             \\
\multicolumn{1}{c|}{}                                   & \multicolumn{1}{c|}{\textit{Chair}}        & 35.70          & 0.985          & 0.009          & 9           & 96              & 35.95          & 0.986          & 0.009          & 12          & 175             & 36.01          & 0.986          & 0.009          & 23          & 703             \\
\hline
\end{tabular}%
}
\end{table}

\begin{table}[h]
\caption{Per-scene results on NSVF with DVGO}
\label{tab:dvgo_2}
\resizebox{\textwidth}{!}{%
\begin{tabular}{ccccccc|ccccc|ccccc}
\hline
\multirow{2}{*}{\textit{Method}}                        & \multirow{2}{*}{\textit{Scene}}            & \multicolumn{5}{c|}{\textit{M=3}}                                                & \multicolumn{5}{c|}{\textit{M=4}}                                                & \multicolumn{5}{c}{\textit{M=5}}                                                  \\ \cline{3-17} 
                                                        &                                            & \textbf{PSNR}~$\ $  & \textbf{SSIM}~$\ $  & \textbf{LPIPS}~$\  $ & $\|w\||_0\  $ & \textbf{GFLOPs} & \textbf{PSNR}~$\ $  & \textbf{SSIM}~$\ $  & \textbf{LPIPS}~$\  $ & $\|w\||_0\  $ & \textbf{GFLOPs} & \textbf{PSNR}~$\ $  & \textbf{SSIM}~$\ $  & \textbf{LPIPS}~$\  $ & $\|w\||_0\  $  & \textbf{GFLOPs} \\ \hline
\multicolumn{1}{c|}{\multirow{9}{*}{\textit{Baseline}}} & \multicolumn{1}{c|}{\textit{Bike}}         & \textit{38.13} & 0.991          & 0.004          & 16          & 269             & 37.70          & 0.990          & 0.004          & 32          & 369             & 37.25          & 0.989          & 0.005          & 48           & 448             \\
\multicolumn{1}{c|}{}                                   & \multicolumn{1}{c|}{\textit{Lifestyle}}    & 34.27          & 0.969          & 0.021          & 20          & 369             & 34.34          & 0.969          & 0.021          & 42          & 505             & 34.24          & 0.968          & 0.022          & 65           & 619             \\
\multicolumn{1}{c|}{}                                   & \multicolumn{1}{c|}{\textit{Palace}}       & 36.59          & 0.974          & 0.015          & 21          & 189             & 37.01          & 0.976          & 0.013          & 43          & 264             & 36.90          & 0.975          & 0.012          & 66           & 319             \\
\hline
\end{tabular}%
}
\end{table}

\begin{table}[h]
\caption{Per-scene results on TanksAndTemple with DVGO}
\label{tab:dvgo_3}
\resizebox{\textwidth}{!}{%
\begin{tabular}{ccccccc|ccccc|ccccc}
\hline
\multirow{2}{*}{\textit{Method}}                        & \multirow{2}{*}{\textit{Scene}}           & \multicolumn{5}{c|}{\textit{M=3}}                                                & \multicolumn{5}{c|}{\textit{M=4}}                                                & \multicolumn{5}{c}{\textit{M=5}}                                                 \\ \cline{3-17} 
                                                        &                                           & \textbf{PSNR}~$\ $  & \textbf{SSIM}~$\ $  & \textbf{LPIPS}~$\  $ & $\|w\||_0\  $ & \textbf{GFLOPs} & \textbf{PSNR}~$\ $  & \textbf{SSIM}~$\ $  & \textbf{LPIPS}~$\  $ & $\|w\||_0\  $ & \textbf{GFLOPs} & \textbf{PSNR}~$\ $  & \textbf{SSIM}~$\ $  & \textbf{LPIPS}~$\  $ & $\|w\||_0\  $ & \textbf{GFLOPs} \\ \hline
\multicolumn{1}{c|}{\multirow{6}{*}{\textit{Baseline}}} & \multicolumn{1}{c|}{\textit{Ignatius}}    & \textit{28.39} & 0.95           & 0.076          & 20          & 1588            & 28.55          & 0.951          & 0.07           & 35          & 1755            & 27.86          & 0.873          & 0.207          & 84          & 3490            \\
\multicolumn{1}{c|}{}                                   & \multicolumn{1}{c|}{\textit{Caterpillar}} & 26.06          & 0.911          & 0.137          & 9           & 1298            & 26.08          & 0.912          & 0.129          & 31          & 1012            & 27.68          & 0.924          & 0.102          & 59          & 1592            \\
\multicolumn{1}{c|}{}                                   & \multicolumn{1}{c|}{\textit{Family}}      & 33.76          & 0.964          & 0.057          & 23          & 1293            & 33.75          & 0.964          & 0.053          & 20          & 1647            & 26.13          & 0.914          & 0.119          & 51          & 2561            \\
\hline
\end{tabular}%
}
\end{table}

\begin{table}[h]
\caption{Per-scene results on LLFF with DVGO}
\label{tab:dvgo_4}
\resizebox{\textwidth}{!}{%
\begin{tabular}{ccccccc|ccccc|ccccc}
\hline
\multirow{2}{*}{\textit{Method}}                        & \multirow{2}{*}{\textit{Scene}}            & \multicolumn{5}{c|}{\textit{M=3}}                                                & \multicolumn{5}{c|}{\textit{M=4}}                                                & \multicolumn{5}{c}{\textit{M=5}}                                                  \\ \cline{3-17} 
                                                        &                                            & \textbf{PSNR}~$\ $  & \textbf{SSIM}~$\ $  & \textbf{LPIPS}~$\  $ & $\|w\||_0\  $ & \textbf{GFLOPs} & \textbf{PSNR}~$\ $  & \textbf{SSIM}~$\ $  & \textbf{LPIPS}~$\  $ & $\|w\||_0\  $ & \textbf{GFLOPs} & \textbf{PSNR}~$\ $  & \textbf{SSIM}~$\ $  & \textbf{LPIPS}~$\  $ & $\|w\||_0\  $  & \textbf{GFLOPs} \\ \hline
\multicolumn{1}{c|}{\multirow{9}{*}{\textit{Baseline}}} & \multicolumn{1}{c|}{\textit{Fern}}         & \textit{24.89} & 0.816          & 0.166          & 28          & 1062            & 24.76          & 0.816          & 0.161          & 49          & 1382            & 24.58          & 0.810          & 0.160          & 95           & 1859            \\
\multicolumn{1}{c|}{}                                   & \multicolumn{1}{c|}{\textit{Flower}}       & 27.51          & 0.850          & 0.132          & 16          & 1018            & 27.65          & 0.854          & 0.127          & 28          & 1295            & 27.67          & 0.856          & 0.122          & 53           & 1690            \\
\multicolumn{1}{c|}{}                                   & \multicolumn{1}{c|}{\textit{Fortress}}     & 29.91          & 0.859          & 0.121          & 8           & 640             & 29.42          & 0.831          & 0.143          & 16          & 907             & 29.38          & 0.821          & 0.149          & 36           & 1407            \\
\hline
\end{tabular}%
}
\end{table}

\begin{table}[h]
\caption{Per-scene results on Synthetic-NeRF with TensorF}
\label{tab:tensorf_1}
\resizebox{\textwidth}{!}{%
\begin{tabular}{ccccccc|ccccc|ccccc}
\hline
\multirow{2}{*}{\textit{Method}}                        & \multirow{2}{*}{\textit{Scene}}         & \multicolumn{5}{c|}{\textit{M=3}}                                                & \multicolumn{5}{c|}{\textit{M=4}}                                                & \multicolumn{5}{c}{\textit{M=5}}                                                 \\ \cline{3-17} 
                                                        &                                         & \textbf{PSNR}  & \textbf{SSIM}  & \textbf{LPIPS} & $\|w\||_0   $ & \textbf{GFLOPs} & \textbf{PSNR}  & \textbf{SSIM}  & \textbf{LPIPS} & $\|w\||_0  $ & \textbf{GFLOPs} & \textbf{PSNR}  & \textbf{SSIM}  & \textbf{LPIPS} & $\|w\||_0   $ & \textbf{GFLOPs} \\ \hline
\multicolumn{1}{c|}{\multirow{9}{*}{\textit{Baseline}}} & \multicolumn{1}{c|}{\textit{Lego}}      & \textit{36.85} & 0.986          & 0.006          & 15          & 376             & 36.73          & 0.986          & 0.006          & 25          & 474             & 36.70          & 0.986          & 0.007          & 38          & 579             \\
\multicolumn{1}{c|}{}                                   & \multicolumn{1}{c|}{\textit{Mic}}       & 34.98          & 0.994          & 0.005          & 8           & 188             & 34.80          & 0.994          & 0.005          & 13          & 259             & 34.45          & 0.994          & 0.005          & 19          & 307             \\
\multicolumn{1}{c|}{}                                   & \multicolumn{1}{c|}{\textit{Chair}}     & 30.89          & 0.879          & 0.088          & 13          & 154             & 30.80          & 0.879          & 0.078          & 22          & 198             & 30.70          & 0.871          & 0.079          & 33          & 255             \\
\hline
\end{tabular}%
}
\end{table}

\begin{table}[h]
\caption{Per-scene results on NSVF with TensoRF}
\label{tab:tensorf_2}
\resizebox{\textwidth}{!}{%
\begin{tabular}{ccccccc|ccccc|ccccc}
\hline
\multirow{2}{*}{\textit{Method}}                        & \multirow{2}{*}{\textit{Scene}}            & \multicolumn{5}{c|}{\textit{M=3}}                                                & \multicolumn{5}{c|}{\textit{M=4}}                                                & \multicolumn{5}{c}{\textit{M=5}}                                                 \\ \cline{3-17} 
                                                        &                                            & \textbf{PSNR}  & \textbf{SSIM}  & \textbf{LPIPS} & $\|w\||_0   $ & \textbf{GFLOPs} & \textbf{PSNR}  & \textbf{SSIM}  & \textbf{LPIPS} & $\|w\||_0   $ & \textbf{GFLOPs} & \textbf{PSNR}  & \textbf{SSIM}  & \textbf{LPIPS} & $\|w\||_0   $ & \textbf{GFLOPs} \\ \hline
\multicolumn{1}{c|}{\multirow{9}{*}{\textit{Baseline}}} & \multicolumn{1}{c|}{\textit{Bike}}         & \textit{39.48} & 0.996          & 0.003          & 14          & 304             & 39.25          & 0.996          & 0.003          & 23          & 362             & 38.59          & 0.995          & 0.004          & 35          & 421             \\
\multicolumn{1}{c|}{}                                   & \multicolumn{1}{c|}{\textit{Lifestyle}}    & 34.76          & 0.975          & 0.015          & 18          & 680             & 34.71          & 0.974          & 0.016          & 30          & 833             & 34.53          & 0.973          & 0.019          & 45          & 990             \\
\multicolumn{1}{c|}{}                                   & \multicolumn{1}{c|}{\textit{Palace}}       & 38.20          & 0.983          & 0.008          & 15          & 541             & 38.51          & 0.984          & 0.008          & 25          & 671             & 38.60          & 0.984          & 0.007          & 38          & 803             \\
\hline
\end{tabular}%
}
\end{table}

\begin{table}[h]
\caption{Per-scene results on TanksAndTemple with TensoRF}
\label{tab:tensorf_3}
\resizebox{\textwidth}{!}{%
\begin{tabular}{ccccccc|ccccc|ccccc}
\hline
\multirow{2}{*}{\textit{Method}}                        & \multirow{2}{*}{\textit{Scene}}            & \multicolumn{5}{c|}{\textit{M=3}}                                                & \multicolumn{5}{c|}{\textit{M=4}}                                                & \multicolumn{5}{c}{\textit{M=5}}                                                 \\ \cline{3-17} 
                                                        &                                            & \textbf{PSNR}  & \textbf{SSIM}  & \textbf{LPIPS} & $\|w\||_0   $ & \textbf{GFLOPs} & \textbf{PSNR}  & \textbf{SSIM}  & \textbf{LPIPS} & $\|w\||_0 $ & \textbf{GFLOPs} & \textbf{PSNR}  & \textbf{SSIM}  & \textbf{LPIPS} & $\|w\||_0  $ & \textbf{GFLOPs} \\ \hline
\multicolumn{1}{c|}{\multirow{6}{*}{\textit{Baseline}}} & \multicolumn{1}{c|}{\textit{Caterpillar}}  & \textit{26.16} & 0.842          & 0.191          & 3           & 3155            & 25.83          & 0.847          & 0.177          & 5           & 3674            & 25.82          & 0.845          & 0.172          & 7           & 4557            \\
\multicolumn{1}{c|}{}                                   & \multicolumn{1}{c|}{\textit{Family}}       & 34.20          & 0.963          & 0.059          & 5           & 3071            & 34.21          & 0.963          & 0.053          & 8           & 3578            & 33.94          & 0.963          & 0.050          & 13          & 4366            \\
\multicolumn{1}{c|}{}                                   & \multicolumn{1}{c|}{\textit{Barn}}         & 27.59          & 0.949          & 0.076          & 0           & 1841            & 27.68          & 0.951          & 0.066          & 1           & 2489            & 27.68          & 0.951          & 0.064          & 1           & 2917            \\
\hline
\end{tabular}%
}
\end{table}

\begin{table}[h]
\caption{Per-scene results on LLFF with TensoRF}
\label{tab:tensorf_4}
\resizebox{\textwidth}{!}{%
\begin{tabular}{ccccccc|ccccc|ccccc}
\hline
\multirow{2}{*}{\textit{Method}}                        & \multirow{2}{*}{\textit{Scene}}            & \multicolumn{5}{c|}{\textit{M=3}}                                                & \multicolumn{5}{c|}{\textit{M=4}}                                                & \multicolumn{5}{c}{\textit{M=5}}                                                 \\ \cline{3-17} 
                                                        &                                            & \textbf{PSNR}  & \textbf{SSIM}  & \textbf{LPIPS} & $\|w\||_0   $ & \textbf{GFLOPs} & \textbf{PSNR}  & \textbf{SSIM}  & \textbf{LPIPS} & $\|w\||_0   $ & \textbf{GFLOPs} & \textbf{PSNR}  & \textbf{SSIM}  & \textbf{LPIPS} & $\|w\||_0   $ & \textbf{GFLOPs} \\ \hline
\multicolumn{1}{c|}{\multirow{9}{*}{\textit{Baseline}}} & \multicolumn{1}{c|}{\textit{Fern}}         & \textit{24.98} & 0.801          & 0.165          & 9           & 4124            & 25.01          & 0.798          & 0.162          & 13          & 4766            & 25.04          & 0.801          & 0.152          & 18          & 5671            \\
\multicolumn{1}{c|}{}                                   & \multicolumn{1}{c|}{\textit{Flower}}       & 28.41          & 0.863          & 0.108          & 9           & 3413            & 28.38          & 0.865          & 0.099          & 13          & 4048            & 28.33          & 0.863          & 0.097          & 19          & 5054            \\
\multicolumn{1}{c|}{}                                   & \multicolumn{1}{c|}{\textit{Fortress}}     & 31.18          & 0.892          & 0.079          & 7           & 3123            & 31.27          & 0.895          & 0.072          & 11          & 3922            & 31.43          & 0.899          & 0.066          & 15          & 4872            \\
\hline
\end{tabular}%
}
\end{table}

\begin{table}[h]
\caption{Per-scene results on Synthetic-NeRF with Instant-NGP}
\label{tab:ngp_1}
\resizebox{\textwidth}{!}{%
\begin{tabular}{ccccccc|ccccc|ccccc}
\hline
\multirow{2}{*}{\textit{Method}}                        & \multirow{2}{*}{\textit{Scene}}            & \multicolumn{5}{c|}{\textit{M=3}}                                                & \multicolumn{5}{c|}{\textit{M=4}}                                                & \multicolumn{5}{c}{\textit{M=5}}                                                 \\ \cline{3-17} 
                                                        &                                            & \textbf{PSNR}  & \textbf{SSIM}  & \textbf{LPIPS} & $\|w\||_0   $ & \textbf{GFLOPs} & \textbf{PSNR}  & \textbf{SSIM}  & \textbf{LPIPS} & $\|w\||_0  $ & \textbf{GFLOPs} & \textbf{PSNR}  & \textbf{SSIM}  & \textbf{LPIPS} & $\|w\||_0   $ & \textbf{GFLOPs} \\ \hline
\multicolumn{1}{c|}{\multirow{9}{*}{\textit{Baseline}}} & \multicolumn{1}{c|}{\textit{Lego}}         & \textit{35.88} & 0.980          & 0.009          & 12          & 23              & 35.98          & 0.981          & 0.008          & 14          & 29              & 36.09          & 0.981          & 0.008          & 15          & 30              \\
\multicolumn{1}{c|}{}                                   & \multicolumn{1}{c|}{\textit{Mic}}          & 36.25          & 0.990          & 0.008          & 10          & 18              & 36.27          & 0.991          & 0.007          & 12          & 20              & 36.66          & 0.991          & 0.007          & 13          & 22              \\
\multicolumn{1}{c|}{}                                   & \multicolumn{1}{c|}{\textit{Ship}}         & 30.69          & 0.892          & 0.080          & 33          & 130             & 30.58          & 0.893          & 0.075          & 46          & 137             & 30.59          & 0.894          & 0.073          & 51          & 145             \\
\hline
\end{tabular}%
}
\end{table}

\begin{table}[h]
\caption{Per-scene results on NSVF with Instant-NGP}
\label{tab:ngp_2}
\resizebox{\textwidth}{!}{%
\begin{tabular}{ccccccc|ccccc|ccccc}
\hline
\multirow{2}{*}{\textit{Method}}                        & \multirow{2}{*}{\textit{Scene}}            & \multicolumn{5}{c|}{\textit{M=3}}                                                & \multicolumn{5}{c|}{\textit{M=4}}                                                & \multicolumn{5}{c}{\textit{M=5}}                                                 \\ \cline{3-17} 
                                                        &                                            & \textbf{PSNR}  & \textbf{SSIM}  & \textbf{LPIPS} & $\|w\||_0   $ & \textbf{GFLOPs} & \textbf{PSNR}  & \textbf{SSIM}  & \textbf{LPIPS} & $\|w\||_0   $ & \textbf{GFLOPs} & \textbf{PSNR}  & \textbf{SSIM}  & \textbf{LPIPS} & $\|w\||_0 $ & \textbf{GFLOPs} \\ \hline
\multicolumn{1}{c|}{\multirow{9}{*}{\textit{Baseline}}} & \multicolumn{1}{c|}{\textit{Bike}}         & \textit{37.86} & 0.990          & 0.004          & 13          & 12              & 38.04          & 0.991          & 0.004          & 17          & 17              & 38.45          & 0.991          & 0.003          & 20          & 17              \\
\multicolumn{1}{c|}{}                                   & \multicolumn{1}{c|}{\textit{Lifestyle}}    & 34.80          & 0.968          & 0.019          & 24          & 31              & 34.95          & 0.970          & 0.018          & 30          & 49              & 35.05          & 0.970          & 0.017          & 35          & 51              \\
\multicolumn{1}{c|}{}                                   & \multicolumn{1}{c|}{\textit{Palace}}       & 37.66          & 0.978          & 0.010          & 27          & 31              & 38.18          & 0.981          & 0.008          & 31          & 35              & 38.38          & 0.982          & 0.008          & 32          & 37              \\
\hline
\end{tabular}%
}
\end{table}

\begin{table}[h]
\caption{Per-scene results on TanksAndTemple with Instant-NGP}
\label{tab:ngp_3}
\resizebox{\textwidth}{!}{%
\begin{tabular}{ccccccc|ccccc|ccccc}
\hline
\multirow{2}{*}{\textit{Method}}                        & \multirow{2}{*}{\textit{Scene}}           & \multicolumn{5}{c|}{\textit{M=3}}                                                & \multicolumn{5}{c|}{\textit{M=4}}                                                & \multicolumn{5}{c}{\textit{M=5}}                                                 \\ \cline{3-17} 
                                                        &                                           & \textbf{PSNR}  & \textbf{SSIM}  & \textbf{LPIPS} & $\|w\||_0   $ & \textbf{GFLOPs} & \textbf{PSNR}  & \textbf{SSIM}  & \textbf{LPIPS} & $\|w\||_0   $ & \textbf{GFLOPs} & \textbf{PSNR}  & \textbf{SSIM}  & \textbf{LPIPS} & $\|w\||_0   $ & \textbf{GFLOPs} \\ \hline
\multicolumn{1}{c|}{\multirow{6}{*}{\textit{Baseline}}} & \multicolumn{1}{c|}{\textit{Ignatius}}    & \textit{28.48} & 0.947          & 0.070          & 63          & 130             & 28.41          & 0.947          & 0.070          & 78          & 137             & 28.44          & 0.948          & 0.069          & 98          & 124             \\
\multicolumn{1}{c|}{}                                   & \multicolumn{1}{c|}{\textit{Caterpillar}} & 26.30          & 0.913          & 0.111          & 47          & 178             & 26.32          & 0.915          & 0.109          & 63          & 174             & 26.42          & 0.915          & 0.107          & 92          & 178             \\
\multicolumn{1}{c|}{}                                   & \multicolumn{1}{c|}{\textit{Family}}      & 34.33          & 0.964          & 0.039          & 27          & 152             & 34.43          & 0.966          & 0.037          & 46          & 155             & 34.47          & 0.967          & 0.037          & 69          & 167             \\
\hline
\end{tabular}%
}
\end{table}

\begin{table}[]
\caption{Per-scene results on LLFF with Instant-NGP}
\label{tab:ngp_4}
\resizebox{\textwidth}{!}{%
\begin{tabular}{ccccccc|ccccc|ccccc}
\hline
\multirow{2}{*}{\textit{Method}}                        & \multirow{2}{*}{\textit{Scene}}            & \multicolumn{5}{c|}{\textit{M=3}}                                                & \multicolumn{5}{c|}{\textit{M=4}}                                                  & \multicolumn{5}{c}{\textit{M=5}}                                                  \\ \cline{3-17} 
                                                        &                                            & \textbf{PSNR}  & \textbf{SSIM}  & \textbf{LPIPS} & $\|w\||_0   $ & \textbf{GFLOPs} & \textbf{PSNR}  & \textbf{SSIM}  & \textbf{LPIPS}  & $\|w\||_0  $  & \textbf{GFLOPs} & \textbf{PSNR}  & \textbf{SSIM}  & \textbf{LPIPS} & $\|w\||_0 $    & \textbf{GFLOPs} \\ \hline
\multicolumn{1}{c|}{\multirow{9}{*}{\textit{Baseline}}} & \multicolumn{1}{c|}{\textit{Fern}}         & \textit{23.48} & 0.741          & 0.149          & 95          & 631             & 23.20          & 0.746          & 0.148           & 146          & 676             & 23.29          & 0.748          & 0.143          & 208          & 697             \\
\multicolumn{1}{c|}{}                                   & \multicolumn{1}{c|}{\textit{Flower}}       & 26.94          & 0.806          & 0.121          & 59          & 611             & 26.84          & 0.807          & 0.120           & 86           & 581             & 26.90          & 0.809          & 0.116          & 112          & 586             \\
\multicolumn{1}{c|}{}                                   & \multicolumn{1}{c|}{\textit{Fortress}}     & 27.13          & 0.821          & 0.089          & 44          & 560             & 28.24          & 0.823          & 0.079           & 56           & 447             & 28.48          & 0.824          & 0.075          & 77           & 761             \\
\hline
\end{tabular}%
}
\end{table}


%
%
\clearpage
\bibliographystyle{splncs04}
\bibliography{main}

\end{document}